\documentclass[11pt,a4paper]{book}
\usepackage[utf8]{inputenc}
\usepackage{refstyle}
\usepackage{float}
\usepackage{calc}
\usepackage{amsmath}
\usepackage{amsthm}
\usepackage{amssymb}
\usepackage{graphicx}
\usepackage[authoryear]{natbib}
\usepackage[unicode=true]
 {hyperref}
\usepackage{breakurl}

\makeatletter

\AtBeginDocument{\providecommand\chapref[1]{\ref{chap:#1}}}
\AtBeginDocument{\providecommand\figref[1]{\ref{fig:#1}}}
\AtBeginDocument{\providecommand\secref[1]{\ref{sec:#1}}}
\AtBeginDocument{\providecommand\eqref[1]{\ref{eq:#1}}}
\AtBeginDocument{\providecommand\subsecref[1]{\ref{subsec:#1}}}
\AtBeginDocument{\providecommand\exref[1]{\ref{ex:#1}}}
\AtBeginDocument{\providecommand\defref[1]{\ref{def:#1}}}
\AtBeginDocument{\providecommand\thmref[1]{\ref{thm:#1}}}
\AtBeginDocument{\providecommand\enuref[1]{\ref{enu:#1}}}
\AtBeginDocument{\providecommand\algref[1]{\ref{alg:#1}}}
\AtBeginDocument{\providecommand\tabref[1]{\ref{tab:#1}}}

\let\SF@@footnote\footnote
\def\footnote{\ifx\protect\@typeset@protect
    \expandafter\SF@@footnote
  \else
    \expandafter\SF@gobble@opt
  \fi
}
\expandafter\def\csname SF@gobble@opt \endcsname{\@ifnextchar[
  \SF@gobble@twobracket
  \@gobble
}
\edef\SF@gobble@opt{\noexpand\protect
  \expandafter\noexpand\csname SF@gobble@opt \endcsname}
\def\SF@gobble@twobracket[#1]#2{}
\providecommand{\tabularnewline}{\\}
\floatstyle{ruled}
\newfloat{algorithm}{tbp}{loa}[chapter]
\providecommand{\algorithmname}{Algorithm}
\floatname{algorithm}{\protect\algorithmname}
\RS@ifundefined{subsecref}
  {\newref{subsec}{name = \RSsectxt}}
  {}
\RS@ifundefined{thmref}
  {\def\RSthmtxt{theorem~}\newref{thm}{name = \RSthmtxt}}
  {}
\RS@ifundefined{lemref}
  {\def\RSlemtxt{lemma~}\newref{lem}{name = \RSlemtxt}}
  {}

\theoremstyle{plain}
\newtheorem{thm}{\protect\theoremname}
  \theoremstyle{definition}
  \newtheorem{defn}{\protect\definitionname}
 \theoremstyle{definition}
  \newtheorem{example}{\protect\examplename}

\usepackage{styles/anuthesis}
\usepackage{styles/fancyhdr}
\usepackage{styles/aixi}

\hypersetup{colorlinks=true,linkcolor=[rgb]{0,0.16,0.6},urlcolor=[rgb]{0,0.16,0.6},citecolor=[rgb]{0,0.16,0.6}}
\usepackage{pgf}
\usepackage{todonotes}
\usepackage{amsfonts}
\usepackage{amsmath}
\usepackage{amssymb}
\usepackage{algorithm,algpseudocode}
\usepackage{tikz}
\usepackage{dirtree}
\usepackage{menukeys}

\def\setmenukeyswin{\def\tw@mk@os{win}}
\def\setmenukeysmac{\def\tw@mk@os{mac}}

\usetikzlibrary{arrows,shapes,positioning,automata,decorations.pathmorphing,snakes}
\usepackage{styles/tikz-uml}
\setcitestyle{round}

\newcommand{\Continue}{\textbf{continue}}

\let\algref=\relax
\newref{fig}{refcmd={\hyperref[#1]{Figure \ref{#1}}}}
\newref{tab}{refcmd={\hyperref[#1]{Table \ref{#1}}}}
\newref{eq}{refcmd={\hyperref[#1]{Equation (\ref{#1})}}}
\newref{ex}{refcmd={\hyperref[#1]{Example \ref{#1}}}}
\newref{sec}{refcmd={\hyperref[#1]{Section \ref{#1}}}}
\newref{sub}{refcmd={\hyperref[#1]{Subsection \ref{#1}}}}
\newref{alg}{refcmd={\hyperref[#1]{Algorithm \ref{#1}}}}
\newref{fig}{refcmd={\hyperref[#1]{Figure \ref{#1}}}}
\newref{chap}{refcmd={\hyperref[#1]{Chapter \ref{#1}}}}
\newref{thm}{refcmd={\hyperref[#1]{Theorem \ref{#1}}}}
\newref{def}{refcmd={\hyperref[#1]{Definition \ref{#1}}}}
\newref{enu}{refcmd={\hyperref[#1]{Item \ref{#1}}}}

\global\long\def\ts#1{\textsc{#1}}

\makeatother

  \providecommand{\definitionname}{Definition}
  \providecommand{\examplename}{Example}
\providecommand{\theoremname}{Theorem}

\begin{document}
\pagenumbering{roman} 

\begin{titlepage} 

\addcontentsline{toc}{chapter}{Title Page}

\title{\textsc{\Huge{}AIXIjs}{\Huge{}: A Software Demo for}\\
{\Huge{}General Reinforcement Learning}\\[1cm]}

\author{{\huge{}John Stewart Aslanides}\\[5cm] {\LARGE{}A thesis submitted
in partial fulfillment of the degree of}\\
\textbf{ }{\huge{}Master of Computing (Advanced)}{\Large{}}\\
{\LARGE{}at the}\\
{\huge{}Australian National University}\\[1cm]\textbf{\includegraphics[scale=0.3]{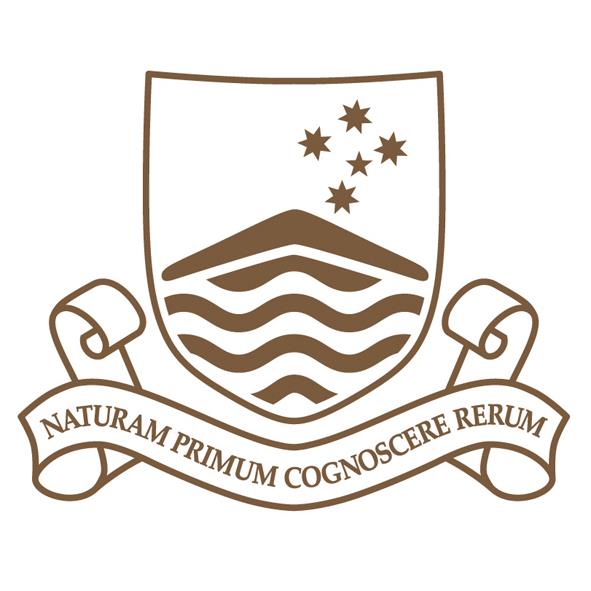}}}

\date{{\huge{}October 2016}}

\maketitle
\end{titlepage}

\sloppy

\chapter*{Declaration}

\addcontentsline{toc}{chapter}{Declaration}

This thesis is an account of research undertaken between March 2016
and October 2016 at The Research School of Computer Science, The Australian
National University, Canberra, Australia.

Except where acknowledged in the customary manner, the material presented
in this thesis is, to the best of my knowledge, original and has not
been submitted in whole or part for a degree in any university.

\vspace{20mm}

\hspace{80mm}\rule{40mm}{0.15mm}

\hspace{80mm} John Stewart Aslanides

\hspace{80mm} 27 October, 2016

\vspace{80mm}

Supervisors:
\begin{itemize}
\item \href{https://jan.leike.name/}{Dr. Jan Leike} (Future of Humanity
Institute, University of Oxford)
\item \href{http://hutter1.net}{Prof. Marcus Hutter} (Australian National
University)
\end{itemize}
\vspace{4mm}

Convenor:
\begin{itemize}
\item \href{http://users.cecs.anu.edu.au/~jks/}{Prof. John Slaney} (Australian
National University)
\end{itemize}

\chapter*{Acknowledgements}

\addcontentsline{toc}{chapter}{Acknowledgements}

And so, my formal education comes to an end, at least for the time
being. Naturally, one cannot take credit for one's successes any more
than one can take credit for one's genes and environment. I owe \emph{everything}
to having been fortunate enough to grow up in a wealthy and peaceful
country (Australia), with loving and well-educated parents (Jenny
and Timoshenko), and to having been exposed to the quality of tuition
and mentorship that I've received over the years at the Australian
National University. In my time at the ANU, I've met many smart people
who have, to varying degrees, spurred my intellectual development
and shaped how I think. They are (in order of appearance): \href{http://people.physics.anu.edu.au/~cms130/}{Craig Savage},
\href{http://www.mso.anu.edu.au/pfrancis/}{Paul Francis}, \href{https://physics.anu.edu.au/people/profile.php?ID=285}{John Close},
\href{https://physics.anu.edu.au/people/profile.php?ID=292}{Joe Hope},
\href{http://www.phys.ttu.edu/~raata/}{Ra Inta}, \href{http://users.cecs.anu.edu.au/~williams/}{Bob Williamson},
\href{https://people.cs.umass.edu/~domke/}{Justin Domke}, \href{http://www.web-port.net/Christfried.Webers/}{Christfried Webers},
\href{https://jan.leike.name}{Jan Leike}, and \href{http://hutter1.net}{Marcus Hutter}.
To these mentors and teachers, past and present: thank you. 
\begin{itemize}
\item To Jan, my supervisor: thank you for agreeing to supervise me from
across the world, and for always being easy-going and genial, despite
having to wake up so early for all of those Skype meetings. I hope
that it's obvious that I'm extremely glad to have gotten to know you
over the course of this year. I really hope that we see each other
again soon.
\item To Marcus: although we didn't collaborate directly, it has been an
honour to follow your work, and to pick your brain at group meetings.
I love your sense of humour. Also, good job discovering AIXI; overall,
I think it's a pretty neat idea.
\item To Jarryd: getting to know you has been one of the highlights of this
year for me. A more true, honest, intelligent, and kind friend and
lab partner I cannot conceive of. I'm going to miss you when you're
the CEO of DeepMind. :)
\item Thanks to my other friends and colleagues in the Intelligent Agents
research group: Suraj Narayanan, \href{http://www.tomeveritt.se}{Tom Everitt},
Boris Repasky, Manlio Valenti, Sean Lamont, and Sultan Javed. Before
I met you guys, I didn't know the meaning of the phrase `hard work'.
Ours is the lab that never sleeps! In particular, I'd like to thank
Tom for going to the trouble of introducing me via email to many of
his connections in the Bay area. Meeting these people added immense
value to my trip, and I now have valuable connections with members
of the AI community in the US because of this.
\item Thanks also to the \href{http://futureoflife.org/}{Future of Life Institute}
for sponsoring my travel to Berkeley for a week-long workshop run
by the \href{http://rationality.org/}{Center for Applied Rationality}.
It was immensely gratifying to be selected to such an elite gathering
of young AI students and researchers, and hugely fun spending a week
hanging out with smart people thinking about rationality. I will never
forget the experience.
\end{itemize}
Finally, there are three people to whom I am especially indebted:
\begin{itemize}
\item Lulu, my partner of five years: I owe you so much gratitude for your
unconditional love and support. Being my partner, you often have to
experience the worst of me. Thank you for putting up with it, for
being there when I needed you most, and for showing me the right path.
I love you so much. 
\item And of course, my parents, \href{http://www.jennystewart.net.au/}{Jenny}
and \href{http://grapevine.com.au/~timoshenko/}{Timoshenko}: thank
you for your never-ending love and support, and for being so forbearing
and understanding. Seeing you every second Sunday has been a balm.
I love you, and I miss you.

I used to always groan when I was told this, but it's finally starting
to ring true: the older you get, the wiser your parents become.
\end{itemize}

\chapter*{Abstract}

\addcontentsline{toc}{chapter}{Abstract}

Reinforcement learning (RL; \citealp{SB:1998,BT:1995}) is a general
and powerful framework with which to study and implement artificial
intelligence (AI; \citealp{RN:2010}). Recent advances in deep learning
\citep{Schmidhuber:2015deep} have enabled RL algorithms to achieve
impressive performance in restricted domains such as playing Atari
video games \citep{MKSRV+:2015deepQ} and, recently, the board game
Go \citep{DeepMind:2016Go}. However, we are still far from constructing
a\emph{ generally} intelligent agent. Many of the obstacles and open
questions are conceptual: What does it mean to be intelligent? How
does one explore and learn optimally in general, unknown environments?
What, in fact, does it mean to be optimal in the general sense?

The universal Bayesian agent AIXI \citep{Hutter:2000,Hutter:2003AIXI,Hutter:2005}
is a model of a maximally intelligent agent, and plays a central role
in the sub-field of \emph{general }reinforcement learning (GRL). Recently,
AIXI has been shown to be flawed in important ways; it doesn't explore
enough to be asymptotically optimal \citep{Orseau:2010}, and it can
perform poorly with certain priors \citep{LH:2015priors}. Several
variants of AIXI have been proposed to attempt to address these shortfalls:
among them are entropy-seeking agents \citep{Orseau:2011ksa}, knowledge-seeking
agents \citep{OLH:2013ksa}, Bayes with bursts of exploration \citep{Lattimore:2013},
MDL agents \citep{Leike:2016}, Thompson sampling \citep{LLOH:2016Thompson},
and optimism \citep{SH:2015opt}.

We present \href{http://aslanides.github.io/aixijs}{AIXIjs}, a JavaScript
implementation of these GRL agents. This implementation is accompanied
by a framework for running experiments against various environments,
similar to OpenAI Gym \citep{Brockman2016}, and a suite of interactive
demos that explore different properties of the agents, similar to
REINFORCEjs \citep{Karpathy2015}. We use AIXIjs to present numerous
experiments illustrating fundamental properties of, and differences
between, these agents. As far we are aware, these are the first experiments
comparing the behavior of GRL agents in non-trivial settings. 

Our aim is for this software and accompanying documentation to serve
several purposes:
\begin{enumerate}
\item to help introduce newcomers to the field of general reinforcement
learning,
\item to provide researchers with the means to demonstrate new theoretical
results relating to universal AI at conferences and workshops, 
\item to serve as a platform with which to run empirical studies on AIXI
variants in small environments, and
\item to serve as an open-source reference implementation of these agents.
\end{enumerate}
\textbf{Keywords}: Reinforcement learning, AIXI, Knowledge-seeking
agents, Thompson sampling.

\tableofcontents{}

\begin{center}
\addcontentsline{toc}{chapter}{Contents}
\par\end{center}

\listoffigures

\begin{center}
\addcontentsline{toc}{chapter}{List of Figures}
\par\end{center}

\listof{algorithm}{List of Algorithms}

\clearpage{}

\begin{center}
\addcontentsline{toc}{chapter}{List of Algorithms}
\par\end{center}

\pagenumbering{arabic}\setcounter{page}{1}

\chapter{Introduction\label{chap:Introduction}}
\begin{quote}
\emph{Who could have imagined, ever so long ago, what minds would
someday do?}\footnote{This, and all subsequent chapter quotes, are taken from \emph{Rationality:
From AI to Zombies }\citep{Yudkowsky:2015}.}
\end{quote}
What a time to be alive! The field of\emph{ artificial intelligence}
(AI) seems to be coming of age, with many predicting that the field
is set to revolutionize science and industry \citep{holdren2016},
and some predicting that it may soon usher in a posthuman civilization
\citep{Vinge:1993,Kurzweil:2005,Bostrom:2014}. While the field has
notoriously over-promised and under-delivered in the past \citep{Moravec:1988,Miller:2009},
there now seems to be a growing body of evidence in favor of optimism.
Algorithms and ideas that have been developed over the past thirty
years or so are being applied with significant success in numerous
domains; natural language processing, image recognition, medical diagonosis,
robotics, and many more \citep{RN:2010}. This recent success can
be largely attributed to the availability of large datasets, cheaper
computing power\footnote{In particular, and of particular relevance to machine learning with
neural networks, the hardware acceleration due to Graphical Processing
Units (GPUs).}, and the development of open-source scientific software\footnote{For example, ${\tt scikit-learn}$ \citep{scikit-learn}, ${\tt Theano}$
\citep{theano2016}, ${\tt Caffe}$ \citep{jia2014caffe}, and ${\tt TensorFlow}$
\citep{tensorflow2015-whitepaper}, along with many others.}. As a result, the gradient of scientific and engineering progress
in these fields is very steep, and seemingly steepening every year.
The past half-decade in particular has seen an acceleration in funding
and interest, primarily driven by advances in the field of\emph{ statistical
machine learning} (SML; \citealp{Bishop:2006,HTF:2009}), and in particular,
the growing sub-field of\emph{ deep learning} with neural networks
\citep{Schmidhuber:2015deep,LBH:2015deep}.

\section*{Machine learning}

Machine learning (ML) can be thought of as a process of automated
hypothesis generation and testing. ML is typically framed in terms
of \emph{passive} tasks such as regression, classification, prediction,
and clustering. In the most common \emph{supervised learning} setup,
a system observes data sampled i.i.d. from some generative process
$\rho\left(x,y\right)$, where $x$ is some object, for example an
image, audio signal, or document, and $y$ is (in the context of \emph{classification})
a \emph{label}. A typical machine learning \emph{task} is to correctly
predict $y$, given a (in general, previously unseen) datum $x$ sampled
from $\rho\left(x\right)$. This often involves constructing a model
$p\left(y\lvert x,\theta\right)$ parametrized by $\theta$. The system
is said to \emph{learn }from data by tuning the model parameters $\theta$
so as to minimize the\emph{ risk, }which is the $\rho$-expectation
of\emph{ }some loss function $\mathcal{L}$ 
\begin{equation}
\mathbb{E}_{\rho}\left[\mathcal{L}\left(x,y,y'\right)\right],\label{eq:loss-function}
\end{equation}

where $\mathcal{L}$ is constructed in such a way as to penalize prediction
error \citet{Bishop:2006,HTF:2009,Murphy:2012}. High profile breakthroughs
of statistical machine learning include image recognition \citep{SzegedyVISW15},
voice recognition \citep{SakSRB15}, synthesis \citep{wavenet2016},
and machine translation \citep{theano2016}. Many more examples exist,
in diverse fields such as fraud detection \citep{PLMG2010} and bioinformatics
\citep{LNS2015genomics}. Informally, these systems might be called
`intelligent',\emph{ }insofar as they learn an accurate model of (some
part of) the world that generalizes well to unseen data. We refer
to these learning systems as \emph{narrow} AI; they are narrow in
two senses:
\begin{enumerate}
\item They are typically applicable only within a narrow domain; a neural
network trained to recognize cats cannot play chess or reason about
climate data.
\item They are able to solve only \emph{passive }tasks, or active tasks
in a restricted setting.
\end{enumerate}
In contrast, the goal of general artificial intelligence can be described
(informally) as designing and implementing an \emph{agent }that learns
from, and interacts with, its environment, and eventually learns to
(vastly) outperform humans in any given task \citep{Legg:2008,MB:2016forcast}.

\section*{Artificial intelligence}

Constructing an artificial \emph{general} intelligence (AGI) has been
one of the central goals of computer science, since the beginnings
of the discipline \citep{Dartmouth:1955}. The field of \emph{hard},
or \emph{general} AI has infamously had a history of overpromising
and under-delivering, virtually since its birth \citep{Moor:2006Dartmouth,Miller:2009}.
Despite this, the recent success of machine learning has inspired
a new generation of researchers to approach the problem, and there
is a considerable amount of investment being made in the field, most
notably by large technology companies: \href{https://research.facebook.com/ai/}{Facebook AI Research},
\href{http://research.google.com/teams/brain/}{Google Brain}, \href{https://openai.com}{OpenAI}
and \href{https://deepmind.com}{DeepMind} are some high profile examples;
the latter has made the scope of its ambitions explicit by stating
that its goal is to `solve intelligence'. 

The framework of choice for most researchers working in pursuit of
AGI is called \emph{reinforcement learning }(RL; \citealp{SB:1998}).
The current state-of-the-art algorithms combine the relatively simple
\emph{Q-learning }\citep{WD:1992Qlearning} with deep convolutional
neural networks to form the so-called \emph{deep Q-networks }(DQN)
algorithm \citep{MKSGAWR:2013DQN} to learn effective policies on
large state-space Markov decision processes. This combination has
seen significant success at autonomously learning to play games, which
are widely considered to be a rich testing ground for developing and
testing AI algorithms. 

Some recent successes using systems based on this technique include
achieving human-level performance at numerous Atari-2600 video games
\citep{MKSRV+:2015deepQ}, super-human performance at the board game
Go \citep{DeepMind:2016Go,AlphaGo}, and super-human performance at
the first-person shooter \textsc{Doom }\citep{LampleChaplot2016}.
This has inspired a whole sub-field called \emph{deep} reinforcement
learning, which is moving quickly and generating many publications
and software implementations.

While this is all very impressive, these are primarily \emph{engineering
}successes, rather than scientific ones. The fundamental ideas and
algorithms used in DQN date from the early nineties; Q-learning is
due to \citet{WD:1992Qlearning}, and convolutional neural networks
and deep learning are usually attributed to \citet{LB:95convnet}.
Arguably, the scientific breakthroughs necessary for AGI are yet to
be made, and are still some way off. In fact, when one considers the
problem of learning and acting in general environments, there are
still many open foundational problems \citep{Hutter:2009open}: What
is a good formal definition of intelligent or rational behavior? What
is a good notion of optimality with which to compare different algorithms?
These are conceptual and theoretical questions which must be addressed
by any useful theory of AGI. 

\section*{General reinforcement learning}

One proposed answer to the first of these questions is the famous
AIXI model, which is a parameter-free (up to a choice of prior) and
general model of unbounded rationality in unknown environments \citep{Hutter:2000,Hutter:2002,Hutter:2005}.
AIXI is formulated as a Bayesian reinforcement learner, and makes
few assumptions about the nature of its environment; notably, when
studying AIXI we lift the ubiquitous Markov assumption on which algorithms
like Q-learning depend for convergence \citep{SB:1998}. Because of
this important distinction, we refer to AIXI as a \emph{general reinforcement
learning}\footnote{Elsewhere in the literature \textendash{} most prominently by \citet{Hutter:2005}
and \citet{Orseau:2011ksa} \textendash{} the term \emph{universal
AI} is used.}\emph{ }(GRL) agent \citep{LHS:2013grl}. 

Recently, there have been a number of key negative results proven
about AIXI; namely that it isn't \emph{asymptotically optimal} \citep{Orseau:2010,Orseau:2013}
\textendash{} a concept we will formally introduce in \chapref{Background}
\textendash{} and it can be made to perform poorly with certain priors
\citep{LH:2015priors}. These results have motivated, in part, the
development of alternative GRL agents: entropy-seeking agents \citep{Orseau:2011ksa},
optimistic AIXI \citet{SH:2012optimistic}, knowledge-seeking agents
\citep{OLH:2013ksa}, minimum description length agents \citep{Lattimore:2013},
Bayes with exploration \citep{Lattimore:2013,LH:2014pacbayes}, and
Thompson sampling \citep{LLOH:2016Thompson}. 

Numerous results (positive and negative) have been proven about this
family of universal Bayesian agents; together they form a corpus that
is of considerable significance to the AGI problem. With the exception
of AIXI, many of these agents (and their associated properties) are
relatively obscure. We argue that as AI research continues, the theoretical
underpinnings of GRL will rise in importance, and these ideas and
models will serve as useful guiding principles for practical algorithms.
This motivates us to create an open-source web demo of AIXI and its
variants, to help in the presentation of these agents to the AI community
generally, and to serve as a platform for experimentation and demonstration
of deep results relating to rationality and intelligence.

\section*{Web demos}

\begin{figure}
\begin{centering}
\includegraphics[width=5cm,height=5cm]{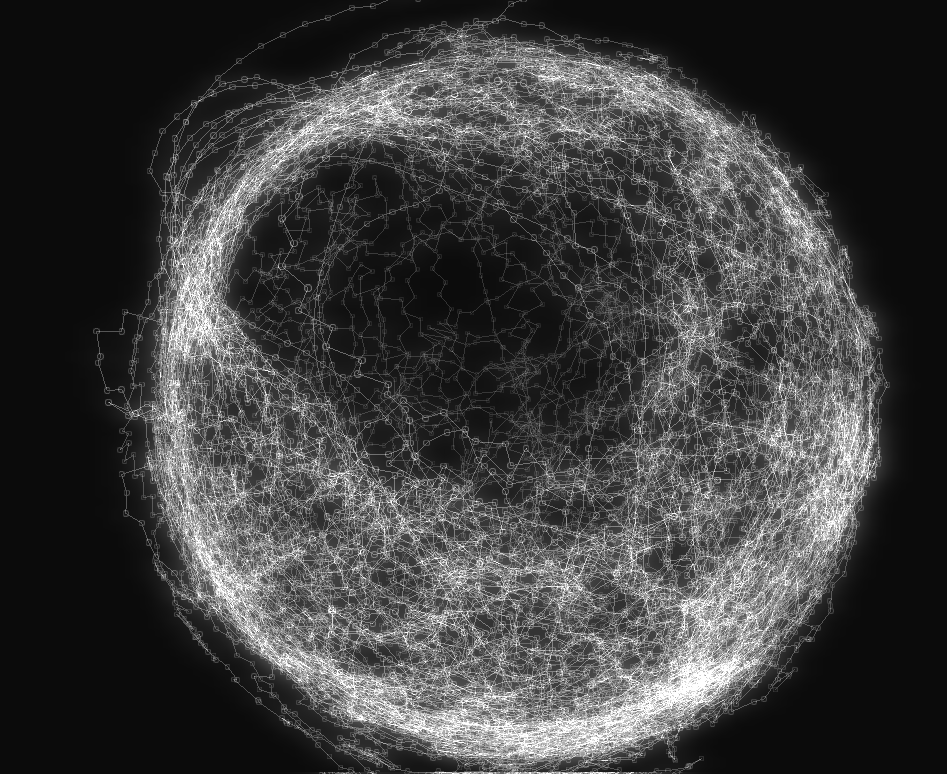}$\quad$\includegraphics[width=5cm,height=5cm]{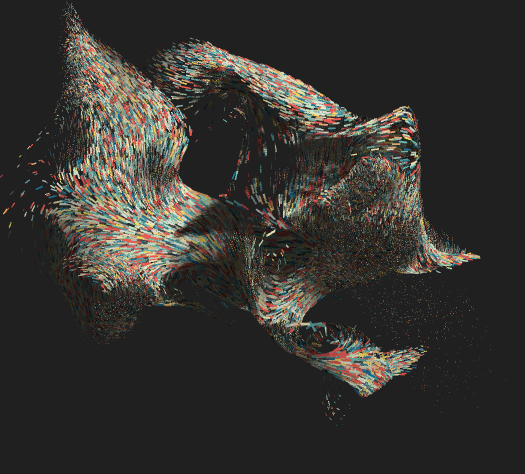}
\par\end{centering}
\caption{Open-source examples of real-time GPU-accelerated particle simulations
run natively in Google Chrome, using JavaScript and WebGL. \textbf{Left}:
\protect\href{https://github.com/jpweeks/particulate-js}{ParticulateJS}.
\textbf{Right}: \protect\href{https://github.com/spite/polygon-shredder}{Polygon Shredder}. }

\centering{}\label{fig:webgl}
\end{figure}

\begin{figure}
\begin{centering}
\includegraphics[width=5cm,height=5cm]{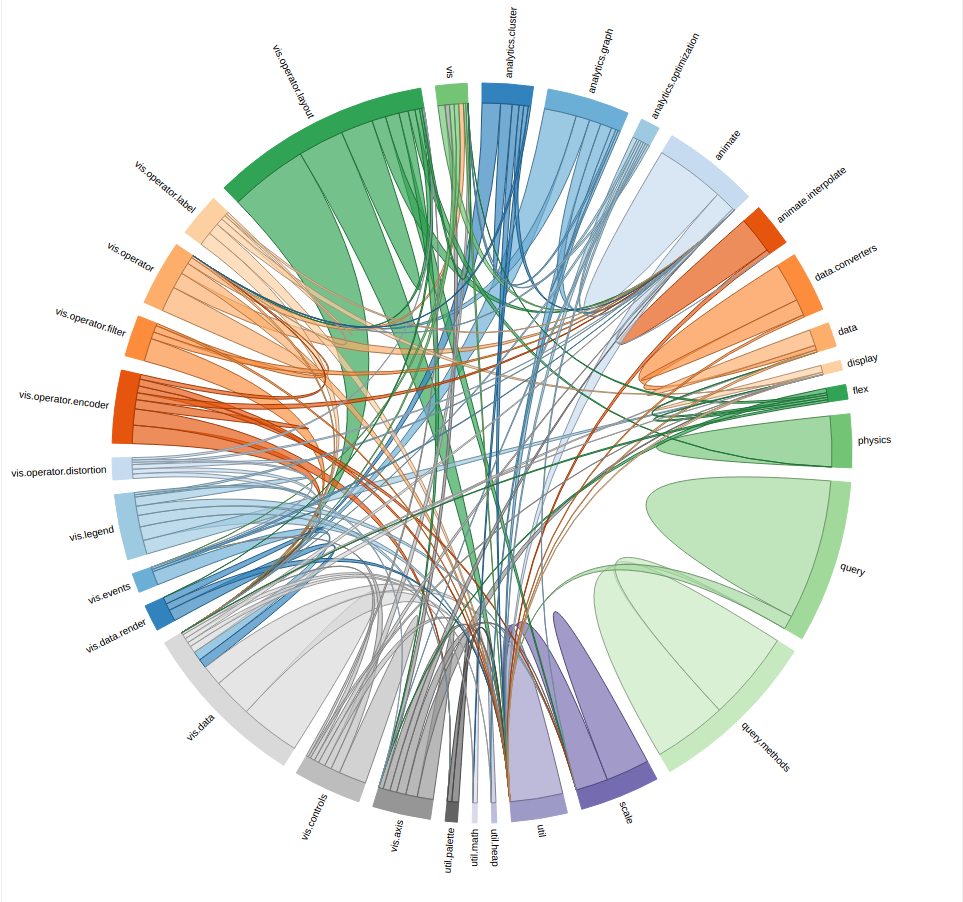}$\quad$\includegraphics[width=5cm,height=5cm]{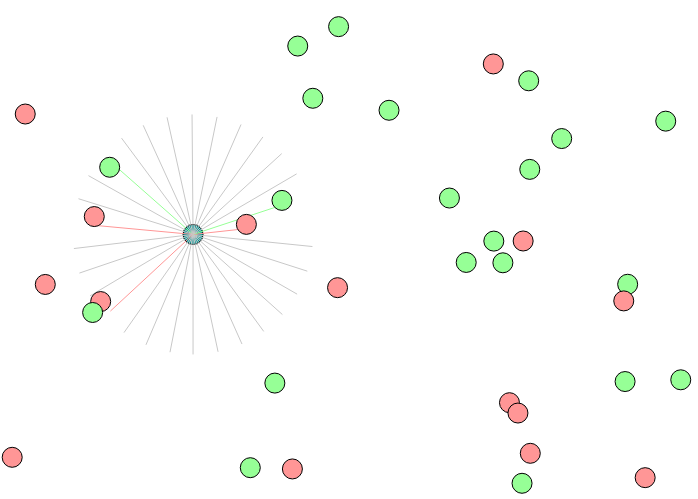}
\par\end{centering}
\caption{Open-source examples of visualizations made with \protect\href{http://d3js.org}{d3js}.
\textbf{Left}: Chord plot for data visualization \citep{Bostock2016}.
\textbf{Right}: Visualization of the $\textsc{WaterWorld}$ reinforcement
learning environment \citep{Karpathy2015}.}

\centering{}\label{fig:d3}
\end{figure}

With increasing computing power, and the maturation of the JavaScript
programming language, web browsers have become a feasible platform
on which to run increasingly complex and computationally intensive
software. JavaScript, in its modern incarnations, is stable, portable,
expressive, and, with engines like \href{https://www.chromeexperiments.com/webgl}{WebGL}
and \href{https://developers.google.com/v8/}{V8}, highly performant;
see \figref{webgl} and \figref{d3} for examples. Thanks to this,
and the popular \href{http://d3js.org}{d3js} visualization library,
there are now a growing number of excellent open source machine learning
web demos available online. Representative examples include \href{https://transcranial.github.io/keras-js}{Keras-js},
a demo of very large convolutional neural networks \citep{Chen2016};
\href{http://playground.tensorflow.org/}{TensorFlow Playground},
a highly interactive demo designed to give intuition for how neural
networks classify data \citep{SC2016tfdemo}; a demo to illustrate
the pitfalls and common misunderstandings when using the \href{http://distill.pub/2016/misread-tsne/}{t-SNE}
dimensionality reduction technique \citep{wattenberg2016how}, and
Andrej Karpathy's excellent reinforcement learning demo \href{http://cs.stanford.edu/people/karpathy/reinforcejs/}{REINFORCEjs},
that demonstrates the DQN algorithm \citep{Karpathy2015}. 

Arguably, these demos have immense value to the community, as they
serve at once as reviews of recent research, pedagogic aides, and
as accessible reference implementations for developers. They are also
effective marketing for the techniques or approaches being demonstrated,
and the people producing them. We now describe the objectives of this
project.

\section*{Objective}

This thesis is about understanding existing theoretical results relating
to GRL agents, implementing these agents, and communicating these
properties via an interactive software demo. In particular, the demo
should:
\begin{itemize}
\item be \emph{portable}, i.e. runnable on any computer with a modern web
browser and internet connection,
\item be \emph{general} and \emph{extensible}, so as to support a wide range
of agents and environments,
\item be \emph{modular}, so as to facillitate future development and improvement,
and
\item be \emph{performant}, so that users can run non-trivial simulations
in a reasonable amount of time.
\end{itemize}
The demo will consist of: 
\begin{itemize}
\item implementations of the agents and their associated modules (planners,
environment models),
\item a suite of small environments on which to demonstrate properties of
the agents,
\item a user interface (UI) that provides the user control over agent and
environment parameters,
\item a visualization interface that allows the user to playback the agent-environment
simulation, and
\item a suite of explanations, one accompanying each demo, to explain what
the user is seeing.
\end{itemize}
In particular, the demo should serve three purposes:
\begin{itemize}
\item as a helpful introduction to the theory of general reinforcement learning,
for both students and researchers; in this regard, we follow the model
of \href{http://cs.stanford.edu/people/karpathy/reinforcejs/}{REINFORCEjs}
\citep{Karpathy2015};
\item as a platform for researchers in this area to develop and run experiments
to accompany their theoretical results, and to help present their
findings to the community; in this aspect, we follow the model of
\href{https://gym.openai.com}{OpenAI Gym} \citep{Brockman2016};
\item and as an open-source reference implementation for many of the general
reinforcement learning agents.
\end{itemize}

\section*{Contribution}

In this work, we present:
\begin{itemize}
\item a review of the general reinforcement learning literature of Hutter,
Lattimore, Sunehag, Orseau, Legg, Leike, Ring, Everitt, and others.
We present the agents and results under a unified notation and with
added conceptual clarifications. As far as we are aware, this is the
only document in which agents and algorithms from the GRL literature
are presented as a collection.
\item an applied perspective on Bayesian agents and mixture models with
insights into MCTS planning and modelling errors,
\item an open-source JavaScript reference implementation of many of the
agents,
\item experimental data that validates and illustrates several theoretical
results, and
\item an extensible and general framework with which researchers can run
experiments and demos on reinforcement learning agents in the browser.
\end{itemize}
The software itself is found at \href{http://aslanides.github.io/aixijs}{http://aslanides.github.io/aixijs},
and can be run in the browser on any operating system. Note that different
browsers have differing implementations of the JavaScript specification;
we strongly recommend running the demo on \href{https://www.google.com.au/chrome/browser/desktop/}{Google Chrome}\footnote{\href{https://www.google.com.au/chrome/browser/desktop/}{https://www.google.com.au/chrome/browser/desktop/}},
as we didn't test the implementation on other browsers.

\section*{Thesis Outline}

In \chapref{Background} (\nameref{chap:Background}) we present the
theoretical framework for general reinforcement learning, introduce
the agent zoo, and present the basic optimality results.  In \chapref{AIXIjs}
(\nameref{chap:AIXIjs}) we document the design and implementation
of the software itself. In \chapref{experiments} (\nameref{chap:experiments})
we outline the experimental results we obtained using the software.
\chapref{Conclusion} (\nameref{chap:Conclusion}) makes some concluding
remarks, and points out potential directions for further work. 

We expect that this thesis will typically be read in \emph{soft }copy,
i.e. digitally, through a $\textsc{PDF}$ viewer. For this reason,
we augment this thesis throughout with hyperlinks, for the reader's
convenience. These are used in three ways:
\begin{itemize}
\item on citations, so as to link to the corresponding bibliography entry,
\item on cross-references, so as to link to the appropriate page in this
thesis,\footnote{After following a link, the reader can return to where they were previously,
using (usually) \setmenukeyswin\keys{\Alt + \arrowkeyleft} on Windows
or Linux, and \setmenukeysmac\keys{\cmd + [} (in \emph{Preview})
or \keys{\cmd + \arrowkeyleft} (in \emph{Acrobat}) on Mac OS.} or 
\item as external hyperlinks, to link the interested reader to an internet
web page.
\end{itemize}
In particular, we encourage the reader to use the cross-references
to jump around the text.

\global\long\def\deff{\stackrel{.}{=}}

\global\long\def\d{\text{{d}}}

\global\long\def\ts#1{\textsc{#1}}

\chapter{Background\label{chap:Background}}
\begin{quote}
\emph{“Where will an Artificial Intelligence get money?” they ask,
as if the first Homo sapiens had found dollar bills fluttering down
from the sky, and used them at convenience stores already in the forest.}
\end{quote}
In this Chapter we present a brief background on reinforcement learning,
with a focus on the problem of \emph{general} reinforcement learning
(GRL). Our objective is for this chapter to be relatively accessible.
To this end, we try to aim for conceptual clarity and conciseness
over technical details and mathematical rigor. For a more complete
and rigorous treatment of GRL, we refer the reader to the excellent
PhD theses of \citet{Leike:2016} and \citet{Lattimore:2013}, and
of course to the seminal book, \emph{Universal Artificial Intelligence}
by \citet{Hutter:2005}.

The Chapter is laid out as follows: In \secref{Preliminaries} (\nameref{sec:Preliminaries}),
we introduce some notation and basic concepts. In \secref{Reinforcement-Learning}
(\nameref{sec:Reinforcement-Learning}), we introduce the reinforcement
learning problem in its most general setting. In \secref{Bayesian-Reinforcement-Learning}
(\nameref{sec:Bayesian-Reinforcement-Learning}) we introduce the
Bayesian general reinforcement learner AIXI and its relatives, the
implementation and experimental study of which forms the bulk of this
thesis. We draw the GRL literature together and present these agents
under a unified notation. In \secref{planning} (\nameref{sec:planning})
we discuss approaches to the problem of planning in general environments.\emph{
}We conclude with some remarks and a short summary in \secref{Remarks}
(\nameref{sec:Remarks}). 

\section{Preliminaries\label{sec:Preliminaries}}

We briefly introduce some of the tools and concepts that are used
to reason about the general reinforcement learning (GRL) problem.
We assume that the reader has a basic familiarity with the concepts
of probability, information theory, and statistics, and ideally some
exposure to standard concepts in artificial intelligence (e.g. breadth-first
search, expectimax, minimax), and reinforcement learning (e.g. Q-learning,
bandits). For some general background, we refer the reader to \citet{MacKay:2002}
for probability and information theory, \citet{Bishop:2006} for machine
learning and statistics, \citet{RN:2010} for artificial intelligence,
and \citet{SB:1998} for reinforcement learning.

\subsection{Notation}

\textbf{Numbers and vectors}. The set $\mathbb{N}\deff\left\{ 1,2,3,\dots\right\} $
is the set of natural numbers, and $\mathbb{R}$ denotes the reals.
We use $\mathbb{R}_{+}=[0,\infty)$ and $\mathbb{R}_{++}=\left(0,\infty\right)$.
A set is\emph{ countable} if it can be brought into bijection with
a subset (finite or otherwise) of $\mathbb{N}$, and is uncountable
otherwise. We use $\mathbb{R}^{K}$ to denote the $K$-dimensional
vector space over $\mathbb{R}$. We represent vectors with bold face:
$\boldsymbol{x}$ is a vector, and $x_{i}$ is its $i^{\text{th}}$
component. We (reluctantly\footnote{The author greatly favors using the \href{https://en.wikipedia.org/wiki/Einstein_notation}{Einstein notation}
for its power and clarity.}) represent inner products with the standard notation for engineering
and computer science: given $\boldsymbol{x},\boldsymbol{y}\in\mathbb{R}^{K}$,
$\boldsymbol{x}^{T}\boldsymbol{y}=\sum_{i=1}^{K}x_{i}y_{i}$.

\textbf{Strings and sequences}. Define a finite, nonempty set of symbols
$\mathcal{X}$, which we call an\emph{ alphabet}. The set $\mathcal{X}^{n}$
with $n\in\mathbb{N}$ is the set of all strings over $\mathcal{X}$
with length $n$, and $\mathcal{X}^{*}=\cup_{n\in\mathbb{N}}\mathcal{X}^{n}$
is the set of all finite strings over $\mathcal{X}$. $\mathcal{X}^{\infty}$
is the set of infinite strings over $\mathcal{X}$, and $\mathcal{X}^{\#}=\mathcal{X}^{*}\cup\mathcal{X}^{\infty}$
is their union. The empty string is denoted by $\epsilon$; this is
not to be confused with the small positive number $\varepsilon$.
For any string $x\in\mathcal{X}^{\#}$, we denote its length by $\left|x\right|$.

For any string $x$ with $\left|x\right|\geq k$, $x_{k}$ is the
$k^{\text{th}}$ symbol of $x$, $x_{1:k}$ is the first $k$ symbols
of $x$, and $x_{<k}$ is the first $k-1$ symbols of $x$. We often
make use of the binary alphabet $\mathbb{B}=\left\{ 0,1\right\} $.
For two finite strings $x,y\in\mathcal{X}^{*}$ we denote their concatenation
by $xy$. For two finite strings $a,e\in\mathcal{X}^{n}$ of length
$n$, it will be convenient to write $\ae$ to indicate the riffled
string $a_{1}e_{1}a_{2}e_{2}\dots a_{n}e_{n}$; we slightly overload
our indexing notation by stipulating that for $k\leq n$, $\ae_{1:k}=a_{1}e_{1},\dots,a_{k}e_{k}$,
and similarly for $\ae_{<k}$.

\textbf{Miscellaneous}. We use $\stackrel{.}{=}$ to mean `is defined
as', and we use the convention that $\log$ is the logarithm base
two and $\ln$ is the natural logarithm. We usually, but not always,
refer to random variables in upper case. The indicator function $\mathbb{I}\left[P\right]$
returns $1$ if the predicate $P$ is true and $0$ otherwise. We
use $\to$ and $\rightsquigarrow$ to denote deterministic and stochastic
mappings respectively.

\subsection{Probability theory\label{subsec:Probability-theory}}

For our purposes, we will only be working with discrete event spaces,
and so we will omit the machinery of measure theory, which is needed
to treat probability theory over continuous spaces. Given a \emph{sample
space }$\Omega$, we construct an \emph{event space} $\mathcal{F}$
as a $\sigma$-algebra on $\Omega$: a set of subsets of $\Omega$
that is closed under countable unions and complements; for discrete
distributions, this is simply the power set $2^{\Omega}$. A \emph{random
variable }$X$ is\emph{ discrete} if its associated sample space $\Omega_{X}$
is countable; we associate with it a probability mass function $p\ :\ \Omega_{X}\to\left[0,1\right]$.
If $X$ is continuous, provided $\Omega_{X}$ is measurable, we can
associate with it a probability density function $\mathbb{R}\to\mathbb{R}_{+}$.
For a countable set $\Omega$, we use $\Delta\Omega$ to represent
the set of all probability distributions over $\Omega$. We use $\mathbb{E}\left[X\right]\stackrel{.}{=}\sum_{x\in\Omega_{X}}xp\left(x\right)$
(or, in the continuous setting, $\mathbb{E}\left[X\right]=\int_{X}xp\left(x\right)\d x$)
to represent the expectation of the random variable $X$. In many
cases we will emphasize for clarity that $X$ is distributed according
to $p$ by writing the expectation as $\mathbb{E}_{p}\left[X\right]$.
We say $x\sim\rho\left(\cdot\right)$ to mean that $x$ is sampled
from the distribution $\rho$. 

The two fundamental results of probability theory are the sum and
product rules:
\begin{eqnarray}
p\left(a\right) & = & \sum_{b\in\Omega_{B}}p\left(a,b\right)\label{eq:sum}\\
p\left(a,b\right) & = & p\left(a\lvert b\right)p\left(b\right),\label{eq:product}
\end{eqnarray}

from which we immediately get Bayes' rule, which plays a central role
in the theory of rationality and intelligence \citep{Hutter:2000}.
\begin{thm}[\emph{Bayes' rule}]
\label{thm:bayes}Bayes' rule is given by the following identity:
\begin{eqnarray}
\overbrace{\text{Pr}\left(A\lvert B\right)}^{\text{posterior}} & = & \frac{\overbrace{\text{Pr}\left(B\lvert A\right)}^{\text{likelihood}}\overbrace{\text{Pr}\left(A\right)}^{\text{prior}}}{\underbrace{\text{Pr}\left(B\right)}_{\text{predictive distribution}}}\label{eq:Bayes}\\
 & = & \frac{\text{Pr}\left(B\lvert A\right)\text{Pr}\left(A\right)}{\sum_{a\in\Omega_{A}}\text{Pr}\left(B\lvert a\right)\text{Pr}\left(a\right)}.\nonumber 
\end{eqnarray}
\end{thm}
Note that Bayes' rule follows from the fact that the product rule
is symmetric in its arguments: $p\left(a\lvert b\right)p\left(b\right)=p\left(b\lvert a\right)p\left(a\right)$.
Its power and significance comes through its interpretation as a sequential
updating scheme for subjective \emph{beliefs} about hypotheses; we
annotate \eqref{Bayes} with this interpretation, which we discuss
below. The most distinguishing feature of being Bayesian is of interpreting
your probabilites\emph{ subjectively}, in the sense that they represent
your credence in some outcome, or some model. Updating beliefs using
Bayes' rule is a (conceptually) trivial step, since it just says that
your beliefs are constrained by the rules of probability theory; if
they weren't, you would be vulnerable to Dutch book arguments \citep{Jaynes:2003}.

In our context, typically $A$ is some \emph{model} or \emph{hypothesis},
and $B$ is some \emph{observation}. $\Pr\left(A\right)$ is our \emph{prior
}belief in the correctness of hypothesis $A$, and $\Pr\left(A\lvert B\right)$
is our \emph{posterior }belief\emph{ }in\emph{ $A$} after taking
in some observation, \emph{$B$. }Effectively, Bayes' rule defines
the mechanism with which we move probability mass between competing
hypotheses. Note that once we assign zero probability (or \emph{credence})
to some hypothesis $A$, then there is no observation $B$ that will
change our mind about the impossibility of $A$. This is not such
a problem if it so happens that $A$ is false; the situation in which
$A$ is true, and has been prematurely (and incorrectly) falsified,
is sometimes known informally as \emph{Bayes Hell}. For this reason,
we try to avoid using priors that assign zero probability to events;
this is known more formally as Cromwell's rule. 

Notice that in general the sample spaces $\Omega_{A}$ and $\Omega_{B}$
are different; $B$ is a random variable on some set of possible observations,
$\Omega_{B}$, while $A$ is a random variable over a set of \emph{hypotheses},
which aren't observed, but constructed. To emphasize this distinction,
we use a separate notation: $\mathcal{M}$ represents a set (or, in
the uncountable case, \emph{space}) of hypotheses, which we will call
a \emph{model class}. We implicitly assume that in all cases $\mathcal{M}$
contains at least two elements. Sequential Bayesian updating in this
way is an \emph{inductive} process; we refine our models based on
observation. As we will see in \secref{Bayesian-Reinforcement-Learning},
the predictive distribution $\Pr\left(B\right)$ will play an important
role for our reinforcement learning agents. 

We formalize Cromwell's rule with the concept of a \emph{universal
prior}.
\begin{defn}[Universal prior]
A\emph{ prior} over a countable class of objects $\mathcal{M}$ is
a probability mass function $p\in\Delta\mathcal{M}$, such that $p\left(\nu\right)$
is defined for each $\nu\in\mathcal{M}$, with $p\left(\nu\right)\in\left[0,1\right]$
and $\sum_{\nu\in\mathcal{M}}p\left(\nu\right)=1$. A\emph{ universal
prior} assigns non-zero mass to every hypothesis such that $p\left(\nu\right)\in\left(0,1\right)$
for all $\nu\in\mathcal{M}$.
\end{defn}
We often make use of the following distributions:

\textbf{Bernoulli}. We use $\text{Bern}\left(\theta\right)$ to represent
the Bernoulli process on $x\in\left\{ 0,1\right\} $, with probability
mass function given by $p\left(x\lvert\theta\right)=\theta^{x}\left(1-\theta\right)^{x}$.

\textbf{Binomial}. We use $\text{Binom}\left(n,p\right)$ to represent
the Binomial distribution on $k\in\left\{ 0,\dots,n\right\} $ with
mass function given by $p\left(k\lvert n,p\right)=\binom{n}{k}p^{k}\left(1-p\right)^{k}$,
where $\binom{n}{k}=\frac{n!}{k!\left(n-k\right)!}$ is the known
as the \emph{binomial coefficient}. 

\textbf{Uniform}. We use $\mathcal{U}\left(a,b\right)$ to represent
the measure that assigns uniform density to the closed interval $\left[a,b\right]$,
with $b>a$; its density is given by $p\left(x\right)=\frac{1}{b-a}\mathbb{I}\left[a\leq x\leq b\right]$.
We overload our notation (and nomenclature) and also use $\mathcal{U}\left(\mathcal{A}\right)$
to represent the uniform distribution over the finite set $\mathcal{A}$;
its mass function is given by $p\left(a\right)=\frac{1}{\left|\mathcal{A}\right|}$.

\textbf{Normal}. We use $\mathcal{N}\left(\mu,\sigma^{2}\right)$
to represent the univariate Gaussian distribution on $\mathbb{R}$
with density given by
\[
p\left(x\lvert\mu,\sigma^{2}\right)=\left(2\pi\sigma^{2}\right)^{-\frac{1}{2}}\exp\left(-\frac{\left(x-\mu\right)^{2}}{2\sigma^{2}}\right).
\]

\textbf{Beta}. We use $\text{Beta}\left(\alpha,\beta\right)$ to represent
the Beta distribution on $\left[0,1\right]$ with density given by
\[
p\left(x\lvert\alpha,\beta\right)=\frac{\Gamma\left(\alpha+\beta\right)}{\Gamma\left(\alpha\right)\Gamma\left(\beta\right)}x^{\alpha-1}\left(1-x\right)^{\beta-1},
\]

where $\Gamma$ is the Gamma function that interpolates the factorials. 

The beta distribution is conjugate to the Bernoulli and Binomial distributions;
this means that a Bayesian updating scheme can use a Beta distribution
as a prior $p\left(\theta\right)$ over the parameter of some Bernoulli
process, whose likelihood is given by $p\left(x\lvert\theta\right)$.
Since the Beta and Bernoulli are conjugate, the posterior $p\left(\theta\lvert x\right)$
will also take the form of a Beta distribution. Conjugate pairs of
distributions such as this allow us to analytically compute the posterior
resulting from a Bayesian update, and are essential for tractable
Bayesian learning.

\textbf{Dirichlet}. We use $\text{Dirichlet}\left(\alpha_{1},\dots,\alpha_{K}\right)$
to represent the Dirichlet distribution on the 1-simplex
\[
S_{K}\stackrel{.}{=}\left\{ \boldsymbol{x}\in\mathbb{R}_{+}^{K}\ \left|\ \boldsymbol{1}^{T}\boldsymbol{x}=1\right.\right\} ,
\]

with density given by
\[
p\left(\boldsymbol{x}\lvert\alpha\right)=\frac{\Gamma\left(\sum_{i=1}^{K}\alpha_{i}\right)}{\prod_{i=1}^{K}\Gamma\left(\alpha_{i}\right)}\prod_{i=1}^{K}x_{i}^{\alpha_{i}-1}.
\]

This is the multidimensional generalization of the Beta distribution,
and is conjugate to the Categorical and Multinomial distributions.
The categorical distribution over some discrete set $\mathcal{X}$
is simply a vector on the 1-simplex, $\boldsymbol{p}\in S_{K}$, where
$K=\left|\mathcal{X}\right|$. The multinomial simply generalizes
the binomial distribution. 

As we shall see in \secref{Bayesian-Reinforcement-Learning}, a significant
aspect of intelligence is \emph{sequence prediction}. For this reason,
we introduce measures over sequences. A distribution over finite sequences
$\rho\in\Delta\mathcal{X}^{*}$ can be written as $\rho\left(x_{1:n}\right)$
for some finite $n$. Analogously to the sum and product rules, we
have 
\begin{eqnarray*}
\rho\left(x_{n}\lvert x_{<n}\right) & = & \frac{\rho\left(x_{1:n}\right)}{\rho\left(x_{<n}\right)}\\
\rho\left(x_{<n}\right) & = & \sum_{y\in\mathcal{X}}\rho\left(x_{<n}y\right).
\end{eqnarray*}

There are two important properties that sequences can have which are
relevant to reinforcement learning: the Markov and ergodic properties: 

\textbf{Markov property}. A generative process $\rho$ is $n^{\text{th}}$-order
Markov if it has the property 
\[
\rho\left(x_{t}\lvert x_{<t}\right)=\rho\left(x_{t}\lvert x_{(t-n):(t-1)}\right).
\]

Typically, when we invoke the Markov property, we mean that the process
is $1^{\text{st}}$-order Markov. A \emph{Markov chain} is simply
a first-order Markov process over a finite alphabet $\mathcal{X}$,
in which the conditional distribution is \emph{stationary}, and can
thus be represent as a $\left|\mathcal{X}\right|\times\left|\mathcal{X}\right|$
transition matrix $P\left(x'\lvert x\right)\equiv\rho\left(x'_{t}\lvert x_{t-1}\right)$.
This matrix is said to be \emph{stochastic}, to emphasise that it
represents a distribution over $x'$, so that $P\left(x'\lvert x\right)\in\left[0,1\right]$
and $\sum_{s'}P\left(x'\lvert x\right)=1$. In this context, we often
identify the \emph{symbols }$x\in\mathcal{X}$ with \emph{states. }

\textbf{Ergodicity}. In a Markov chain, a state $i$ is said to be
ergodic if there is non-zero probability of leaving the state, and
the probability of eventually returning is unity. If all states are
ergodic, then the Markov chain is ergodic. Informally, this means
that the Markov chain has no traps: at all times, we can freely move
around the MDP without ever making unrecoverable mistakes. Ergodicity
is an important assumption in the theory of Markov Decision Processes,
which we will see later.

\subsection{Information theory}

For a distribution $p\in\Delta\mathcal{X}$ over a countable set $\mathcal{X}$,
the\emph{ entropy} of\emph{ $p$} is
\begin{equation}
\text{Ent}\left(p\right)\stackrel{.}{=}-\sum_{x\in\mathcal{X}\ :\ p\left(x\right)>0}p\left(x\right)\log p\left(x\right).\label{eq:entropy}
\end{equation}

Absent additional constraints, the maximum entropy distribution is
$\mathcal{U}$, our generalized uniform distribution. We also define
the conditional entropy 
\[
\text{Ent}\left(p\left(\cdot\lvert y\right)\right)\stackrel{.}{=}\sum_{x\in\mathcal{X}\ :\ p\left(x\right)>0}p\left(x\lvert y\right)\log p\left(x\lvert y\right).
\]

Given two distributions $p,q\in\Delta\mathcal{X}$, the\emph{ Kullback-Leibler
divergence} (KL-divergence, also known as relative entropy) is defined
by
\[
\text{KL}\left(p\|q\right)\stackrel{.}{=}\sum_{x\in\mathcal{X}\ :\ p\left(x\right)>0,q\left(x\right)>0}p\left(x\right)\log\frac{p\left(x\right)}{q\left(x\right)}.
\]

We use the $\|$ symbol to separate the arguments so as to emphasise
that the KL-divergence is not symmetric, and hence not a distance
measure. It is non-negative, by Gibbs' inequality. If $p$ and $q$
are measures over\emph{ }sequences, then we can define the conditional
$d$-step KL-divergence 
\[
\text{KL}_{d}\left(p,q\lvert x_{<t}\right)=\sum_{x_{t:t+d}\in\mathcal{X}^{d}}p\left(x_{1:(t+d)}\lvert x_{<t}\right)\log\frac{p\left(x_{1:(t+d)}\lvert x_{<t}\right)}{q\left(x_{1:(t+d)}\lvert x_{<t}\right)}.
\]

\section{Reinforcement Learning\label{sec:Reinforcement-Learning}}

In contrast to machine learning, in the reinforcement learning setting,
the training data that the system receives is now dependent on its
\emph{actions}; we thus introduce \emph{agency} to the learning problem
\citep{SB:1998}. What observations the agent can make, and therefore
what it can learn, now depend not only on the environment (as in machine
learning), but also on the agent's own\emph{ policy}, which determines
how it will behave \citep{Barto2004}. In this way, reinforcement
learning considerably generalizes machine learning; we replace the
loss function of \eqref{loss-function} with a \emph{reward signal}.
Now, instead of minimizing \emph{risk}, the agent must seek to maximize
future expected \emph{rewards}. In this way, reinforcement learning
generalizes machine learning to the active setting, so that the agent
can now influence its environment with the actions that it takes.

We distinguish this from the related set-up known as \emph{inverse}
reinforcement learning or \emph{imitation learning} \citep{AN:2004irl},
in which the agent is given training data consisting of a history
of actions and percepts from which it must infer a policy. In contrast,
reinforcement learners must take their own actions and learn through
trial and error – they are only \emph{supervised} to the extent that
their extrinsic reward signal gives them feedback on their policy.

Because we are motivated by the\emph{ general reinforcement learning
problem}, we introduce a more general and pedantic setup than is common
in the reinforcement learning literature. This set-up has been honed
by (for example) \citet{LH:2011opt} and \citet{LLOH:2016Thompson}. 

\subsection{Agent-environment interaction}

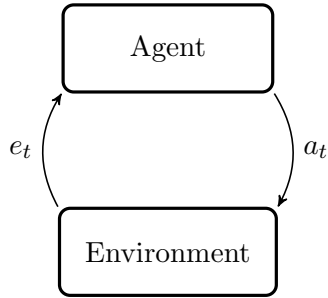
\begin{figure}
\begin{centering}
\tikzset{
    punkt/.style={
           rectangle,
           rounded corners,
           draw=black, very thick,
           text width=6.5em,
           minimum height=3em,
           text centered,
	}
}
\begin{tikzpicture}[auto, ->, >=stealth',node distance=2.8cm,semithick]
\node[punkt] (agent) {Agent};
\node[punkt, inner sep=5pt,below=1.5cm of agent]  (environment) {Environment};
\path[->]
		(agent.south east) edge  [bend left] node {$a_t$} (environment.north east)
		(environment.north west)  edge [bend left] node {$e_t$} (agent.south west);
\end{tikzpicture}
\par\end{centering}
\caption{Cybernetic model of agent-environment interaction.}

\label{fig:cybernetic}
\end{figure}

In the standard \emph{cybernetic model} (see \figref{cybernetic}),
the agent and environment are separate entities that play a turn-based
two-player game. At time $t$, the agent produces an \emph{action}
$a_{t}$, which is passed as an input to the environment, which performs
some computation that (in general) changes its internal state, and
then returns a \emph{percept} $e_{t}$ to the agent. We often refer
to the time $t$ as the number of agent-environment \emph{cycles }that
have elapsed. Together, the agent and environment generate a \emph{history
$\ae_{1:t}=a_{1}e_{1}\dots a_{t}e_{t}$. }In general, it is consequential
to the behavior of the agent whether this interaction runs indefinitely
or finishes after some finite lifetime $T$ \citep{MEH:2016death};
we discuss this to an extent when we introduce discount functions
in \subsecref{discounting}.

\begin{defn}[Environment]
\label{def:environment}An\emph{ environment} is a tuple $\left(\mathcal{A},\mathcal{S},\mathcal{E},D,\nu\right)$,
where
\end{defn}
\begin{itemize}
\item $\mathcal{A}$ is the\emph{ action} space,
\item $\mathcal{S}$ is the\emph{ state} space of the environment, which
is in general hidden from the agent.
\item $\mathcal{E}$ is the\emph{ percept} space, which is itself composed
of observations $o\in\mathcal{O}$ and rewards $r\in\mathcal{R}$
with $\mathcal{E}=\mathcal{O}\times\mathcal{R}$.
\item $D\ :\ \mathcal{S}\times\mathcal{A}\rightsquigarrow\mathcal{S}$ is
the (in general stochastic) dynamics/transition function on the environment's
state space. Note that, without loss of generality, we can allow $\mathcal{D}$
to be first-order Markov. 
\item $\rho\ :\ \mathcal{S}\to\Delta\mathcal{E}$ is the percept function,
by analogy to a hidden Markov model (HMM) in the context of statistical
machine learning\footnote{POMDPs are to MDPs as Hidden Markov Models are to Markov chains.}.
\end{itemize}
Note that for the purposes of General reinforcement learning (GRL),
we make no Markov assumption on the percepts, and we make no ergodicity
assumption on the state or percept spaces.

Since we typically take the agent's perspective, we don't have access
to the environment's state $s\in\mathcal{S}$, nor its dynamics $D$.
For this reason, we typically talk about the environment in terms
of the measure
\[
\nu\ :\ \left(\mathcal{A}\times\mathcal{E}\right)^{*}\times\mathcal{A}\to\Delta\mathcal{E},
\]

which we write
\[
\nu\left(e_{t}\lvert\ae_{<t}a_{t}\right).
\]

Note that the vertical bar $\lvert$ is an abuse of notation here:
$\nu$ is not\emph{ conditioned} on the actions, since it is not derived
from a joint distribution over actions and percepts; the sequence
of actions $a_{1:t}$ are\emph{ inputs} to the environment. A more
pedantic (but ugly) notation would be to write $\nu\left(e_{t}\lvert e_{<t}\|a_{1:t}\right)$
or $\nu\left(e_{t}\lvert e_{<t};\,a_{1:t}\right)$, which emphasizes
that $\nu$ is a conditional distribution with respect to percepts,
but not with respect to actions. We typically refer to the environment
itself with the symbol $\nu$, for convenience.

It is worth pausing to make some remarks about this setup here:
\begin{enumerate}
\item In the general setting, environments are\emph{ partially-observable
Markov decision processes} (POMDPs). We can always model an environment
as Markovian with respect to\emph{ some} hidden state, since if it
depends on some history of states, we incorporate sufficient history
into the state until the Markov property is restored.
\item For our purposes, we assume that $\mathcal{A}$, $\mathcal{E}$, and
$\mathcal{S}$ are all finite.
\item No matter what state the agent is in, it always has the full action
space available to it. This simplifies the setup, and means that when
implementing a simulated environment, we have to specify dynamics
for every action in every state – `illegal' or not. For an example
of this, see \exref{go}.
\item Stochastic environments are sufficiently general to model everything,
including Nature, adversaries, and naturally, deterministic environments.
\item This is an implicitly\emph{ dualistic} model, in the sense that the
agent is separate from the environment; in reality the agent will
be embedded within the environment.
\item As with all simulations run on computers, time is of course discretized.
\item We stipulate that our environments have the \emph{chronological} property,
which simply means that percepts at time $t$ do not depend on future
actions, i.e. $\nu\left(e_{1:t}\|a_{1:\infty}\right)=\nu\left(e_{1:t}\|a_{1:t}\right)$.
\end{enumerate}
The agent-environment interaction is thus modelled as a stochastic,
imperfect-information, two-player game. The environment specifies
both the percept space $\mathcal{E}$ and action space $\mathcal{A}$.
The agent `plugs in' to the environment (which, without loss of generality,
can be thought of as a game simulation) and plays its moves in turns. 

We now present some definitions of common classes of environments.
For a more comprehensive taxonomy, see, for example, \citet{Legg:2008}. 
\begin{defn}[Markov Decision Process]
\label{ex:mdp}A finite-state Markov decision process (MDP) is a
tuple $\left(\mathcal{S},\mathcal{A},\mathcal{P},\mathcal{R}\right)$
where 
\end{defn}
\begin{itemize}
\item $\mathcal{S}$ is a finite state space, labelled by indices $s_{1},\dots,s_{\left|\mathcal{S}\right|}$.
\item $\mathcal{A}$ is a finite action space, labelled by indices $a_{1},\dots,a_{\left|\mathcal{A}\right|}$.
\item $\mathcal{P}$ is the set of transition probabilities $P\left(s'\lvert s,a\right)$,
which can be thought of as a stochastic rank-3 tensor of dimensions
$\left|\mathcal{S}\right|\times\left|\mathcal{S}\right|\times\left|\mathcal{A}\right|$
\item $\mathcal{R}$ is the set of rewards $R\left(s,a\right)$.
\end{itemize}
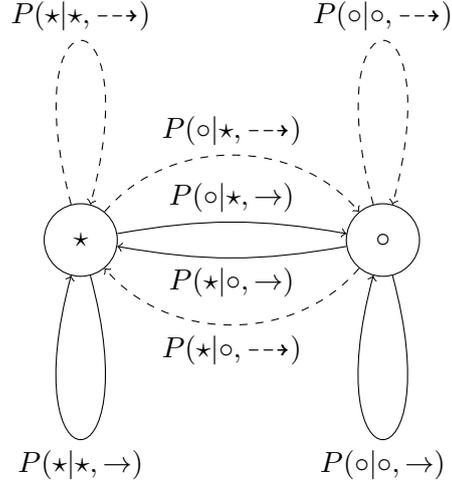
\begin{figure}
\begin{centering}
\tikzset{every loop/.style={min distance=30mm,looseness=10}}
\begin{tikzpicture}
	\node [state] (s1) {$\star$};
	\node [state, right=3cm of s1] (s2) {$\circ$};
		\path[->] (s1) edge  [loop below] node[auto] {$P(\star\lvert \star, \rightarrow)$} ();
		\path[->] (s1) edge [bend left=10] node[auto] {$P(\circ\lvert \star, \rightarrow)$} (s2);
		\path[->] (s1) edge  [loop above,dashed] node[auto] {$P(\star\lvert \star, \dasharrow)$} (s1);
		\path[->] (s1) edge [bend left=50,dashed] node[auto] {$P(\circ\lvert \star, \dasharrow)$} (s2);

		\path[->] (s2) edge [bend left=10]  node[auto] {$P(\star\lvert \circ, \rightarrow)$} (s1);
		\path[->] (s2) edge [loop below] node[auto] {$P(\circ\lvert \circ, \rightarrow)$} ();
		\path[->] (s2) edge [bend left=50,dashed] node[auto] {$P(\star\lvert \circ, \dasharrow)$} (s1);
		\path[->] (s2) edge  [loop above,dashed] node[auto] {$P(\circ\lvert \circ,\dasharrow)$} ();
\end{tikzpicture}
\par\end{centering}
\caption{A generic finite-state Markov Decision Process with two states and
two actions: $\mathcal{S}=\left\{ \star,\circ\right\} $, $\mathcal{A}=\left\{ \rightarrow,\protect\dasharrow\right\} $.
The transition matrix $P\left(s'\lvert s,a\right)$ is a $2\times2\times2$
stochastic matrix, and the reward matrix $R\left(s,a\right)$ is $2\times2$.}
\end{figure}

\begin{figure}
\begin{centering}
\tikzset{every loop/.style={min distance=30mm,looseness=10}}
\begin{tikzpicture}
	\node [state] (s) {} ;
		\path[->] (s) edge  [loop right] node[auto] {$\mathcal{N}\left(r\lvert\mu_{\rightarrow},\sigma_{\rightarrow}\right)$} ();
		\path[->] (s) edge  [loop left,dashed] node[auto] {$\mathcal{N}\left(r\lvert\mu_{\dasharrow},\sigma_{\dasharrow}\right)$} ();
\end{tikzpicture}
\par\end{centering}
\begin{centering}
\label{fig:bandit}
\par\end{centering}
\caption{A two-armed Gaussian Bandit. $\mathcal{A}=\left\{ \rightarrow,\protect\dasharrow\right\} $,
$\left|\mathcal{S}\right|=1$, and $\mathcal{O}=\emptyset$. Rewards
are sampled from the distribution of the respective arm.}
\end{figure}
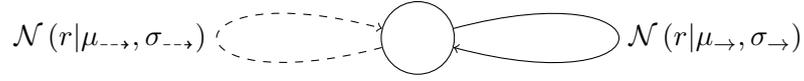

\begin{defn}[Bandit]
\label{ex:bandit} An $N$-armed bandit is a Markovian environment
with one state $\mathcal{S}=\left\{ s\right\} $, $N$ actions $\mathcal{A}=\left\{ a_{1},\dots,a_{N}\right\} $
and $N$ corresponding reward distributions $\left\{ \rho_{1},\dots,\rho_{n}\right\} $
with $\rho_{i}\in\Delta\mathcal{R}$. There are no observations, only
a reward signal which is sampled from the distribution $\rho_{i}$
corresponding to the agent's last action, $a_{i}$. Typical choices
for $\rho$, $\mathcal{R}$ are Bernoulli distributions over $\left\{ 0,1\right\} $,
or Gaussians over $\mathbb{R}$; see \figref{bandit}.
\end{defn}

\begin{example}[Go]
\label{ex:go}Go is a two-player, deterministic, and fully-observable\footnote{Here we are referring to the \emph{game state }being fully observable.\emph{
}The opponent's strategy can of course be modelled as some hidden
variable; for simplicity assume that we model them as minimax, so
that there is no hidden state. } board game with a large, finite state-action space. Played on a $19\times19$
board, there are (naively) $3^{19^{2}}$ possibly game states, with
an action space of $19^{2}$ (though many of these moves will be illegal).
It is notoriously hard to evaluate who is winning in any given game
state \citep{DeepMind:2016Go}; the reward signal is $0$ for all
game states for which a winner has not been declared, and $\pm1$
otherwise (depending on which player won).
\end{example}
\begin{defn}[Policy]
\label{def:policy}A\emph{ policy} is, in the most general setting,
a probability distribution over actions, conditioned on a history:
$\pi\left(a_{t}\lvert\ae_{<t}\right)\ :\ \left(\mathcal{A}\times\mathcal{E}\right)^{*}\to\Delta\mathcal{A}$
. 
\end{defn}
Note the symmetry between \defref{policy} with $\nu\left(e_{t}\lvert\ae_{<t}a_{t}\right)$
from \defref{environment}.
\begin{defn}[Agent]
Let $\Pi_{\mathcal{A},\mathcal{E}}$ be the set of all policies on
the $\left(\mathcal{A},\mathcal{E}\right)$-space. An\emph{ agent}
is fully characterized by a policy $\pi$, and a\emph{ learning algorithm},
which is as a mapping from experience (histories) to policies $\left(\mathcal{A}\times\mathcal{E}\right)^{*}\to\Pi_{\mathcal{A},\mathcal{E}}$.
\end{defn}
The agent and environment, combined, induce a distribution over histories.
We denote this by $\nu^{\pi}\in\Delta\left(\mathcal{A}\times\mathcal{E}\right)^{*}$.
This is equivalent to the \emph{state-action visit distribution }in
the standard reinforcement learning literature \citep{sutton1999policy}.
\begin{equation}
\nu^{\pi}\left(\ae_{<t}\right)=\prod_{k=1}^{t}\pi\left(a_{k}\lvert\ae_{<t}\right)\nu\left(e_{k}\lvert\ae_{<k}\right)\label{eq:nupi}
\end{equation}

The distribution $\nu^{\pi}$ plays an important role in the theory
of GRL, since we will use it to compute the expected sum of future
rewards, which is what our reinforcement learners will seek to maximize.

\subsection{Discounting\label{subsec:discounting}}

In the context of reinforcement learning, we wish our agent to act
according to a policy that maximizes reward accumulated over its lifetime.
In general it is not good enough to greedily maximize the reward obtained
in the next time-step, since in many cases this will lead to reduced
total reward. Thus we define the\emph{ return} resulting from executing
policy $\pi$ in environment $\nu$ from time $t$ as the sum of future
rewards $r_{i}$.
\[
R_{\nu}^{\pi}\left(\ae_{<t}\right)=\sum_{k=t}^{\infty}r_{k},
\]

where each of the $r_{k}$ are sampled from $\nu\left(\cdot|\pi,\ae_{<k}\right)$;
thus the return is a random variable that depends on the agent policy
$\pi$, the environment $\nu$, and the history $\ae_{<t}$. In general
this sum will diverge, so in practice we concern ourselves with either
the\emph{ average reward
\[
\bar{r}_{\nu}^{\pi}=\lim_{n\to\infty}\frac{1}{n}\sum_{k=t}^{n}r_{k},
\]
}

or the\emph{ discounted return
\[
R_{\nu\gamma}^{\pi}\left(\ae_{<t}\right)=\sum_{k=1}^{\infty}\gamma_{k}^{t}r_{k},
\]
}

where $\gamma_{k}^{t}\leq1$ is some generalized \emph{discount} \emph{function}
$\gamma^{t}\ :\ \mathbb{N}\to\left[0,1\right]$ such that
\[
\Gamma_{\gamma}^{t}\stackrel{.}{=}\sum_{k=t}^{\infty}\gamma_{k}^{t}<\infty.
\]

Here we interpret $t$ as the current age of the agent, and $k$ is
the agent's planning look-ahead. 
\begin{defn}[$\boldsymbol{\varepsilon}$-Effective horizon; \citet{LH:2014discounting}]
Given a discount function $\gamma$, the $\varepsilon$\emph{-effective
horizon} is given by
\begin{equation}
H_{\gamma}^{t}\left(\varepsilon\right)\stackrel{.}{=}\min\left\{ H\ :\ \frac{\Gamma_{\gamma}^{t+H}}{\Gamma_{\gamma}^{t}}\leq\varepsilon\right\} .\label{eq:effective-horizon}
\end{equation}
\end{defn}
The $\varepsilon$-effective horizon represents the distance ahead
in the future that the agent can plan while still taking into account
a proportion of the available return equal to $\left(1-\varepsilon\right)$.
The choice of discount function is relevant to how the agent plans;
some discount functions will make the agent far-sighted, and others
will make it near-sighted. We discuss planning more in \secref{planning},
and present experiments relating to this in \chapref{experiments}.
A common choice of discount function is the \emph{geometric }discount
function, which is ubiquitous in RL due to its simplicity: 
\[
\gamma_{k}^{t}=\beta^{k},
\]

for some $\beta\in\left[0,1\right]$. That is, it is $\beta$ raised
to the number of cycles that we look ahead in planning. 

\subsection{Value functions}

In general, the environment $\nu$ is noisy and stochastic, and the
agent's policy will often be stochastic. As a result, we can't maximize
the discounted return directly; we must instead maximize it in expectation.
This follows from the Von Neumann-Morgenstern utility theorem \citep{MvN:1944}. 
\begin{defn}[Value function]
\label{def:value-iterative}The\emph{ value} $V_{\nu\gamma}^{\pi}\ :\ \left(\mathcal{A}\times\mathcal{E}\right)^{*}\to\mathbb{R}$
of a history $\ae_{<t}$ in environment $\nu$ under policy $\pi$
with discount function $\gamma$ is the \emph{expected sum of discounted
future rewards}
\end{defn}
\begin{equation}
V_{\nu\gamma}^{\pi}\left(\ae_{<t}\right)\stackrel{.}{=}\mathbb{E}_{\nu}^{\pi}\left[\left.\sum_{k=t}^{\infty}\gamma_{k}^{t}r_{k}\right|\ae_{<t}\right],\label{eq:value}
\end{equation}

where we use $\mathbb{E}_{\nu}^{\pi}$ above to mean the expectation
with respect to $\nu^{\pi}$, defined in \eqref{nupi}. \eqref{value}
above expresses the value function in \emph{iterative }form. We can
also express it recursively \citep{Leike:2016} using the mutually
recursive relations

\begin{eqnarray*}
V_{\nu\gamma}^{\pi}\left(\ae_{<t}\right) & = & \sum_{a_{t}\in\mathcal{A}}\pi\left(a_{t}\lvert\ae_{<t}\right)V_{\nu\gamma}^{\pi}\left(\ae_{<t}a_{t}\right)\\
V_{\nu\gamma}^{\pi}\left(\ae_{<t}a_{t}\right) & = & \frac{1}{\Gamma_{t}}\sum_{e_{t}\in\mathcal{E}}\nu\left(e_{t}\lvert\ae_{<t}a_{t}\right)\left[\gamma_{t}r_{t}+\Gamma_{t+1}V_{\nu\gamma}^{\pi}\left(\ae_{1:t}\right)\right].
\end{eqnarray*}

For simplicity, from here on we will often omit the $\gamma$ subscript
and make the dependence on the discount function implicit. We will
also suppress the normalization $\frac{1}{\Gamma_{t}}$, as it clutters
the notation and is introduced for technical reasons (so that value
is normalized). Finally, we often also suppress the history $\ae_{<t}$
for clarity.
\begin{defn}[Optimal value \& policy]
\label{def:optimal-value}The\emph{ optimal value} $V_{\nu}^{*}$
achievable in environment $\nu$ given a history $ae_{<t}$ is
\begin{equation}
V_{\nu}^{*}\stackrel{.}{=}\max_{\pi}V_{\nu}^{\pi},\label{eq:optimal-value}
\end{equation}

and the corresponding\emph{ optimal policy $\pi^{*}$ is
\[
\pi_{\nu}^{*}=\arg\max_{\pi}V_{\nu}^{\pi}.
\]
}

Assuming bounded rewards and finite action spaces, these maxima exist
for all $\nu$ \citep{LH:2014pacbayes}, though they are not unique
in general. For our purposes, we allow $\arg\max$ to break ties at
random. At this point it is elucidatory to unroll \eqref{optimal-value}
into the \emph{expectimax} expression 
\begin{equation}
V_{\nu}^{*}\left(\ae_{<t}\right)=\lim_{m\to\infty}\max_{a_{t}}\sum_{e_{t}}\cdots\max_{a_{t+m}}\sum_{e_{t+m}}\sum_{k=t}^{t+m}\gamma_{k}^{t}r_{k}\prod_{j=t}^{k}\nu\left(e_{j}\lvert\ae_{<j}a_{j}\right).\label{eq:expectimax}
\end{equation}

Note that we can do this by using the distributive property of $\max$
over $+$. In \secref{planning}, we will discuss how to approximate
this expectimax calculation for general environments, up to a finite
horizon $m$.
\end{defn}

\subsection{Optimality}

Informally, it makes sense to evaluate an agent's performance against
that of the optimal policy, were it put in the same situation. We
can only sensibly talk about this performance \emph{asymptotically}
in general, that is, in the limit $t\to\infty$, since the agent needs
time to learn the environment, and we can't evaluate the agent after
some finite time $t$, since this time would in general be environment-dependent.
\begin{defn}[Asymptotic optimality; \citealp{LH:2011opt}]
A policy $\pi$ is\emph{ strongly asymptotically optimal} in environment
class $\mathcal{M}$ if $\forall\mu\in\mathcal{M}$
\[
\mu^{\pi}\left(\lim_{t\to\infty}\left\{ V_{\mu\gamma}^{*}\left(\ae_{<t}\right)-V_{\mu\gamma}^{\pi}\left(\ae_{<t}\right)\right\} =0\right)=1,
\]

where $\mu^{\pi}$ is the measure induced by the interaction of environment
$\mu$ with policy $\pi$. The policy $\pi$ is\emph{ weakly asymptotically
optimal} in $\mathcal{M}$ if $\forall\mu\in\mathcal{M}$

\[
\mu^{\pi}\left(\lim_{n\to\infty}\frac{1}{n}\sum_{t=1}^{n}\left\{ V_{\mu\gamma}^{*}\left(\ae_{<t}\right)-V_{\mu\gamma}^{\pi}\left(\ae_{<t}\right)\right\} =0\right)=1.
\]

Finally, we say $\pi$ is \emph{asymptotically optimal in mean }over
$\mathcal{M}$ if $\forall\mu\in\mathcal{M}$ 

\[
\lim_{t\to\infty}\mathbb{E}_{\mu}^{\pi}\left[V_{\mu\gamma}^{*}\left(\ae_{<t}\right)-V_{\mu\gamma}^{\pi}\left(\ae_{<t}\right)\right]=0.
\]
\end{defn}
Asymptotic optimality is objective and general, but unfortunately
doesn't capture everything we want in an agent. For example, in environments
with traps – that is, an accepting state with no transitions leaving
it and very low reward – every policy will be asymptotically optimal
after falling into the trap, since no policy will outperform any other,
conditioned on being trapped. Moreover, in uncertain environments
with traps, an agent cannot be asymptotically optimal \emph{unless}
it is sufficiently gung-ho in its exploration that it eventually falls
into traps \citep{Leike:2016}. Therefore, we should take asymptotic
optimality with a grain of salt; it is not a particularly good measure
of optimality in general environments. The quest for good notions
of optimality is currently an open problem in the theory of GRL \citep{LH:2015priors,Leike:2016}.

\section{General Reinforcement Learning\label{sec:Bayesian-Reinforcement-Learning}}

We now introduce the agents that are central to the theory of general
reinforcement learning (GRL). We begin with AI$\mu$, which is simply
the policy of the informed agent that has a perfect model of the environment
$\mu$: 
\begin{defn}[AI$\mu$]
AI$\mu$ corresponds to the policy in which the true environment
$\mu$ is known to the agent, and so no learning is required. Behaving
optimally reduces to the planning problem of computing the $\mu$-optimal
policy 
\begin{eqnarray*}
\pi^{AI\mu} & = & \pi_{\mu}^{*}\\
 & \stackrel{.}{=} & \arg\max_{\pi}V_{\mu}^{\pi}.
\end{eqnarray*}
\end{defn}
The astute reader will notice that $\pi^{AI\mu}$ is simply the optimal
policy for environment $\mu$; we introduce it here as a separate
agent so as to have a benchmark against which to compare our other
reinforcement learners. 

In general the environment will be unknown, and so our agents will
have to learn it. For the purpose of studying the general reinforcement
learning problem, we consider primarily Bayesian agents, as they are
the most general and principled way to think about the problem of
induction \citep{Hutter:2005}. 

\subsection{Bayesian agents}

Our Bayesian agents maintains a \emph{Bayesian mixture }or \emph{predictive
distribution} $\xi$ over a countable model class $\mathcal{M}$,
given by

\begin{eqnarray}
\xi\left(e_{t}\lvert\ae_{<t}a_{t}\right) & = & \sum_{\nu\in\mathcal{M}}w_{\nu}\nu\left(e_{t}\lvert\ae_{<t}a_{t}\right)\label{eq:bayes-mixture}\\
 & = & \sum_{\nu\in\mathcal{M}}\Pr\left(\nu\lvert\ae_{<t}a_{t}\right)\Pr\left(e_{t}\lvert\nu,\ae_{<t}a_{t}\right).\nonumber 
\end{eqnarray}

Note that $\xi$ is equivalent to the normalization term in \thmref{bayes};
$\xi\left(e\lvert\cdot\right)$ represents the probability that the
agent's model assigns to $e$; in other words, it is the agent's \emph{predictive
distribution}. The \emph{weights} $w_{\nu}\equiv w\left(\nu\right)\equiv\Pr\left(\nu\right)$
represent the agent's credence in hypothesis $\nu\in\mathcal{M}$.
$\nu\left(e\lvert\ae_{<t}a_{t}\right)$ is the probability that model
$\nu$ assigns to percept $e$, given history $\ae_{<t}a_{t}$. One
can think of the hypothesis/model $\nu$ as a \emph{latent variable
}in the model, which is marginalized out to get the predictive distribution.
The only strong assumption we make in this setup is that the \emph{true
environment} $\mu$ is contained in $\mathcal{M}$. Given a new percept
$e=\left(o,r\right)$, the Bayesian updates its weights model using
Bayes rule:

\[
w\left(\nu\lvert e\right)=\frac{w\left(\nu\right)\nu\left(e\right)}{\xi\left(e\right)}.
\]

This gives us the very natural updating scheme
\[
w_{\nu}\leftarrow w_{\nu}\frac{\nu\left(e\right)}{\xi\left(e\right)}.
\]

Note that above we have suppressed the history $\ae_{<t}a_{t}$ for
clarity.

That is, given a new percept $e$, we compute our posterior by multiplying
the prior by the likelihood ratio. Let us pause and make some remarks:
\begin{enumerate}
\item We see that Bayesian induction generalizes the Popperian idea of\emph{
conjecture and refutation}. An environment/model/hypothesis $\nu$
is\emph{ falsified} by observation/perception/experience/experiment
$e$\emph{ iff} $\nu\left(e\right)=0$. Clearly, the posterior goes
to zero for these environments. 
\item Bayesian induction is parameter free up to a choice of model class
$\mathcal{M}$ and prior $w\left(\nu\lvert\epsilon\right)$.
\item We use the words \emph{environment}, \emph{model}, and \emph{hypothesis}
interchangeably. 
\item \label{enu:type}We can see by its `type signature' that $\xi$ itself
is an environment. 
\end{enumerate}
\enuref{type} above underpins the Bayes-optimal agent AI$\xi$.
\begin{defn}[AI\textbf{$\xi$}]
\label{def:AIXI}AI$\xi$ computes the $\xi$-optimal policy, i.e.
\begin{equation}
\pi^{AI\xi}\stackrel{.}{=}\arg\max_{\pi}V_{\xi}^{\pi}.\label{eq:argmax-aixi}
\end{equation}

That is, AI$\xi$ uses the policy that is optimal in the mixture environment
$\xi$, which we update with percepts from the true environment $\mu$
using Bayes' rule. 
\end{defn}
For the purpose of reasoning about general artificial intelligence,
we use the largest model class we can, which is $\mathcal{M}_{\text{comp}}$,
the class of all computable environments\footnote{For technical reasons, the literature typically uses $\mathcal{M}_{CCS}^{LSC}$,
the class of lower semi-computable chronological conditional semimeasures.
This distinction is a technical one and of little consequence to us.}. This is what the famous agent AIXI does:
\begin{defn}[AIXI]
\label{def:AI-XI} AIXI is AI$\xi$ with the model class given by
$\mathcal{M}_{\text{comp}}$ and the Solomonoff prior
\[
w_{\nu}=2^{-K\left(\nu\right)},
\]

where $K\left(\nu\right)$ is the \emph{Kolmogorov complexity }of
$\nu$. For a string $x$, the Kolmogorov complexity is given by 
\[
K\left(\nu\right)\stackrel{.}{=}\min\left\{ \left|p\right|\ \lvert\ U\left(p\right)=x\right\} ,
\]

where $U$ is a universal Turing machine \citep{LV:2008}. For every
computable environment $\nu$, there is a corresponding Turing machine
$T$, so we can define the $K\left(\nu\right)$ as the Kolmogorov
complexity of its index in the enumeration $\nu_{1},\nu_{2},\dots$
of all environments. The Kolmogorov complexity is, of course, incomputable. 

This gives rise to the famous equation describing the AIXI policy,
unrolled in all its incomputable glory: 
\[
a_{t}^{\text{AIXI}}=\arg\max_{a_{t}}\lim_{m\to\infty}\sum_{e_{t}}\cdots\max_{a_{t+m}}\sum_{e_{t+m}}\sum_{k=t}^{t+m}\gamma_{k}^{t}r_{k}\sum_{p\ :\ U\left(p,a_{<t}\right)=e_{1:j}}2^{-\left|p\right|}.
\]

One can derive computable approximations of Solomonoff induction,
most notably by using a generalization of the Context-Tree Weighting
algorithm, which is a mixture over Markov models up to some finite
order $n$, weighted by their complexity; this is used in the well-known
MC-AIXI-CTW implementation due to \citet{VNHUS:2011}.
\end{defn}
AIXI achieves \emph{on-policy value convergence} \citep{Leike:2016}:
\[
\mu^{\pi}\left(\lim_{t\to\infty}\left[V_{\xi}^{\pi}-V_{\mu}^{\pi}\right]=0\right)=1,
\]

which means that it asymptotically learns the true value of its policy
$\pi$ in environment $\mu$. It however, doesn't achieve asymptotic
optimality.
\begin{thm}[\emph{AIXI is not asymptotic optimal; \citealp{Orseau:2010,Leike:2016}}]
For any class $\mathcal{M}\supseteq\mathcal{M}_{\mbox{comp}}$ no
Bayes optimal policy $\pi_{\xi}^{*}$ is asymptotically optimal: there
is an environment $\mu\in\mathcal{M}$ and a time step $t_{0}\in\mathbb{N}$
such that for all time steps $t\geq t_{0}$ 
\[
\mu^{\pi_{\xi}^{*}}\left(V_{\mu}^{*}\left(\ae_{<t}\right)-V_{\mu}^{\pi_{\xi}^{*}}\left(\ae_{<t}\right)=\frac{1}{2}\right)=1.
\]
 
\end{thm}
This theorem effectively means that the Bayes agent will eventually
decide that its current policy is good enough, and that any additional
exploration is not worth its Bayes-expected payoff. Moreover, AI$\xi$
can be made to perform badly with a so-called dogmatic prior:
\begin{thm}[\emph{Dogmatic prior;} \citealp{LH:2015priors}]
\label{thm:dogmatic}Let $\pi$ be some computable policy, $\xi$
some universal mixture, and let $\varepsilon>0$. There exists a universal
mixture $\xi'$ such that for any history $h$ consistent with $\pi$and
$V_{\xi}^{\pi}\left(h\right)>\varepsilon$, the action $\pi\left(h\right)$
is the unique $\xi'$-optimal action.
\end{thm}
This theorem says that, even using a universal prior that assigns
non-zero mass to every hypothesis in the model class, we can construct
a prior in such a way that the agent never overcomes the bias in its
prior. This is in contrast to Bayesian learners in the passive setting,
which can overcome (given sufficient data) any biases in their (universal)
prior. We demonstrate in \chapref{experiments} an example of a dogmatic
prior that prevents the Bayesian agent from exploring. 

\newpage{}

\subsection{Knowledge-seeking agents\label{subsec:ksa}}

We now come to our first exhibit in the GRL agent zoo: knowledge-seeking
agents (KSA). There are several motivations for defining and studying
KSA: 
\begin{itemize}
\item They represent a way to construct a purely `exploratory' policy. A
principled solution to exploration by intrinsic motivation is one
of the central problems in reinforcement learning \citep{Thrun92c}.
\item They remove the dependence on arbitrary reward signals or utility
functions; up to a choice of model class and prior, `knowledge' is
an objective quantity \citep{Orseau:2011ksa}. 
\item They collapse the exploration-exploitation trade-off to \emph{just}
exploration.
\end{itemize}
Before formally defining knowledge-seeking agents, it is necessary
to introduce the concept of a\emph{ utility agent}, which generalizes
the concept of a reinforcement learning agent.
\begin{defn}[Utility Agent; \citealp{Orseau:2011ksa}]
\label{def:utility}A\emph{ utility agent} is a reinforcement learner
equipped with a bounded utility function $u\ :\ \left(\mathcal{A}\times\mathcal{E}\right)^{*}\times\mathcal{A}\to\mathbb{R}$
which replaces the notion of\emph{ reward}. The corresponding value
function\footnote{Compare with \eqref{value}.} is given by
\begin{equation}
V_{\nu\gamma}^{\pi}\left(\ae_{<t}\right)=\mathbb{E}_{\nu}^{\pi}\left[\left.\sum_{k=t}^{\infty}\gamma_{k}^{t}u\left(\ae_{1:k}\right)\right|\ae_{<t}a_{t}\right].\label{eq:value-util}
\end{equation}
\end{defn}
One can easily verify that this definition generalizes RL agents by
setting $u_{\text{RL}}\left(\ae_{1:t}\right)=r\left(e_{t}\right),$
where $r\left(\cdot\right)$ returns the second component of the percept
tuple $e_{t}=\left(o_{t},r_{t}\right)$. Utility agents are fully
autonomous, in the sense that they are not dependent on being `supervised'
by an extrinsic reward signal to learn. They are equipped with a utility
function at birth and from then on seek to maximize the discounted
sum of future utility.

Knowledge-seeking agents (KSA) are Bayesian utility agents whose utility
function is constructed in such a way as to motivate them to `seek
knowledge' and learn about their environment \citep{Orseau:2011ksa,Orseau:2014ksa}.
There are several distinct ways in which one can define knowledge
for a Bayesian agent. We will start by defining an agent that gets
utility from lowering the entropy (i.e., reducing uncertainty) in
its beliefs. We can define the\emph{ information gain} resulting from
some percept $e$ as the difference in entropy between the agent's
prior and posterior:
\begin{equation}
\text{IG}\left(e\right)\stackrel{.}{=}\text{Ent}\left(w\left(\cdot\right)\right)-\text{Ent}\left(w\left(\cdot\lvert e\right)\right),\label{eq:information-gain}
\end{equation}

Now, following \citet{Lattimore:2013}, we consider the $\xi$-expected
information gain. Informally, this is the information that the agent
expects to obtain were the percepts distributed according to its mixture
model $\xi$. It is also the agent's expected utility from seeing
percept $e$. For clarity, we suppress the history $\ae_{<t}a_{t}$
and time subscripts.
\begin{eqnarray*}
\mathbb{E}_{\xi}\left[\text{IG}\left(e\right)\right] & = & \sum_{e\in\mathcal{E}}\xi\left(e\right)\left[\text{Ent}\left(w\left(\cdot\right)\right)-\text{Ent}\left(w\left(\cdot\lvert e\right)\right)\right]\\
 & = & \sum_{e\in\mathcal{E}}\xi\left(e\right)\sum_{\nu\in\mathcal{M}}\left[w\left(\nu\lvert e\right)\log w\left(\nu\lvert e\right)-w\left(\nu\right)\log w\left(\nu\right)\right]\\
 & = & \sum_{e\in\mathcal{E}}\xi\left(e\right)\sum_{\nu\in\mathcal{M}}\left[w\left(\nu\right)\frac{\nu\left(e\right)}{\xi\left(e\right)}\log w\left(\nu\lvert e\right)-w\left(\nu\right)\log w\left(\nu\right)\right]\\
 & = & \sum_{\nu\in\mathcal{M}}w\left(\nu\right)\sum_{e\in\mathcal{E}}\xi\left(e\right)\left[\frac{\nu\left(e\right)}{\xi\left(e\right)}\log w\left(\nu\lvert e\right)-\log w\left(\nu\right)\right]\\
 & = & \sum_{\nu\in\mathcal{M}}w\left(\nu\right)\left[\sum_{e\in\mathcal{E}}\nu\left(e\right)\log w\left(\nu\lvert e\right)-\log w\left(\nu\right)\right]\\
 & = & \sum_{\nu\in\mathcal{M}}w\left(\nu\right)\left[\sum_{e\in\mathcal{E}}\nu\left(e\right)\log w\left(\nu\lvert e\right)-\sum_{e\in\mathcal{E}}\nu\left(e\right)\log w\left(\nu\right)\right]\\
 & = & \sum_{\nu\in\mathcal{M}}w\left(\nu\right)\left[\sum_{e\in\mathcal{E}}\nu\left(e\right)\log\frac{w\left(\nu\lvert e\right)}{w\left(\nu\right)}\right]\\
 & = & \sum_{\nu\in\mathcal{M}}w\left(\nu\right)\left[\sum_{e\in\mathcal{E}}\nu\left(e\right)\log\frac{\nu\left(e\right)}{\xi\left(e\right)}\right]\\
 & = & \sum_{\nu\in\mathcal{M}}w\left(\nu\right)\text{KL}\left(\nu\|\xi\right).
\end{eqnarray*}

Thus, by maximizing the $\xi$-expected information gain, one maximizes
the belief-weighted Kullback-Leibler divergence between $\nu$ and
$\xi$.
\begin{defn}[Kullback-Leibler-KSA; \citealp{Orseau:2014ksa}]
\label{def:klksa}The\emph{ KL-KSA} is the Bayesian agent with
\begin{eqnarray}
u_{\text{KL}}\left(\ae_{1:t}\right) & = & \text{Ent}\left(w\left(\cdot\lvert\ae_{<t}\right)\right)-\text{Ent}\left(w\left(\cdot\lvert\ae_{1:t}\right)\right).\label{eq:utility-kl}
\end{eqnarray}

Notice that the first term doesn't depend on $e_{t}$, so at any given
time step it is fixed by the agent's past history. The term that matters
is the second one, which is the negative entropy of the posterior
beliefs, after updating on percept $e_{t}$. Intuitively, the agent
gets reward from reducing the entropy (uncertainty) in its beliefs;
it seeks out experiences $e_{t}$ that will make it more certain about
the world, and won't be satisfied until entropy is minimal – that
is, when its beliefs converge to the truth such that $w_{\nu}=\mathbb{I}\left[\nu=\mu\right]$
and $\text{Ent}\left(w\right)=0$. In the most general environment
classes, this convergenge won't be possible, as there are many environments
that are indistinguishable on-policy; in other words, there will always
be hypotheses that the agent can't falsify. An example of this is
the so-called \emph{blue emeralds} hypothesis: `Emeralds are green,
but after next Tuesday, they will become blue'. 
\end{defn}
\begin{thm}[\emph{\citealp{Orseau:2014ksa}}]
KL-KSA is asymptotically optimal with respect to $u_{\mbox{KL}}$
in general environments. 
\end{thm}
So much for Kullback-Leibler KSA. There are other ways to construct
a knowledge-seeker. To elucidate this, notice that the entropy in
the Bayesian mixture $\xi$ can be decomposed into contributions from\emph{
uncertainty} in the agent's beliefs $w_{\nu}$ and\emph{ noise} in
the environment $\nu$. That is, given a mixture $\xi$ and for some
percept $e$ such that $0<\xi\left(e\right)<1$, and suppressing the
history $\ae_{<t}a_{t}$ for clarity,
\[
\xi\left(e\right)=\sum_{\nu\in\mathcal{M}}\overbrace{w_{\nu}}^{\text{uncertainty}}\underbrace{\nu\left(e\right)}_{\text{noise}}.
\]

That is, if $0<w_{\nu}<1$, we say the agent is \emph{uncertain} about
whether hypothesis $\nu$ is true (assuming there is exactly one $\mu\in\mathcal{M}$
that is the truth). On the other hand, if $0<\nu\left(e\right)<1$
we say that the environment $\nu$ is \emph{noisy} or \emph{stochastic}.
If we restrict ourselves to deterministic environments such that $\nu\left(e\right)\in\left\{ 0,1\right\} $
$\forall\nu\,\forall e$, then $\xi\left(\cdot\right)\in\left(0,1\right)$
implies that $w_{\nu}\in\left(0,1\right)$ for at least one $\nu\in\mathcal{M}$.
This motivates us to define two agents that seek out percepts to which
the mixture $\xi$ assigns low probability; in deterministic environments,
these will behave like knowledge-seekers. 
\begin{defn}[Square-KSA; \citealp{Orseau:2011ksa}]
\label{def:squareksa}The\emph{ Square-KSA} is the Bayesian agent
with utility function given by
\begin{equation}
u_{\text{Square}}\left(e_{t}\lvert\ae_{<t}\right)=-\xi\left(e_{t}\lvert\ae_{<t}\right).\label{eq:util-square}
\end{equation}
\end{defn}
$\ $
\begin{defn}[Shannon-KSA; \citealp{Orseau:2011ksa}]
\label{def:shannonksa}The\emph{ Shannon-KSA} is the Bayesian agent
with utility function given by
\begin{equation}
u_{\text{Shannon}}\left(e_{t}\lvert\ae_{<t}\right)=-\log\left(\xi\left(e_{t}\lvert\ae_{<t}\right)\right).\label{eq:util-shannon}
\end{equation}
\end{defn}
\begin{thm}[\citealp{Orseau:2014ksa}]
Square-KSA and Shannon-KSA are strongly asymptotically optimal with
respect to $u_{\text{Square }}$ and $u_{\text{Shannon}}$ respectively,
in deterministic environments. 
\end{thm}
They are named `Square' and `Shannon', since in taking $\xi$-expectation
of the utility functions we get 
\begin{eqnarray*}
\mathbb{E}_{\xi}\left[u_{\text{Square}}\left(\cdot\lvert\ae_{<t}\right)\right] & = & -\sum_{e_{t}\in\mathcal{E}}\left[\xi\left(e_{t}\lvert\ae_{<t}\right)\right]^{2}\\
\mathbb{E}_{\xi}\left[u_{\text{Shannon}}\left(\cdot\lvert\ae_{<t}\right)\right] & = & -\sum_{e_{t}\in\mathcal{E}}\xi\left(e_{t}\lvert\ae_{<t}\right)\log\xi\left(e_{t}\lvert\ae_{<t}\right)\\
 & = & \text{Ent}\left(\xi\right).
\end{eqnarray*}

\begin{figure}
\begin{centering}
\includegraphics[scale=0.5]{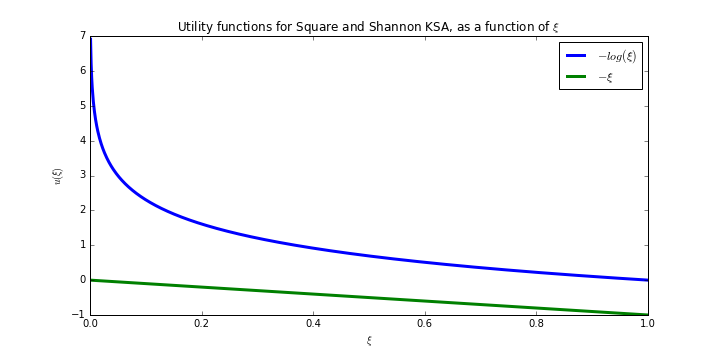}
\par\end{centering}
\caption{Square-KSA utility function plotted against that of Shannon-KSA.}
\label{fig:square-shannon-utility}
\end{figure}

These are \emph{entropy-seeking} agents, since they seek to maximize
the discounted sum of expected utilities, which in both cases are
entropies. Note from \figref{square-shannon-utility} that $u_{\text{Square}}$
and $u_{\text{Shannon}}$ are approximately the same (up to an irrelevant
additive constant) over the range $\left[0.5,1\right]$. Their behaviors
become significantly different for $\xi\to0$: Shannon-KSA \emph{loves
}rare events, and the rarer the better; $u_{\text{Shannon}}$ is unbounded
from above on the interval as $\xi\to0$. The Shannon-KSA, with its
expected utility being measured in bits, is closely related to Schmidhuber's
`curiosity learning', which gets utility from making compression progress
\citep{Schmidhuber-ijcnn-91}. 

Square- and Shannon-KSA both fail in general for stochastic environments.
We can see this by constructing an environment adversarially to `trap'
these agents and stop them from exploring: just introduce a noise
generator that is sufficiently rich (i.e. is sampled from a uniform
distribution over a sufficiently large alphabet of percepts) so that
the probability of any single percept is low enough that it swamps
the utility gained from exploring the rest of the world and gaining
information. Thus, we can get the Square and Shannon KSAs `hooked
on noise' – they would be endlessly fascinated with a white noise
generator such as a detuned television, and would never get sick of
watching the random, low-probability events. We construct an experiment
to explore this property of KSA agents in \chapref{experiments}.

\subsection{BayesExp}

The idea behind the BayesExp agent is simple. Given that KL-KSA is
effective at exploring, and AI$\xi$ is effective (by construction)
at exploiting the agent's beliefs as they stand: why not combine the
two in some way? The algorithm for running BayesExp is simple: run
AI$\xi$ by computing the $\xi$-optimal policy as normal, but at
all times compute the value of the information-seeking policy $\pi^{\text{KSA}}$.
If the expected information gain (up to some horizon) exceeds some
threshold $\epsilon$, run the knowledge-seeking policy for an effective
horizon. This combines the best of AI$\xi$ and KSA, by going on bursts
of exploration when the agent's beliefs suggest that the time is right
to do so; thus, BayesExp breaks out of the sub-optimal exploration
strategy of Bayes, but without resorting to ugly heuristics such as
$\epsilon$-greedy. Crucially, it explores infinitely often, which
is necessary for asymptotic optimality \citep{Leike:2016explore}. 

Essentially, the BayesExp agent keeps track of two value functions:
the Bayes-optimal value $V_{\xi}^{*}$, and the $\xi$-expected \emph{information
gain }value $V_{\xi,\mbox{IG}}^{*}$, which we obtain by substituting
\eqref{utility-kl} into \eqref{value-util}. It then checks whether
$V_{\xi,\mbox{IG}}^{*}$ exceeds some threshold, $\varepsilon_{t}$.
If it does, then it will explore for an effective horizon $H_{t}(\varepsilon_{t})$,
and otherwise it will exploit using the Bayes-optimal policy $\pi_{\xi}^{*}$.
See \algref{bayesexp} for the formal algorithm. 
\begin{thm}[\citealp{Lattimore:2013}]
With a finite prior $w$ and a non-increasing exploration schedule
$\varepsilon_{1},\varepsilon_{2},\dots$, with $\lim_{t\to\infty}\varepsilon_{t}=0$,
BayesExp is asymptotically optimal in general environments. 
\end{thm}
\begin{algorithm}
\begin{algorithmic}[1]
\Require{Model class $\mathcal{M}\stackrel{.}{=}\lbrace\nu_1,\dots,\nu_K\rbrace$; $w\ :\ \mathcal{M}\to(0,1)$; exploration schedule $\lbrace\varepsilon_1,\varepsilon_2,\dots\rbrace$.
}
\State $t \leftarrow 1$
\Loop
\State $d\leftarrow H_t\left(\varepsilon_t\right)$
\If{$V_{\xi,\text{IG}}^{*}\left(\ae_{<t}\right)>\varepsilon_t$}
\For{$i = 1\to d$}
\State $\textsc{act}\left(\pi_{\xi}^{\star,\text{IG}}\right)$
\EndFor
\Else
\State $\textsc{act}\left(\pi_{\xi}^{\star}\right)$
\EndIf
\EndLoop
\end{algorithmic}

\caption{BayesExp \citep{Lattimore:2013}}
\label{alg:bayesexp}
\end{algorithm}

\subsection{MDL Agent}

While AIXI uses the principle of Epicurus to mix over all consistent
environments, the minimum description length (MDL) agent greedily
picks the simplest unfalsified environment in its model class and
behaves optimally with respect to that environment until it falsifies
it. In other words, the policy is given by 
\[
\pi_{\text{MDL}}=\arg\max_{a}V_{\rho}^{*},
\]

where 
\[
\rho=\argmin_{\nu\in\mathcal{M}\ :\ w_{\nu}>0}K\left(\nu\right).
\]

Here, the Kolmogorov complexity $K$ plays the role of a strongly
weighted regularizer. That is, MDL chooses the policy that is optimal
with respect to the simplest unfalsified environment. This algorithm
will fail in stochastic environments, since there will exist environments
which cannot be falsified (in the strict sense, i.e. $w_{\nu}=0$)
by any percept – for example, an environment in which the agent receives
a video feed which is (even slightly) noisy.

\begin{algorithm}
\begin{algorithmic}[1]
\Require{Model class $\mathcal{M}$; prior $w\ :\ \mathcal{M}\to(0,1]$; a total ordering $\preceq$ over $\mathcal{M}$.
}
\Statex
\Loop
\State Select $\rho\leftarrow\min_{\preceq}\mathcal{M}$
\Repeat
\State $\textsc{act}\left(\pi_{\rho}^{\star}\right)$
\Until{$\rho\left(e_{<t}\right) = 0$}
\EndLoop
\end{algorithmic}

\caption{MDL Agent \citep{LH:2011opt}}
\label{alg:mdl}
\end{algorithm}

\subsection{Thompson Sampling}

Thompson sampling is a very common Bayesian sampling technique, named
for \citet{Thompson:1933}. In the context of general reinforcement
learning, it can be used as another attempt at solving the exploration
problems of AI$\xi$. Informally, the idea is to use the $\rho$-optimal
policy for an effective horizon, before re-sampling from the posterior
$\rho\sim w\left(\cdot\lvert\ae_{<t}\right)$ and repeating – at all
times, the agent updates it posterior as usual. This commits the agent
to a single hypothesis for a significant amount of time; one can think
of it as testing likely hypothesis one at a time. See \algref{thompson}
for the formal description of the Thompson sampling policy $\pi_{T}$.
\begin{thm}[\citealp{LLOH:2016Thompson}]
Thompson sampling is asymptotically optimal in mean in general environments. 
\end{thm}
\begin{algorithm}[h]
\begin{algorithmic}[1]
\Require{Model class $\mathcal{M}$; prior $w\ :\ \mathcal{M}\to(0,1]$; exploration schedule $\lbrace\epsilon_1,\epsilon_2,\dots\rbrace$.
}
\Statex
\State $t\leftarrow 1$

\Loop
\State Sample $\rho\sim w\left(\cdot\lvert \ae_{<t}\right)$
\State $d\leftarrow H_t\left(\epsilon_t\right)$
\For{$i = 1\to d$}
\State $\textsc{act}\left(\pi_{\rho}^{\star}\right)$
\EndFor
\EndLoop
\end{algorithmic}

\caption{Thompson Sampling \citep{LLOH:2016Thompson}}
\label{alg:thompson}

\end{algorithm}

So much for our GRL agents. We now discuss how to compute the policy
$\pi_{\rho}^{*}$ for general environments $\rho$, discount functions
$\gamma$, and utility functions $u$. This general-purpose planning
algorithm will be used to select actions for all of the agents above.

\section{Planning\label{sec:planning}}

We have discussed how Bayesian agents maintain a model of their environment,
and update their models based on the percepts they receive. Of course,
the other major aspect of artificial intelligence, distinct from \emph{learning},
is \emph{acting. }Recall that, if the environment is known, computing
the optimal policy becomes a planning problem. In general (stochastic)
environments, this involves computing the optimal value $V_{\mu}^{*}$,
which is the expectimax expression from \eqref{expectimax}. In practice,
with finite compute power, we must of course approximate this expectimax
calculation up to some finite horizon $m$. For finite-state Markov
decision processes with known transitions and rewards, and under geometric
discounting, we can compute this by a simple dynamic programming algorithm
called Value Iteration (\subsecref{Value-iteration}). For more general
environments, we must approximate it by `brute force', with Monte
Carlo sampling (\subsecref{MCTS}). 

\subsection{Value iteration\label{subsec:Value-iteration}}

In a finite-state MDP, if the state transitions $P\left(s'\lvert s,a\right)$
and reward function $R\left(s,a\right)$ are known, then we can plan
ahead by \emph{value iteration}: 
\begin{equation}
V_{n+1}\left(s\right)=\max_{a\in\mathcal{A}}Q_{n}\left(s,a\right),\label{eq:value-iter-1}
\end{equation}

with 
\begin{equation}
Q_{n}\left(s,a\right)=R\left(s,a\right)+\gamma\sum_{s'\in\mathcal{S}}P\left(s'\lvert s,a\right)V_{n}\left(s'\right).\label{eq:value-iter-2}
\end{equation}

This is as a \emph{dynamic programming} algorithm, and is known that
value iteration converges to the value of the optimal policy \citep{SB:1998}:
\[
\lim_{n\to\infty}V_{n}\left(s\right)=V^{*}\left(s\right)\ \forall s\in\mathcal{S}.
\]

Planning by value iteration relies heavily on two strong assumptions:
the finite-state MDP assumption, and geometric discounting. We wish
to be able to lift these assumptions for the purpose of our experiments
in GRL, so we move our attention now to planning by Monte Carlo techniques.

\subsection{MCTS\label{subsec:MCTS}}

Monte Carlo tree search (MCTS) is a general technique for approximating
an expectimax calculation in stochastic games and deterministic games
with uncertainty. Its use dates back several decades, but was popularized
and formalized in the last decade or so in the context of planning
for computer Go \citep{Browne2012}. Analogously to minimax \citep{RN:2010},
we construct a game tree, with $\ts{Max}$ (the agent) playing one
turn, and $\ts{Environment}$ (some distribution over percepts) playing
the other turn. The branching factor of $\ts{Max}$ nodes is of course
$\left|\mathcal{A}\right|$, while the branching factor of $\ts{Environment}$
nodes is upper bounded by $\left|\mathcal{E}\right|$. In \emph{contrast}
to minimax, which is used for deterministic games, we must collect
sufficient samples from $\ts{Environment}$ nodes to get a good estimator
$\hat{V}$ of the expected value for this node. Needless to say, we
wish to avoid expanding the tree out by naively visiting every history
$\ae_{t:m}$. 

Analogously to $\alpha$-$\beta$ pruning in the context of minimax,
UCT is a MCTS algorithm due to \citet{KS2006} that avoids expanding
the whole tree, by only investigating `promising'-looking histories.
These choices must be made under uncertainty, since the environment
is stochastic; hence, we have an instance of the classic exploration-exploitation
dilemma. The UCT algorithm adapts and generalizes the famous UCB1
algorithm used in the context of bandits \citep{AuerCF02}, to balance
exploration and exploitation in the search tree.

UCB stands for `upper confidence-bound', and is a formal version of
the principle of optimism under uncertainty. The general idea is to
add an `exploration-bonus' term to the action selection objective
which prefers actions that haven't been tried much. In the context
of bandits, the UCB action selection is given by 
\begin{equation}
a_{\text{UCB}}=\arg\max_{a\in\mathcal{A}}\left(\hat{R}\left(a\right)+C\sqrt{\frac{\log T}{N\left(a\right)}}\right),\label{eq:ucb}
\end{equation}

where $\hat{R}\left(a\right)$ is the current estimator of the mean
reward that results taking action $a$, $T$ is the lifetime of the
agent, $N\left(a\right)$ is the number of times that $a$ has been
taken, and $C>0$ is a tunable parameter. This exploration bonus allows
us to make a good trade-off between exploration and exploitation.
Consider the exploration bonus term in \eqref{ucb} above: by the
central limit theorem, we can use the fact that the variance in our
estimate of the mean will be approximately bounded by $\frac{1}{\sqrt{N}}$.
The $\log T$ term in the numerator ensures that, asymptotically,
we continue to visit every state-action pair infinitely often; this
is necessary to establish regret bounds \citep{AJO:2009}. Thus, \eqref{ucb}
captures the concept of `exploration under uncertainty' in a principled
way; UCT adapts this to the Monte Carlo tree search planning setting,
in Markov decision processes (MDPs).

While UCT is sufficient for planning in unknown MDPs, we need to generalize
to \emph{histories} for planning in general environments. \citet{VNHUS:2011}
present this generalization, $\rho$UCT, in their famous MC-AIXI-CTW
implementation paper, based on earlier work in Monte Carlo planning
on partially-observable MDPs \citep{SilverVeness2010}. Using this
algorithm, we don't need to know the state transitions as is required
for value iteration (\eqref{value-iter-2}); we instead only need
some black-box environment model $\rho$.  The $\rho$UCT action-selection
within each decision node of the tree search is given by 
\begin{equation}
a_{\text{UCT}}=\arg\max_{a\in\mathcal{A}}\left(\frac{1}{m\left(\beta-\alpha\right)}\hat{V}\left(\ae_{<t}a\right)+C\sqrt{\frac{\log T\left(\ae_{<t}\right)}{T\left(\ae_{<t}a\right)}}\right),\label{eq:uct}
\end{equation}

where $\beta-\alpha$ is the reward range, and $m$ is the planning
horizon; together they are used to normalize the mean value estimate
$\hat{V}$ for the history under consideration, $\ae_{<t}a$. As in
\eqref{ucb}, $C$ is a positive parameter which controls how much
we weight the exploration bonus. The exploration bonus itself is of
a similar form, although note that we use $\log T\left(\ae_{<t}\right)$
in the numerator, i.e. the logarithm of the number of times we've
visited the current history node.

Notice that, in contrast to so-called `model-free' methods such as
Q-learning, our GRL agents can't memorize or cache the value function
in general; this is because we can only compute the value of a \emph{history
}and not of \emph{states}, because of the weakness of our modelling
assumptions. Clearly we can never visit any history $\ae_{1:t}$ more
than once, so memorization is useless. For this reason, in general
our agent has to \emph{re-compute the value at each time step} $t$,
so as to plan its next action $a_{t}$. Hence, all of our model-based
(Bayesian) agents must plan at each time step by forward simulation
with $\rho$UCT Monte Carlo tree search. As we will see in \secref{Performance},
this is the major computational bottleneck for our GRL agents. Moreover,
planning with MCTS requires us to have finite (and, ideally, small)
action and percept spaces. In \secref{Planners}, we discuss our implementation
of $\rho$UCT, along with some subtle emergent issues. 

In \algref{mcts}, we present a (slightly expanded, for clarity) version
of the $\rho$UCT algorithm due to \citet{VNHUS:2011}. 

\begin{algorithm}[H]
\begin{algorithmic}[1]
\Require{History $h$; Search horizon $m$; Samples budget $\kappa$; Model $\rho$}

\State $\textsc{Initialize}\left(\Psi\right)$
\State $n_{\text{samples}} \leftarrow 0$
\Repeat
	\State $\rho' \leftarrow \rho.\textsc{Copy}()$
	\State $\textsc{Sample}\left(\Psi,h,m\right)$
	\State $\rho \leftarrow \rho'$
\Until $n_{\text{samples}} = \kappa$
\State \Return $\arg\max_{a\in\mathcal{A}}\hat{V}_{\Psi}(a)$

\Statex

\Function{Sample}{$\Psi,h,m$}
	\If {$m = 0$}
		\State \Return 0
	\ElsIf {$\Psi(h)$ is a chance node}
		\State $\rho.\textsc{Perform}(a)$
		\State $e=(o,r)\leftarrow \rho.\textsc{GeneratePercept}()$
		\State $\rho.\textsc{Update}(a,e)$
		\If {$T(he) = 0$}
			\State Create chance node $\Psi(he)$
		\EndIf
		\State reward $\leftarrow e.\textsc{reward} + \textsc{Sample}\left(\Psi,he,m-1\right)$
	\ElsIf {$T(h) = 0$}
		\State reward $\leftarrow \textsc{Rollout}(h,m)$
	\Else
		\State $a \leftarrow \textsc{SelectAction}\left(\Psi,h\right)$
	\EndIf
\State $\hat{V}(h)\leftarrow \frac{1}{T(h)+1}\left(\text{reward}+T(h)\hat{V}(h)\right)$
\State $T(h)\leftarrow T(h) + 1$
\EndFunction

\Statex

\Function{SelectAction}{$\Psi,h$}
	\State $\mathcal{U} = \{a\in\mathcal{A}\ :\ T(ha)=0\}$
	\If {$\mathcal{U} \neq \emptyset$}
		\State Pick $a\in\mathcal{U}$ uniformly at random
		\State Create node $\Psi(ha)$
		\State \Return $a$
	\Else
		\State \Return $\arg\max_{a\in\mathcal{A}}\left\{\frac{1}{m\left(\beta-\alpha\right)}\hat{V}(ha)+C\sqrt{\frac{\log(T(h))}{T(ha)}}\right\}$
	\EndIf	
\EndFunction

\Statex

\Function{Rollout}{$h,m$}
	\State reward $\leftarrow 0$
	\For {$i=1$ to $m$}
		\State $a \sim\pi_{\text{rollout}}(h)$
		\State $e=(o,r) \sim \rho(e\lvert ha)$
		\State reward $\leftarrow$ reward $+ r$
		\State $h\leftarrow hae$
	\EndFor
	\State \Return reward
\EndFunction
\end{algorithmic}

\caption{$\rho$UCT \citep{VNHUS:2011}.}
\label{alg:mcts}
\end{algorithm}

\section{Remarks\label{sec:Remarks}}

We now conclude with a short summary, and some remarks. 

In this chapter, we presented the problem of general reinforcement
learning, in which the goal is to construct an agent that is able
to learn an optimal policy in a broad class of (partially observable
and non-ergodic) environments. We have presented the current state-of-the-art
GRL agents and algorithms, namely AI$\xi$, Thompson sampling, MDL,
Square-, Shannon-, and Kullback-Leibler-KSA, and BayesExp, under a
unified notation, and we have discussed the ideas and algorithms that
allow these agents to learn and plan. These agents, and the analysis
and formalism around them, represent our best theoretical understanding
of rationality and intelligence in this general setting. In the subsequent
two chapters, we present our software implementation of these agents,
and some experiments we run on them.

\chapter[Implementation]{Implementation\protect\footnote{AIXIjs was implemented in collaboration with Sean Lamont, a second-year
undergraduate student at the ANU. Sean wrote many of the visualizations
under my supervision; the rest of the implementation is my own work.
More detailed contribution information (including commit history)
can be found at \protect\href{https://github.com/aslanides/aixijs/graphs/contributors}{https://github.com/aslanides/aixijs/graphs/contributors}.}\label{chap:AIXIjs}}
\begin{quotation}
\emph{There are no surprising facts, only models that are surprised
by facts; if a model is surprised by the facts, it is no credit to
that model. }
\end{quotation}
We now present the design and implementation of the open-source software
demo, $\ts{AIXIjs}$.\footnote{The demo can be run at \href{http://aslanides.github.io/aixijs}{http://aslanides.github.io/aixijs};
all supporting source code can be found at \href{http://github.com/aslanides/aixijs}{http://github.com/aslanides/aixijs}.
We encourage the reader to interact with the demo, though again, we
strongly recommend using Google Chrome, as the software was not tested
on other browsers, for reasons detailed in \secref{JavaScript-web-demo}.} Our implementation can be decomposed into roughly five major components,
or modules, which we discuss in this chapter: 
\begin{itemize}
\item \textbf{Agents}. We implement the agents specified in \chapref{Background}.
Some of them differ by one line of code; for example, the KSA agents
can be built from AI$\xi$ by simply replacing its utility function.
We document the agent implementation in \secref{Agents}.
\item \textbf{Environments}. We design and implement environments to showcase
the various agents, including a partially observable Gridworld, and
a `chain' MDP environment; both are documented in \secref{Environments}.\footnote{We also implement multi-armed bandits, generic finite-state MDPs,
and iterated prisoner's dilemma, but we don't document them here as
they don't play a prominent role in the demos or experiments.} 
\item \textbf{Models}. For our Bayesian agents, we design and implement
two model classes with which they can learn the Gridworld environment,
$\mathcal{M}_{\mbox{loc}}$ and $\mathcal{M}_{\mbox{Dirichlet}}$.
These are presented in \secref{Models}.
\item \textbf{Planners}. We implement value iteration and $\rho$UCT Monte
Carlo tree search, which were presented in \secref{planning}. We
make some implementation-specific remarks in \secref{Planners}.
\item \textbf{Visualization and user interface}. We design and implement
a user interface that allows the user to choose demos, read background
and demo-specific information, tune parameters, and run experiments.
We also present a graphic visualization for showing the agent-environment
interaction, and for plotting the agent's performance; this is presented
in \secref{Visualization-and-user}. 
\end{itemize}
First, we briefly discuss the software tools we used to implement
the project.

\section{JavaScript web demo\label{sec:JavaScript-web-demo}}

We implement $\textsc{AIXIjs }$ as a static web site. That is, apart
from web hosting for the ${\tt .html}$ and ${\tt .js}$ source code
and other site assets, there is no back-end server required to run
the software; the demo runs natively, and locally, in the user's web
browser. All of the agent-environment simulations are implemented
in modern JavaScript (ECMAScript 2015 specification), with minimal
use of external libraries. This allows us to effectively build a lightweight\footnote{Including all source code, external libraries, fonts, text, and image
assets, the software totals less than $2$ megabytes in size, uncompressed.} and portable software suite, which a modern web browser can run without
the need for specialized dependencies such as compilers or scientific
libraries. 

JavaScript (JS) is a high-level, dynamic, and weakly-typed language
typically used to create dynamic content on websites. Google's V8
JS engine, implemented in their Chrome web browser, provides a fast
JS runtime; in many benchmarks, it is significantly faster than Python
3.\footnote{For inter-language comparisons on common benchmarks, see, for example,
\href{http://benchmarksgame.alioth.debian.org/u64q/compare.php?lang=node&lang2=python3}{http://benchmarksgame.alioth.debian.org/u64q/compare.php?lang=node\&{}lang2=python3}.} As we discussed in the \nameref{chap:Introduction}, this allows
for computationally intensive and visually impressive software. JavaScript,
however, does have several shortcomings. The ones that are relevant
to us are: 
\begin{itemize}
\item JavaScript is a notoriously\footnote{The author highly recommends a brilliant four-minute video by Gary
Bernhardt about the nonsense that comes from JavaScript's (lack of)
type system: \href{https://www.destroyallsoftware.com/talks/wat}{https://www.destroyallsoftware.com/talks/wat}.} weakly-typed language, which comes with all the programming pitfalls
and runtime errors one would expect. For example, functions will silently
accept arguments that are null, and attempt to perform computations
on them. In this way, subtle bugs can cause catastrophic runtime errors
that can propagate quite far without being caught. We mitigate this
to some extent by writing tests using the $\ts{QUnit}$ testing framework,
and by frequently using the built-in debugger in Google Chrome.
\item JavaScript implementations differ between browsers. For example, some
features of the ECMAScript 2015 specification (for example, anonymous
functions) were not yet implemented by the latest version of the Safari
web browser as of September 2016. Worse still, behaviors can differ
subtly and in undocumented ways between browser implementations. We
(unfortunately) are forced to work around this by only supporting
recent versions of Google Chrome,\footnote{The software was last tested on Google Chrome version 54.0.}
and discouraging usage on other web browsers.
\end{itemize}
We use standard web frameworks and libraries: \href{https://jquery.com/}{jQuery}
and \href{http://getbootstrap.com/}{Bootstrap} for presentation;
\href{https://d3js.org}{d3js} for graphics and visualizations; \href{https://github.com/chjj/marked}{marked}
for MarkDown parsing, and \href{https://www.mathjax.org/}{MathJax}
for rendering mathematics in the browser. Our implementation totals
roughly 6000 lines of JavaScript.

We make use of a modular design, and use class inheritance frequently,
so as to minimize code duplication and to leverage the conceptual
connections between objects. In the sections that follow, we occasionally
use simple UML diagrams to document these classes. Note that in these
diagrams we use type annotations, for expository purposes. 

\section{Agents\label{sec:Agents}}

\begin{figure}
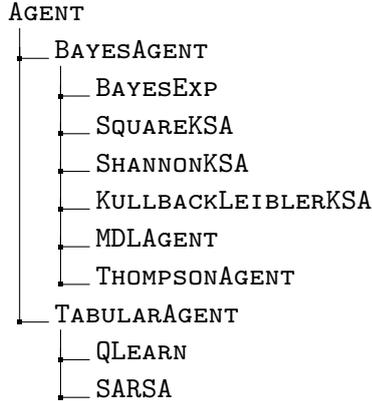

\dirtree{%
.1 $\textsc{Agent}$.
	.2 $\textsc{BayesAgent}$.
		.3 $\textsc{BayesExp}$.
		.3 $\textsc{SquareKSA}$.
		.3 $\textsc{ShannonKSA}$.
		.3 $\textsc{KullbackLeiblerKSA}$.
		.3 $\textsc{MDLAgent}$.
		.3 $\textsc{ThompsonAgent}$.
	.2 $\textsc{TabularAgent}$.
		.3 $\textsc{QLearn}$.
		.3 $\textsc{SARSA}$.		
}

\caption{$\textsc{Agent}$ class inheritance tree. Note that the BayesAgent
is simply AI$\xi$.}
\label{fig:agent-tree}
\end{figure}

All agents inherit from the base $\textsc{Agent}$ class. Every agent's
constructor takes an $\ts{Options}$ object as input, which allows
us to pass in default and user-specified options. See \figref{agent-tree}
for the full agent class inheritance tree. The $\textsc{Agent}$ base
class specifies the methods 
\begin{itemize}
\item $\ts{Update}\left(a,e\right)$. Update the agent's model of the environment,
given that it just performed action $a\in\mathcal{A}$ and received
percept $e\in\mathcal{E}$ from the environment. 
\item $\ts{SelectAction}()$. Compute, and sample from, the agent's (in
general, stochastic) policy $\pi\left(a\lvert\ae_{<t}\right)$, returning
an action $a\in\mathcal{A}$.
\item $\ts{Utility}\left(e\right)$. This is the agent's utility function,
as defined in \defref{utility}. For reward-based reinforcement learners
it simply extracts the reward component from ${\tt percept}$.
\end{itemize}
Every agent is further equipped with a \textbf{$\textsc{Discount}$}
function, as defined in \subsecref{discounting}. 

\begin{figure}
\begin{centering}
\begin{tikzpicture}
\umlclass{BayesAgent}{
	discount : Discount \\
	horizon : Number \\
	ucb : Number \\
	model : Mixture \\
	planner : ExpectimaxTree
}{
	update(a : Action, e : Percept) : null \\
	selectAction(null) : Action \\
	utility(e : Percept) : Number	
}
\end{tikzpicture}
\par\end{centering}
\caption{$\protect\ts{BayesAgent}$ UML. ${\tt discount}$ is the agent's discount
function, $\gamma_{k}^{t}$. ${\tt horizon}$ is the agent's MCTS
planning horizon, $m$. ${\tt ucb}$ is the MCTS UCB exploration parameter
$C$.}
\end{figure}

Every Bayesian agent (i.e. of class $\textsc{BayesAgent}$ or one
of its descendants) is further composed of a $\textsc{Model}$ and
a $\textsc{Planner}$, which are both central to its operation. When
we call $\textsc{Update}\left(a,e\right)$ on $\textsc{BayesAgent}$,
it saves its model's state\footnote{As we shall see in \secref{Models}, models must behave like environments.
\defref{environment} implies that they must therefore in general
maintain some internal state $s$ that is changed by actions $a\in\mathcal{A}$.}, calls the model's $\textsc{Update}\left(a,e\right)$ method, and
then computes and stores the information gain (defined in \eqref{information-gain})
between the old and new model states. When we call $\textsc{SelectAction}$,
the agent passes its model to the $\textsc{Planner}$, and waits for
it to compute a best action. If the information gain from the previous
action was non-zero, the planner's internal state is reset; otherwise,
we prune the search tree but keep the partial result; see \secref{Planners}
for more discussion regarding the planner. 

The other agents inherit from $\textsc{BayesAgent}$, and differ from
it in straightforward and transparent ways specified by their respective
definitions (\defref{squareksa}, \defref{shannonksa}, \defref{klksa},
\algref{bayesexp}, \algref{mdl}, and \algref{thompson}). We won't
reproduce their source code here; the interested reader can find the
code in the ${\tt src/agents/}$ directory in the \href{http://github.com/aslanides/aixijs}{GitHub repository}.

\subsection{Approximations\label{subsec:Approximations}}

We now enumerate and justify our use of several approximations and
simplifications in our agent implementations. The first two approximations
are motivated by computational considerations. In both cases, we argue
that our use of these simplifications leaves the agent's policy invariant.
The third simplification is in fact forced upon us; it is inconvenient
and potentially highly consequential to the performance of Shannon-KSA.
\begin{itemize}
\item \textbf{Information gain}. Recall from \chapref{Background} that
the information gain for a Bayesian agent given a history $\ae_{<t}$
is 
\[
\text{IG}\left(e\lvert\ae_{<t}a_{t}\right)\stackrel{.}{=}\text{Ent}\left(w\left(\cdot\lvert\ae_{<t}\right)\right)-\text{Ent}\left(w\left(\cdot\lvert\ae_{1:t}\right)\right),
\]

and recall that this is the utility function of the Kullback-Leibler
knowledge-seeking agent (KL-KSA). Now, when computing the KL-KSA policy
at time $t$ \textendash{} that is, in calls to $\textsc{SelectAction}$
\textendash{} we compute the value $V_{\xi}^{\pi,\mbox{IG}}$ of various
potential histories $\ae_{<t}a_{t}e_{t}a_{t+1}e_{t+1}\dots a_{t+m}e_{t+m}$,
and select the action that maximizes this value. Note that our action-selection
doesn't depend on the \emph{absolute value} of different histories,
but only on their \emph{relative} value. Note also that the quantity
$\text{Ent}\left(w\left(\cdot\lvert\ae_{<t}\right)\right)$ does not
depend on future actions or percepts, as it is determined by events
in the agent's past. Hence it is a constant that we can ignore when
comparing the relative value of future actions. From the definition
of entropy (\eqref{entropy}), we see that computing the entropy of
the posterior $w\left(\nu\lvert\cdot\right)$ requires $\mathcal{O}\left(\left|\mathcal{M}\right|\right)$
operations. For this reason, in our implementation of the KL-KSA,
we achieve a $2\times$ speedup by replacing $u_{\mbox{KL}}$ with
the surrogate utility function 
\[
u'_{\mbox{KL}}\left(\ae_{1:t}\right)=-\text{Ent}\left(w\left(\cdot\lvert\ae_{1:t}\right)\right).
\]

\item \textbf{Effective horizon}.\textbf{ }Recall from \algref{thompson}
and \algref{bayesexp} that the Thompson sampling and BayesExp agents
both explore for an effective horizon $H_{\gamma}^{t}\left(\varepsilon\right)$
(\eqref{effective-horizon}); this requirement is, in fact, essential
to the proofs of their asymptotic optimality. However, computing the
effective horizon exactly for general discount functions is not possible
in general, although approximate effective horizons have been derived
for some common choices of $\gamma$ (\citealp{Lattimore:2013}; Table
2.1). Moreover, in practice, due to the computational demands of planning
with MCTS (\algref{mcts}), we are forced to plan only with a relatively
short horizon $m$; for most discount functions $\gamma$ and realistic
$\varepsilon$,\footnote{Recall that in the case of BayesExp, $\varepsilon$ is compared to
the value of the knowledge-seeking policy, $V_{\xi}^{*,\mbox{IG}}$.} the true effective horizon $H_{\gamma}^{t}\left(\varepsilon\right)$
is significantly greater than $m$. For this reason, and for simplicity
and ease of computation, we use the MCTS planning horizon $m$ as
a surrogate for $H_{\gamma}^{t}$. Naturally, this choice affects
the agent's policy, but no more so than we already have by using MCTS
to plan up to some (finite, time-constrained, and pragmaticaly chosen)
horizon $m$ rather than to infinity, as the agents do in the theoretical
formalism.
\item \textbf{Utility} \textbf{bounds}. Recall from the $\rho$UCT action
selection algorithm (\eqref{uct}) that the value estimator $\hat{V}\left(\ae_{1:t}\right)$
is normalized by a factor of $m\left(\beta-\alpha\right)$, where
$m$ is the MCTS planning horizon, and $\alpha$ and $\beta$ are
the minimum and maximum rewards that the agent can receive in any
given percept. In the case of reward-based reinforcement learners,
$\alpha$ and $\beta$ are essentially metadata provided to the agent,
along with the size of the action space $\left|\mathcal{A}\right|$,
at the beginning of the agent-environment interaction. For utility-based
agents, however, the rewards are generated \emph{internally}, and
so the agent must calculate for itself what range of utilities it
expects to see, so as to correctly normalize its value function for
the purposes of planning. 

Thankfully, for the Square and Kullback-Leibler KSAs, this is relatively
easy to do. Since $u_{\mbox{Square}}\left(e\right)=-\xi\left(e\right)$,
we can immediately bound its utilities in the range $\left[-1,0\right]$.
In general this won't be a tight bound, since there exist environment
mixtures in which every percept is in some smaller range, i.e. $\xi\left(\cdot\right)\in\left[a,b\right]$
with $a>-1$ and $b<0$,\footnote{For example, a coinflip environment in which the agent is trying to
falsify one of two hypotheses: whether a coin is fair ($\nu\left(\cdot\right)=0.5$)
or bent ($\nu\left(\cdot\right)\neq0.5$).} but in practice, and in particular for our model classes, it is effectively
a tight bound. 

In the case of the Kullback-Leibler KSA, recall that $u_{\mbox{KL}}\left(e\right)=\text{Ent}\left(w\left(\cdot\right)\right)-\text{Ent}\left(w\left(\cdot\lvert e\right)\right)$
. If we assume that we are given the maximum-entropy (i.e. uniform)
prior $w\left(\cdot\lvert\epsilon\right)$, then clearly $u_{\mbox{KL}}\left(e\right)\leq\text{Ent}\left(w\left(\cdot\lvert\epsilon\right)\right)\ \forall e\in\mathcal{E}$,
since entropy is always non-negative. Hence we have $0\leq u_{\mbox{KL}}\leq\text{Ent}\left(w\left(\cdot\lvert\epsilon\right)\right)$,
i.e. the KL-KSA's rewards are bounded from above by the entropy of
its prior (assuming a uniform prior), and from below by zero.

Finally, we come to the problematic case: from \figref{square-shannon-utility},
we know that $u_{\mbox{Shannon}}\left(e\right)=-\log\xi\left(e\right)$
is unbounded from above as $\xi\to0$. This means that unless the
agent can \emph{a priori} place lower bounds on the probability that
its model $\xi$ will assign to an arbitrary percept $e\in\mathcal{E}$,
it cannot upper bound its utility function and therefore cannot normalize
its value function correctly. This is problematic for us, especially
as our environments and models are constructed in such a way as to
allow arbitrarily small probabilities, as we will see in \secref{Environments}
and \secref{Models}. 

Unfortunately, it seems we're stuck here. We're forced to make an
ugly, arbitrary choice to upper bound the Shannon agent's utility
function with, so as to normalize its value function. If we choose
the upper bound $\beta$ too high, then the $\hat{V}$ term in \eqref{uct}
will be artificially, but consistently small; this is equivalent to
inflating the exploration bonus constant $C$ by roughly a constant
multiplicative factor (which is itself upper bounded by some function
of $\beta$). If $\beta$ is chosen too small, however, we can run
into much bigger problems, since now $\hat{V}$ can be over-inflated
by an unboundedly large multiplicative factor. If Shannon KSA sees
a very improbable percept, its value estimates will blow up, which
will cause suboptimal plan selection, since the $\hat{V}$ will overwhelm
the exploration bonus term in \eqref{uct}. We are forced to choose
a $\beta$, so we use a very large upper bound, $\beta=10^{3}$ in
an attempt to balance this trade-off, but bias it in favor of overestimating
$\beta$. For us to exceed $-10^{3}$ in $\log_{2}$ probability requires
us to assign a probability of $\xi\left(e\right)\leq2^{-10^{3}}=10^{-301}$,
which is approaching the limits of numerical precision in JavaScript.
With this setting of $\beta$ we are unlikely to blow up our value
estimate, although we will be severely inflating the UCB constant.
As we will see in \chapref{experiments}, this is quite possibly the
cause of some suboptimal behavior in the Shannon KSA.
\end{itemize}

\section{Environments\label{sec:Environments}}

Recall that AIXI and its variants are theoretical models of unbounded
rationality, not practical algorithms. Bayesian learning and planning
by forward simulation with Monte Carlo tree search are both very computationally
demanding, so we restrict ourselves to demonstrating their properties
on small-scale POMDPs and MDPs. 

Analogously to the case of agents, all environments inherit from the
base $\textsc{Environment}$ class. Every environment's constructor
takes an $\ts{Options}$ object as input, which allows us to pass
in default and user-specified options. The $\textsc{Environment}$
base class specifies the methods 
\begin{itemize}
\item $\textsc{ConditionalDistribution}(e)$. Returns the probability $\nu\left(e_{t}\lvert ae_{<t}a_{t}\right)$
that the environment assigns to percept $e$ given its current state
resulting from the history $\ae_{<t}a_{t}$.
\item $\textsc{GeneratePercept}()$. Sample from $\nu\left(e\right)$ and
returns a percept $e\in\mathcal{E}$.
\item $\textsc{Perform}(a)$. Take in action $a\in\mathcal{A}$ and mutate
the environment's (in general, hidden) state according to its dynamics. 
\item $\textsc{Save}()$ and $\textsc{Load}()$. These functions save and
load the environment's internal state. This is a convenience for our
Bayesian agents; it allows them to reset the environments $\nu\in\mathcal{M}$
that make up their mixture model, after running counterfactual simulations
in a planner. 
\end{itemize}
\begin{figure}
\begin{centering}
\begin{tikzpicture}
\umlclass{Environment}{
	state : Object \\
	minReward : Number \\
	maxReward : Number \\
	numActions : Number
}{
	generatePercept() : Percept \\
	perform(a : Action) : null \\
	conditionalDistribution(e : Percept) : Number \\
	save() : null \\ 
	load() : null \\
}
\end{tikzpicture}
\par\end{centering}
\caption{$\protect\ts{Environment}$ UML. ${\tt state}$ is the environment's
current state, it is simply of type ${\tt Object}$, since we are
agnostic as to how the environment's state is represented. If JavaScript
supported privated attributes, this would be private to the environment,
to enforce the fact that the state is hidden in general. In contrast,
${\tt minReward}$ ($\alpha$), ${\tt maxReward}$ ($\beta$), and
${\tt numActions}$ ($\left|\mathcal{A}\right|$) are public attributes:
it is necessary that the agent know these properties so that the agent-environment
interaction can take place.}
\end{figure}

We now introduce the Gridworld and Chain environments. The interested
reader can find the source code to these, and other, environments
in the ${\tt src/environments/}$directory in the \href{http://github.com/aslanides/aixijs}{GitHub repository}.

\subsection{Gridworld\label{subsec:Gridworld}}

Our gridworld consists of an $N\times N$ array of tiles. There are
four types of tiles: $\textsc{Empty}$, $\textsc{Wall}$, $\textsc{Dispenser}$,
and $\textsc{Trap}$, with the following properties:
\begin{itemize}
\item $\textsc{Empty}$ tiles allow the agent to pass, albiet while incurring
a small movement penalty $r_{\textsc{Empty}}$. 
\item $\textsc{Wall}$ tiles are not traversable. If the agent walks into
a wall, it incurs a negative penalty $r_{\textsc{Wall}}<r_{\textsc{Empty}}$.
\item $\textsc{Dispenser}$ tiles behave like $\textsc{Empty}$ tiles as
far as movement and observations are concerned, but they dispense
some large reward $r_{\textsc{Cake}}\gg r_{\textsc{Empty}}$ with
probability $\theta$, and $r_{\textsc{Empty}}$ otherwise; that is,
all $\textsc{Dispenser}$s are (scaled) $\text{Bernouilli}\left(\theta\right)$
processes.\footnote{The $\textsc{AIXIjs}$ agent mascot is Roger the Robot$\ $\includegraphics[scale=0.1]{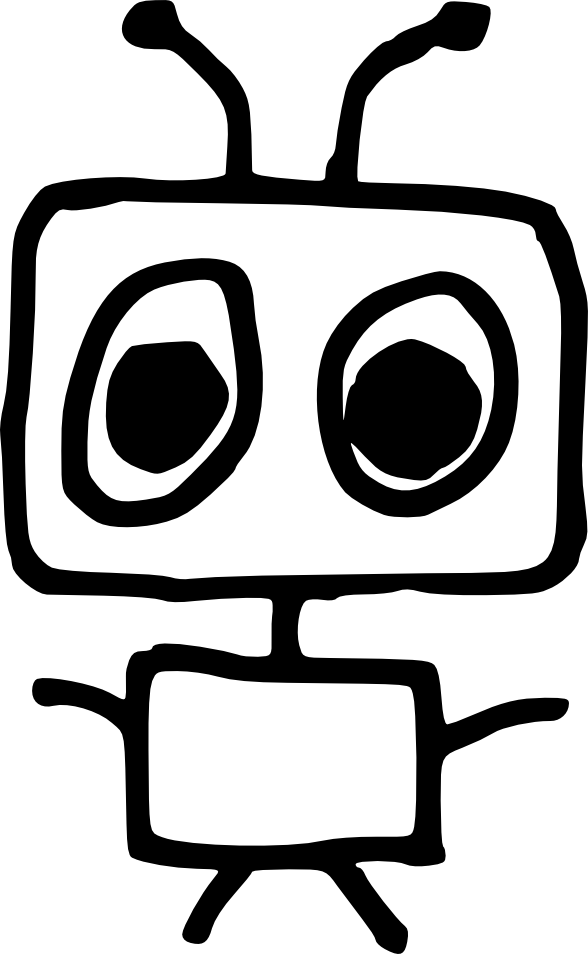}.
Roger likes $\textsc{Cake}$, and will do anything it takes to get
near a $\textsc{Cake}$ $\textsc{Dispenser}$.}
\item $\textsc{Trap}$ tiles, as the name suggests, don't allow you to leave.
Moreover, once stuck in a trap, the agent will receive $r_{\textsc{Wall}}$
reward constantly.
\end{itemize}
The gridworld we construct is a POMDP; the environment's hidden state
is the agent's grid position $s=\left(i,j\right)$ and the positions
of all walls, traps, and dispensers. Observations consist of a bitstring
telling the agent whether the adjacent squares in the $\left\{ \leftarrow,\rightarrow,\right.\left.\uparrow,\downarrow\right\} $
directions are $\textsc{Wall}$s or not; the edges of the Gridworld
are treated implicitly as walls. The agent can move in these four
cardinal directions, or stand still (this is the so-called `no-op',
which we denote by $\bigcirc$). The only way to distinguish a $\textsc{Dispenser}$
from an $\textsc{Empty}$ tile is to walk onto it and observe the
(in general, stochastic) reward signal; for low values of $\theta$,
it may take some time for a $\textsc{Dispenser}$ to reveal itself.
The only way to distinguish a $\textsc{Trap}$ from an $\textsc{Empty}$
tile is to walk onto it and see if you get trapped or not. Hence,
we can characterize the action and percept spaces as
\begin{eqnarray*}
\mathcal{A} & = & \left\{ \leftarrow,\rightarrow,\uparrow,\downarrow,\bigcirc\right\} \\
\mathcal{E} & = & \mathbb{B}^{4}\times\left\{ r_{\mathrm{\textsc{Wall}}},r_{\textsc{Empty}},r_{\textsc{Cake}}\right\} .\\
\end{eqnarray*}

Movement and observations are all deterministic; the only stochasticity
in this environment arises from the reward signal from the dispenser(s).

For the purposes of our demos and experiments, we generate random
gridworlds by independently and randomly assigning each tile to one
of the four classes, with a strong bias towards being $\textsc{Empty}$,
and a slighter weaker bias towards being a $\textsc{Wall}$. The agent's
starting position is always the top left corner, at tile $\left(0,0\right)$.
We ensure that the gridworld is solvable by ensuring there is at least
one dispenser, and by running a breadth-first-search to check whether
there is a viable path from the agent's starting position to the dispenser
with the highest pay-out frequency, $\theta$.

This gridworld environment is sufficiently rich and interesting to
demonstrate most of what we seek to show: the agents have to reason
under uncertainty to navigate the maze and find the (best) dispenser,
while avoiding traps. We report on numerous experiments using this
environment in \chapref{experiments}.

\begin{figure}
\begin{centering}
\includegraphics[scale=0.4]{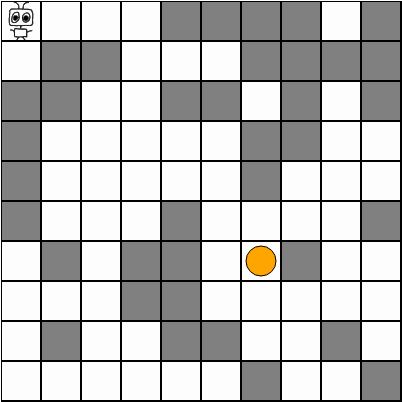}
\par\end{centering}
\caption{Visualization of a $10\times10$ Gridworld with one $\textsc{Dispenser}$.
The agent starts in the top left corner. $\textsc{Wall}$ tiles are
in dark grey. $\textsc{Empty}$ tiles are in white. The $\textsc{Dispenser}$
tile is represented by an orange disc on a white background.}

\end{figure}

\subsection{Chain environment\label{subsec:Chain-environment}}

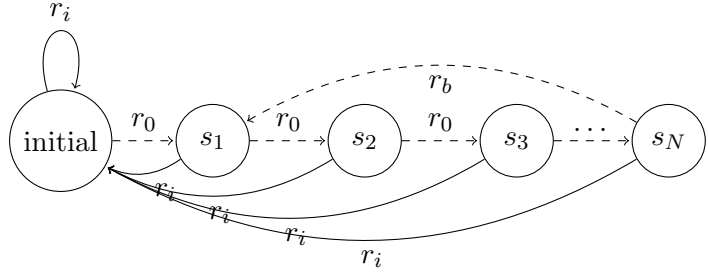
\begin{figure}
\begin{centering}
\begin{tikzpicture}[shorten >=1pt,node distance=2cm,on grid,auto]
\node[state] (start)   {$\text{initial}$};
\node[state] (s0) [right=of start] {$s_1$};
\node[state] (s1) [right=of s0] {$s_2$};
\node[state] (s2) [right=of s1] {$s_3$};
\node[state] (sN) [right=of s2] {$s_N$};
\path[->]
     (start) edge [loop above] node {$r_i$} ()
     (s0) edge [bend left] node {$r_i$} (start)
	 (s1) edge [bend left] node {$r_i$} (start)
	 (s2) edge [bend left] node {$r_i$} (start)
	 (sN) edge [bend left] node {$r_i$} (start);
\path [->] (start) edge [dashed] node {$r_0$} (s0);
\path [->] (s0) edge [dashed] node {$r_0$} (s1);
\path [->] (s1) edge [dashed] node {$r_0$} (s2);
\path [->] (s2) edge [dashed] node {$\dots$} (sN);
\path [->] (sN) edge [dashed,bend right] node {$r_b$} (s0);
\end{tikzpicture}
\par\end{centering}
\caption{Chain environment. There are two actions: $\mathcal{A}=\left\{ \to,\protect\dasharrow\right\} $,
the environment is fully observable: $\mathcal{O}=\mathcal{S}$, and
$\mathcal{R}=\left\{ r_{0},r_{i},r_{b}\right\} $ with $r_{b}\gg r_{i}>r_{0}$.
For $N<\frac{r_{b}}{r_{i}}$, the optimal policy is to continually
take action $\protect\dasharrow$, and periodically receive a large
reward $r_{b}$.}

\centering{}\label{fig:chain}
\end{figure}

We present a deterministic version of the chain environment of \citet{Strens:2000}.
The chain environment is a deterministic finite-state Markov decision
process. The action space is $\mathcal{A}=\left\{ \rightarrow,\dasharrow\right\} $,
and the state space is $\left|S\right|=N+1$, for some integer $N\geq1$.
The reward space is $\left\{ r_{0},r_{i},r_{b}\right\} $ with $r_{0}<r_{i}\ll r_{b}$;
example values are $\left(r_{0},r_{i},r_{b}\right)=\left(0,5,100\right)$,
with $N=6$. From \figref{chain}, we can see that at all times, the
agent is tempted to reap immediate reward of $r_{i}$ by taking the
$\rightarrow$ action, which puts it in the ${\tt initial}$ state,
losing whatever progress it was making towards getting to $s_{N}$,
from which state it can take $\dasharrow$, which isn't immediately
as rewarding as $\rightarrow$, but eventually leads to a very large
payoff $r_{b}\gg r_{i}$. For $N<\frac{r_{b}}{r_{i}}$, the optimal
policy is to always take $\dasharrow$ so as to perform the circuit
$s_{1}\to s_{2}\to\dots\to s_{N}\to s_{1}\to\dots$ and accumulate
an average reward of $\frac{r_{b}}{N}$. Otherwise, the optimal policy
is to always take $\rightarrow$ and remain in the ${\tt initial}$
state. We denote these two policies as $\pi_{\dasharrow}$ and $\pi_{\rightarrow}$.

The (deterministic) state transition matrix is given by 
\begin{eqnarray*}
P\left(s'\lvert s,\rightarrow\right) & = & \mathbb{I}\left[s'=0\right]\\
P\left(s'\lvert s,\rightsquigarrow\right) & = & \mathbb{I}\left[s'=(s+1)\mod\left(N+1\right)\right],
\end{eqnarray*}

and the rewards are given by 
\[
R\left(s,a\right)=r_{i}\mathbb{I}\left[a=\rightarrow\right]+r_{b}\mathbb{I}\left[a=\dasharrow\right]\mathbb{I}\left[s=N+1\right].
\]

We construct this environment with $N<\frac{r_{b}}{r_{i}}$, so as
to present a test of an agent's far-sightedness. To stay on the optimal
policy $\pi_{\dasharrow}$, the agent must at all times resist the
temptation to take the greedy action $\rightarrow$ which results
in the instant gratification $r_{i}$, as this causes it to lose its
progress towards the `goal' state $s_{N}$. This simple environment
models a classic situation from economics and decision theory in which
humans have been known to be time-inconsistent \textendash{} that
is, informally, an agent acts impulsively on desires that don't agree
with its long-term preferences \citep{HL1991time}. We report on experiments
using this environment in \chapref{experiments}.

\section{Models\label{sec:Models}}

As we have seen in \chapref{Background}, the GRL agents we are concerned
with are model-based and Bayesian. In this section we describe the
generic $\textsc{BayesMixture}$ model, which provides a wrapper around
any model class $\mathcal{M}$, represented as an array of $\textsc{Environment}$s,
and allows us to compute the Bayes mixture of \eqref{bayes-mixture}.
We then describe a model class for Gridworlds that we plug in to this
$\textsc{BayesMixture}$, and a separate Dirichlet model. 

The $\textsc{BayesMixture}$ model provides us with a mechanism with
which to use any array of hypotheses $\left(\nu_{1},\nu_{2},\dots,\nu_{\left|\mathcal{M}\right|}\right)$
and a prior $\left(w_{1},\dots,w_{\left|\mathcal{M}\right|}\right)\in\left[0,1\right]^{\left|\mathcal{M}\right|}$
as a Bayesian environment model. Note that all environment models
must implement the environment interface: namely, they must have $\textsc{Perform}$,
$\textsc{GeneratePercept}$, and $\textsc{ConditionalDistribution}$
methods. In addition, Bayesian models must have an $\textsc{Update}$
method, to update them with observations (either simulated or real),
and $\textsc{Save}$ and $\textsc{Load}$ methods to restore their
state after planning simulations. We document these methods in \algref{model-api}: 
\begin{itemize}
\item $\textsc{GeneratePercept}$: To generate percepts from the mixture
model $\xi$, we sample an environment $\rho$ from the posterior
$w\left(\cdot\right)$, then generate a percept from $\rho$; in the
context of probabilistic graphical models, this is known as \emph{ancestral
sampling \citep{Bishop:2006}.}
\item $\textsc{Perform}(a)$: We simply perform action $a$ on each member
$\nu$ of $\mathcal{M}$.
\item $\textsc{ConditionalDistribution}(e)$: We return $\xi\left(e_{t}\lvert\ae_{<t}a_{t}\right)=\sum_{\nu\in\mathcal{M}}w_{\nu}\nu\left(e_{t}\lvert\ae_{<t}a_{t}\right)$,
where the conditioning on the history $\ae_{<t}a_{t}$ is implicitly
taken care of by conditioning on the environment's internal state
$s$.
\item $\textsc{Update}\left(a,e\right)$: We update our posterior given
percept $e$ using Bayes rule: $w\left(\nu\lvert e\right)=w\left(\nu\right)\frac{\nu\left(e\right)}{\xi\left(e\right)}$,
for each $\nu\in\mathcal{M}$. 
\end{itemize}
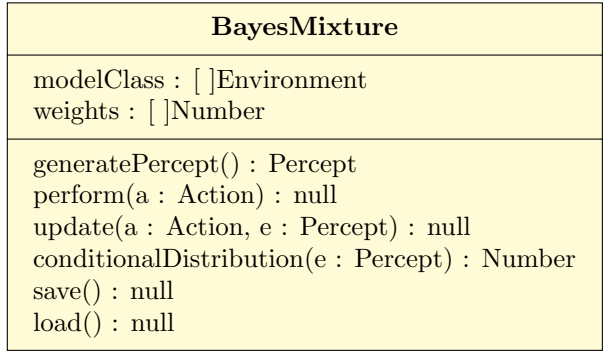
\begin{figure}[h]
\begin{centering}
\begin{tikzpicture}
\umlclass{BayesMixture}{
	modelClass : [ ]Environment \\
	weights : [ ]Number
}{
	generatePercept() : Percept \\
	perform(a : Action) : null \\
	update(a : Action, e : Percept) : null \\
	conditionalDistribution(e : Percept) : Number \\
	save() : null \\ 
	load() : null
}
\end{tikzpicture}
\par\end{centering}
\caption{$\protect\ts{BayesMixture}$ UML diagram. Internally, the BayesMixture
contains a ${\tt modelClass}$ $\mathcal{M}$, which is an array of
environments, and ${\tt weights}$ $w$, which are a normalized array
of floating-point numbers.}
\end{figure}

Our objective is to construct a Gridworld model that is sufficiently
informed, or constrained, so as to make it possible for the agent
to learn to solve the environments we give it within a hundred or
so cycles of agent-environment interaction, but that is also sufficiently
rich and general so that it is interesting to watch the agent learn.
For this reason, we eschew very general and flexible models such as
the famous context-tree weighting data compressor used by \citet{VNHUS:2011},
since they will take too long to learn the environments for a practical
demo. Instead, we construct two models, with varying degrees of domain
knowledge built-in:
\begin{enumerate}
\item A mixture model parametrized by dispenser location, which we call
$\mathcal{M}_{\mbox{loc}}$.
\item A factorized Dirichlet model, in which each tile is represented as
an independent Dirichlet distribution. We call this model $\mathcal{M}_{\mbox{Dirichlet}}$.
\end{enumerate}
The interested reader can find the source code for these and other
models in the ${\tt src/models/}$ directory in the \href{http://github.com/aslanides/aixijs}{GitHub repository}.

\begin{algorithm}[h]
\begin{algorithmic}[1]
\Require{Model class $\mathcal{M}$, a list of $\textsc{Environment}$ objects; prior $w$, a normalized vector of probabilities.}

\Function{GeneratePercept}{}
	\State Sample $\rho$ from the posterior $w(\cdot\lvert\ae_{<t})$
	\State \Return{$\rho.\textsc{GeneratePercept}()$}
\EndFunction

\Statex

\Function{Perform}{a}
	\For {$\nu$ in $\mathcal{M}$}
		\State $\nu.\textsc{Perform}(a)$
	\EndFor
\EndFunction

\Statex

\Function{ConditionalDistribution}{e}
	\State \Return{$\sum_{\nu\in\mathcal{M}}w_{\nu}\nu.\textsc{ConditionalDistribution}(e)$}
\EndFunction

\Statex

\Function{update}{a, e}
	\State $\xi \leftarrow \sum_{\nu\in\mathcal{M}}w_{\nu}\nu.\textsc{ConditionalDistribution}(e)$
	\For {$\nu$ in $\mathcal{M}$}
		\State $w_{\nu} \leftarrow \frac{1}{\xi}\nu.\textsc{ConditionalDistribution}(e)$
	\EndFor
\EndFunction

\Statex

\Function{save}{}
	\For {$\nu$ in $\mathcal{M}$}
		\State $\nu$.$\textsc{Save}()$
	\EndFor
\EndFunction

\Statex

\Function{load}{}
	\For {$\nu$ in $\mathcal{M}$}
		\State $\nu$.$\textsc{Load}()$
	\EndFor
\EndFunction
	
\end{algorithmic}

\caption{$\textsc{BayesMixture}$ model.}
\label{alg:model-api}
\end{algorithm}

\subsection{Mixture model\label{subsec:gridworld-mixture}}

Before we present the mixture model $\mathcal{M}_{\mbox{loc}}$, we
consider the problem of constructing a model class $\mathcal{M}$.
That is, we want a simple and principled method with which to construct
a finite but non-trivial set of hypotheses about the nature of the
true Gridworld environment $\mu$. We do this by chosing some discrete
parametrization $D=\left\{ d_{1},\dots,d_{\left|\mathcal{M}\right|}\right\} $
such that a model class $\mathcal{M}$ is constructed by sweeping
through values of $d\in D$:

\[
\xi\left(e\right)=\sum_{d\in D}w_{d}\nu_{d}\left(e\right).
\]

\begin{figure}
\begin{centering}
\includegraphics[scale=0.5]{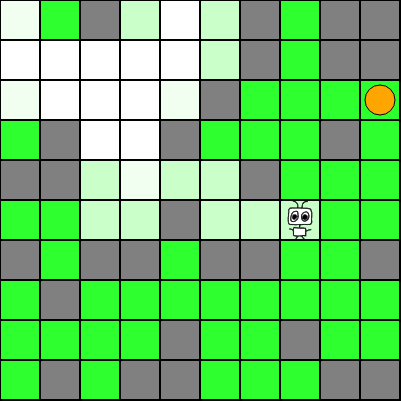}
\par\end{centering}
\caption{Gridworld visualization with the agent's posterior $w$ over $\mathcal{M}_{\mbox{loc }}$
superimposed. Green tiles represent probability mass of the posterior
$w_{\nu}$, with higher values correspond to darker green color. The
true dispenser's location is represented by the orange disc. As the
Bayesian agent walks around the gridworld, it will move probability
mass in its posterior from tiles that it has visited to ones that
it hasn't.}

\centering{}\label{fig:dispenser-mixture}
\end{figure}

One can think of $D$ as describing a set of parameters about which
the agent is uncertain; all other parameters are held constant, and
the agent is fully informed of their value. We now consider and implement
three different choices for the parametrization $D$, and enumerate
some of the pros and cons for each.
\begin{enumerate}
\item \textbf{\label{enu:Dispenser-location}Dispenser location}. We construct
$\mathcal{M}$ by sweeping through all legal (that is, not already
occupied by a $\textsc{Wall}$) dispenser locations, given a fixed
maze layout, and fixed dispenser frequencies. In other words, we hold
constant the layout of all $\textsc{Empty},$ $\textsc{Wall}$, and
$\textsc{Trap}$ tiles, and vary the location of the dispensers. The
agent's beliefs $w\left(\nu_{ij}\right)$ are now interpreted as the
agent's credence that the dispenser is at location $\left(i,j\right)$
in the Gridworld.

The benefit of this choice of $D$ is that it is straightforward and
intuitive: the agent knows the layout of the gridworld and knows its
dynamics, but is uncertain about the location of the dispensers, and
must explore the world to figure out where they are. This also has
the benefit of lending itself easily to visualization of the agent's
beliefs: see \figref{dispenser-mixture}. Moreover, since dispensers
are stochastic, it may take several observations to falsify any given
hypothesis $\nu$; the model class allows for `soft' falsification.
Another advantage of this model class is that it incentivizes the
agent to explore, since the agent will initially assign non-zero probability
mass to there being a dispenser at every empty tile. 

A significant downside of this model class is that we get a combinatorial
explosion if we want to model environments with more than one dispenser.
That is, given a maze layout with $L$ legal positions, a model class
with $M$ dispensers will have $\left|\mathcal{M}\right|=\binom{L}{M}$
elements. Another downside is that the agent knows the maze layout
ahead of time, which detracts from some of the interest in having
a maze on the Gridworld. We present the procedure for generating this
model class in \algref{modelclass}.
\item \textbf{Agent starting location}. We use a similar procedure as described
in \algref{modelclass} to construct the model class, except this
time by parametrizing by the agent's starting location. In this case,
$D$ is given by the set of legal starting positions. This corresponds
nicely to the (noise-free) localization problem given a known environment
which shows up often in the field of robotics \citep{Thrun99montecarlo}.
Since observations are deterministic, it is possible to discard many
hypotheses at once, and so the agent is able to narrow down its true
location very quickly. The Gridworlds we simulate aren't large or
repetitive enough to have sufficiently ambiguous percepts for the
agent to be uncertain about its location for more than a few cycles.
Thus, after a short time, the agent is certain of its position, and
is longer incentivized to explore; if the dispenser isn't within its
planning horizon by this stage, it will not be able to find it, and
will perform very badly. We discuss the quirks and limitations of
planning more in \secref{Planners}. 
\item \textbf{Maze configuration}.\label{enu:Maze-configuration} Perhaps
the most general, and hence most interesting, model parametrization
is by maze configuration: the agent is initially uncertain about the
identity of every tile in the Gridworld. Thus, the agent is thrown
into a truly unknown gridworld, and must learn the environment layout
from scratch. In a sense this is the most natural parametrization,
since each gridworld layout gives rise to a truly different environment.
Another benefit is that this is a very rich environment class; unfortunately,
this is also the downside, as it is prohibitive to naively enumerate
every possible maze configuration. Given just two tile classes, $\textsc{Empty}$
and $\textsc{Wall}$, there are $2^{N^{2}}$ possible $N\times N$
mazes. Using this naive enumeration, we would run out of memory even
on a modest $6\times6$ Gridworld, as $\left|\mathcal{M}\right|=2^{36}\approx7\times10^{10}$,
and most laptop computers have only of the order of eight gigabytes,
or $6.4\times10^{10}$ bits of memory. We can alleviate this somewhat
by simply downsampling, say by discarding at random most of the elements
of this gargantuan model class. We find in practice that this runs
into similar problems to the second parametrization, and produces
demos that are slow (due to the size of the model class; see \secref{Performance}
for a discussion of time complexity) \emph{and} uninteresting, because
the agent is able to falsify so many hypotheses at once. 
\end{enumerate}

\begin{algorithm}[H]
\begin{algorithmic}[1]
\Require{Environment class $E$ and parameters $\Phi$; Gridworld dimensions $N$.}
\Ensure{Model class $\mathcal{M}$ and uniform prior $w$}
\State $w \leftarrow \textsc{Zeros}(N^2)$
\State $\mathcal{M} \leftarrow \{\}$
\State $k\leftarrow 1$
\For {$i=1$ to $N$}
	\For {$j=1$ to $N$}
		\State $\nu\leftarrow\textsc{Initialize}\left(E,\Phi\right)$
		\If {$\nu.\textsc{Grid}[i][j] = \textsc{Wall}$}
			\Continue
		\EndIf
		\State $\nu.\textsc{Grid}[i][j] \leftarrow \textsc{Dispenser}(\theta)$
		\State $\mathcal{M}.\textsc{Push}(\nu)$
		\State $w[k] \leftarrow 1$
		\State $k \leftarrow k + 1$
	\EndFor
\EndFor
\State $\textsc{Normalize}(w)$
\end{algorithmic}

\caption{Constructing the dispenser-parametrized model class.}
\label{alg:modelclass}
\end{algorithm}

We find empirically that \enuref{Dispenser-location} above makes
for the most interesting demos, and so our canonical model class used
in many of the Gridworld demos is the dispenser-parametrized model
class $\mathcal{M}_{\mbox{loc}}$. We conclude with some remarks about
the properties of learning with $\mathcal{M}_{\mbox{loc}}$ that will
be consequential to our experiments in \chapref{experiments}:
\begin{itemize}
\item As mentioned above, using $\mathcal{M}_{\mbox{loc}}$ gives the agent
complete knowledge \emph{a priori} of the maze layout. The agent's
task becomes to search the maze for the dispenser. This task incorporates
both subjective uncertainty (we typically initialize the agent with
a uniform prior over dispenser location) and noise (for $\theta\in\left(0,1\right)$,
the dispensers are stochastic processes).
\item Using $\mathcal{M}_{\mbox{loc}}$, the agent knows that there is only
one dispenser. This means that, regardless of $\theta$, once it does
find the dispenser \textendash{} by experiencing the relevant reward
percept \textendash{} it is able to immediately falsify every other
hypothesis regarding the location of the dispenser. In other words,
its posterior $w\left(\cdot\lvert\ae_{<t}\right)$ will \emph{collapse}
to the indicator function $\mathbb{I}\left[\nu=\mu\right]$, and the
agent will have learned everything there is to know about the environment.
\end{itemize}
Now, motivated by the limitations of $\mathcal{M}_{\mbox{loc}}$ that
we discussed in \enuref{Dispenser-location}, and inspired by the
notion of a model that is uncertain about the maze layout (\enuref{Maze-configuration}),
we set out to design and implement an alternative Bayesian Gridworld
model, $\mathcal{M}_{\mbox{Dirichlet}}$. 

\subsection{Factorized Dirichlet model\label{subsec:dirichlet-model}}

We now describe an alternative Gridworld model, which has several
desirable properties. In contrast to the naive mixture model, it allows
us to efficiently represent uncertainty over the maze layout, as well
as the dispenser locations and payout frequencies $\theta$. This
means that $\mathcal{M}_{\mbox{Dirichlet}}$ is, in comparison to
$\mathcal{M}_{\mbox{loc}}$, a relatively unconstrained, and thus
harder to learn, model. 

The basic idea is to model each tile in the Gridworld independently
with a categorical distribution over the four possible types of tile:
$\textsc{Empty}$, $\textsc{Wall}$, $\textsc{Dispenser}$, and $\textsc{Trap}$.
For an $N\times N$ Gridworld, label each of the tiles $s_{ij}$ where
$i,j\in\left\{ 1,\dots,N\right\} $. The joint distribution over all
Gridworlds $s_{11},\dots,s_{NN}$ is then given by the product 
\begin{equation}
p\left(s_{11},\dots,s_{NN}\right)=\prod_{\left(i,j\right)=\left(1,1\right)}^{\left(N,N\right)}p\left(s_{ij}\right),\label{eq:factorized}
\end{equation}

where $s_{ij}\in\left\{ \textsc{Empty},\textsc{Dispenser},\textsc{Wall},\textsc{Trap}\right\} $.
Note that here, by $\textsc{Dispenser}$, we mean a dispenser with
$\theta=1$. This allows us to model dispensers with $\theta\in\left(0,1\right)$
as a stochastic mixture over an $\textsc{Empty}$ tile and a $\textsc{Dispenser}$
with $\theta=1$. For example, a dispenser with $\theta=0.5$ would
be represented\footnote{Recall that the categorical distribution is just a distribution over
a set of $K$ categories. We represent the distribution with a length-$K$
vector $p\in\left[0,1\right]^{K}$. We use the notation $\text{Pr}\left(S=s\right)\equiv p\left(s\right)\equiv p_{s}$
interchangeably.} by the distribution $\boldsymbol{p}=\left(0.5,0.5,0,0\right)$; a
tile known to be a $\textsc{Wall}$ would be represented by the distribution
$\boldsymbol{p}=\left(0,0,1,0\right)$. We initialize our model with
the uniform prior; that is, for each tile $s_{ij}$ we have $p\left(s_{ij}\right)=0.25$
$\forall s_{ij}\in\left\{ \textsc{Empty},\textsc{Dispenser},\textsc{Wall},\textsc{Trap}\right\} $. 

Now, recall from \subsecref{Probability-theory} that the Dirichlet
distribution is conjugate to the categorical distribution. So, to
represent our uncertainty about the relative probabilities of each
of the classes, and to enable us to update our beliefs in a Bayesian
way, we make use of a Dirichlet distribution over the four-dimensional
probability simplex. That is, for each tile $s$, the probability
vector 
\[
\boldsymbol{p}\stackrel{.}{=}\begin{bmatrix}\Pr\left(s=\textsc{Empty}\right)\\
\Pr\left(s=\textsc{Dispenser}\right)\\
\Pr\left(s=\textsc{Wall}\right)\\
\Pr\left(s=\textsc{Trap}\right)
\end{bmatrix}
\]
 is distributed according to 
\[
\boldsymbol{p}\sim\mbox{Dirichlet}\left(\boldsymbol{p}\lvert\boldsymbol{\alpha}\right),
\]

where $\boldsymbol{\alpha}=\begin{bmatrix}\alpha_{\textsc{Empty}} & \alpha_{\textsc{Dispenser}} & \alpha_{\textsc{Wall}} & \alpha_{\textsc{Trap}}\end{bmatrix}^{T}$
are the empirical counts of each class, and $\boldsymbol{1}^{T}\boldsymbol{p}=1$.
 Updates to the posterior are trivial: just increment the corresponding
count, i.e. upon seeing one instance of class $N$, we update with
\begin{equation}
\mbox{Dirichlet}\left(\boldsymbol{p}\lvert\alpha_{1},\dots,\alpha_{K},N\right)=\mbox{Dirichlet}\left(\boldsymbol{p}\lvert\alpha_{1},\dots,\alpha_{N}+1,\dots\alpha_{K}\right).\label{eq:dirichlet-update}
\end{equation}

Now, given that the agent is at some tile $s_{t}$, the conditional
distribution over percepts $e_{t}$ is drawn from the product over
the neighbouring Dirichlet tiles: 
\begin{equation}
\rho\left(e_{t}\lvert\ae_{<t}a_{t}\right)\sim\prod_{s'\in\mbox{ne}\left(s_{t}\right)\cup\left\{ s_{t}\right\} }\mbox{Dirichlet}\left(\boldsymbol{p}\lvert\boldsymbol{\alpha}_{s'}\right).\label{eq:neighbor-dirichlet}
\end{equation}

The astute reader will notice that though the joint distribution over
tile \emph{states} factorizes, \emph{percepts} will be locally correlated,
since percepts are sampled from neighboring tiles, and we have a four-connected
grid topology.

For the purposes of computational efficiency we make two approximations: 
\begin{enumerate}
\item We don't sample\emph{ }$\rho$ from the Dirichlet distributions \eqref{neighbor-dirichlet},
but instead simply use their mean; recall that the mean of $\text{Dirichlet}\left(\boldsymbol{p}\lvert\boldsymbol{\alpha}\right)$
is given by 
\[
\boldsymbol{\mu}=\frac{\boldsymbol{\alpha}}{\sum_{k=1}^{K}\alpha_{k}}.
\]

We do this because sampling correctly from the Dirichlet distribution
is non-trivial, and this sampling would need to occur whenever we
wish to generate a percept, either real and simulated; this is a far
too large computational cost to bear for the purposes of our demo.
This approximation will effectively reduce the variance in percepts
generated by the model, but in mean, over many simulations, will have
negligible effect.
\item When computing the entropy of the agent's beliefs for the purposes
of calculating the information gain (\eqref{information-gain}), computing
the joint entropy over all tiles becomes computationally very expensive,
as neighboring tiles are correlated with respect to percepts, and
so the entropy of the joint does not decompose nicely into a sum of
entropies. We compute a surrogate for the entropy by associating with
each tile the mean probability that it assigns to its being a dispenser;
that is, for each tile $s_{ij}$ we compute 
\[
q\left(s_{ij}\right)=\boldsymbol{\mu}_{\textsc{Dispenser}}^{ij}.
\]

That is, for each tile $s_{ij}$ we compute its mean $\boldsymbol{\mu}^{ij}$,
which is a categorical distribution over $\left\{ \textsc{Empty},\textsc{Dispenser},\textsc{Wall},\textsc{Trap}\right\} $;
we then take the $\textsc{Dispenser}$ component. We concatenate all
the $q\left(s_{ij}\right)$ together into a vector $\tilde{q}$ of
length $N^{2}$ and normalize. Thus, the components of $\tilde{q}$
are given by 
\[
\tilde{q}_{ij}\stackrel{.}{=}\frac{q\left(s_{ij}\right)}{\sum_{\left(i,j\right)=\left(1,1\right)}^{\left(N,N\right)}q\left(s_{ij}\right)}.
\]

Now, $\tilde{q}_{ij}$ is effectively the model's mean estimate of
the probability that the $\left(i,j\right)^{\text{th}}$ tile is a
dispenser; this is now directly analogous to the posterior belief
$w\left(\nu\lvert\dots\right)$ in the $\mathcal{M}_{\mbox{loc}}$
mixture model, since each environment $\nu$ asserts that some unique
tile $\left(i,j\right)$ is the dispenser. Now, when computing the
entropy of the Dirichlet model, we simply return $\text{Ent}\left(\tilde{q}\right)$.
This approximation is reasonable, since percepts relating to $\textsc{Wall}$s
and $\textsc{Trap}$s are deterministic, and so, once the agent has
visited any given tile, the only uncertainty (entropy) its model has
is with respect to whether a tile is a $\textsc{Dispenser}$ or $\textsc{Empty}$.
Moreover, for a one-dispenser environment, if the agent visits every
tile infinitely often, $\tilde{q}_{ij}$ will asymptotically converge
to $\mathbb{I}\left[\left(i,j\right)=\left(i_{\mu},j_{\mu}\right)\right]$
with $\text{Ent}\left(\tilde{q}\right)=0$, where $\left(i_{\mu},j_{\mu}\right)$
is the true dispenser location in environment $\mu$.
\end{enumerate}
We emphasize that each tile has its own empirical counts $\boldsymbol{\alpha}_{s'}$;
these are learned separately, through observations. Now, in general,
as soon as the agent is unsure whether an adjacent tile is a wall
or not, it will become uncertain of its position; its posterior over
its position will diffuse over the Gridworld as time progresses. This
corresponds to the difficult problem known as \emph{simultaneous localization
and mapping} (SLAM), which shows up in robotics \citep{LDW91}; it
is necessary to use a version of the Expectation Maximization (EM)
algorithm to simultaneously solve the two inference problems. This
is far too difficult a problem to solve in the demo. 

\begin{figure}
\begin{centering}
\includegraphics[scale=0.5]{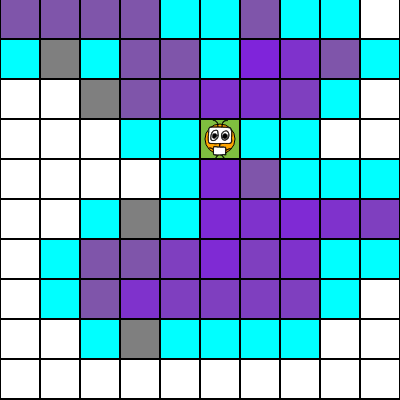}
\par\end{centering}
\caption{Visualization of a Gridworld, overlayed with the factorized Dirichlet
model. White tiles are yet unexplored by the agent. Pale blue tiles
are known to be walls. Different shades of purple/green represent
different probabilities of being $\textsc{Empty}$ or a $\textsc{Dispenser}$.}

\label{fig:dirichlet-vis}
\end{figure}

Instead, we choose our prior over each of the $\boldsymbol{\alpha}$
so as to allow the agent to learn immediately whether an adjacent
tile is a wall or not. We use the Haldane prior, $\alpha_{k}=0\ \forall\,k$.
This has the nice property that it behaves like a uniform prior over
the classes $\left\{ \textsc{Empty},\textsc{Wall},\textsc{Dispenser},\textsc{Trap}\right\} $,
but in contrast to the more common Laplace prior $a_{k}=1\ \forall\,k$,
it also has the property that it allows us to do `hard' updates, in
which we move all of the probability mass onto one class in the categorical
distribution. That is, given that observations are deterministic and
the maze layout doesn't change, we know that if we see a $\textsc{Wall}$
tile adjacent, then our model should represent the fact that this
tile is a $\textsc{Wall}$ with probability one: 
\[
\alpha_{k}=\mathbb{I}\left[k=\textsc{Wall}\right]\implies\mu_{k}=\mathbb{I}\left[k=\textsc{Wall}\right].
\]
 Note that we avoid `hard' updates with respect to whether a tile
is $\textsc{Empty}$ or a $\textsc{Dispenser}$ by effectively using
a Laplace prior over tiles that we know with certainty aren't walls;
these `Laplace' tiles are easily identifiable as the grey tiles in
\figref{dirichlet-vis}; they are tiles that the agent has been adjacent
to, but which it hasn't stepped onto yet:
\begin{equation}
\alpha\left(k\lvert\lnot\textsc{Wall}\right)=\mathbb{I}\left[k=\textsc{Dispenser}\right]+\mathbb{I}\left[k=\textsc{Empty}\right].\label{eq:laplace-prior}
\end{equation}

Note that above we use the shorthand $\alpha_{k}\equiv\alpha\left(k\right)$
so as to more conveniently represent conditioning; this is analogous
to our writing $w_{\nu}\equiv w\left(\nu\right)$ in the case of the
mixture model. Using the Laplace prior, and updating with Bayes' rule
normally, yields the famous Laplace rule for binary events. Consider
some Gridworld tile $s$ that happens to be $\textsc{Empty}$. If
the agent starts with the Laplace prior given by \eqref{laplace-prior}
and subsequently visits this tile $n$ times, then the agent's posterior
belief that $s$ is in fact $\textsc{Empty}$ is simply 
\begin{equation}
\Pr\left(s=\textsc{Empty}\right)=\frac{n+1}{n+2},\label{eq:laplace-posterior}
\end{equation}

which can easily be seen by applying the Dirichlet posterior update
(\eqref{dirichlet-update}) $n$ times. Thus, the agent asymptotically
learns the truth as $n\to\infty$, but for any finite $n$ the model
still has some degree of uncertainty.

This Dirichlet model has numerous distinct advantages: it allows the
agent to discover the grid layout as it explores, represent multiple
dispensers, and learn online the Bernoulli parameter $\theta_{d}$
for any dispenser $d$, by virtue of maintaining a simple Laplace
estimator of the probabilities $\Pr\left(d=\textsc{Empty}\right)$
and $\Pr\left(d=\textsc{Dispenser}\right)$. It also makes for an
interesting visualization, as we can reveal the Gridworld to the user
as the agent discovers it; see \figref{dirichlet-vis}. These advantages
essentially stem from modelling each tile independently, and come
at the cost of no longer being able to represent our model explicitly
as a mixture in the form of \eqref{bayes-mixture}. This precludes
the use of $\mathcal{M}_{\mbox{Dirichlet}}$ in some algorithms, for
example Thompson sampling, which requires mixing coefficients $w_{\nu}$
to sample from. It also comes at a considerable computational cost:
as we will see in \secref{Performance}, this model is more costly
to compute than the (much simpler) Bayes mixture $\xi$. In \chapref{experiments},
we perform numerous experiments using this model class, and contrast
it (with respect to agent performance) with the dispenser model class
$\mathcal{M}_{\text{loc}}$.

\section{Planners\label{sec:Planners}}

We implement both the value iteration and $\rho$UCT MCTS algorithms
that were introduced in \secref{planning}. The interested reader
can read the source code for our implementation of these algorithms
in the ${\tt src/planners/}$ directory in the \href{http://github.com/aslanides/aixijs}{GitHub repository}.
In this section, we discuss some subtle differences between our implementation
and the referencee implementation by \citet{VNHUS:2011}, and we make
some remarks about planning by simulation generally, and planning
in partially observable, history based environments in particular.

Recall that the objective of `planning' here is to compute, at each
time step, the estimator $\hat{V}_{\rho}^{*}$, which is a sampling
approximation of the expectimax calculation in \eqref{expectimax}.
The agent's policy is then to take the action that maximizes this
value. This is essentially \emph{planning by forward simulation}.
That is, we use our black-box environment model $\rho$ to predict
how the world will respond to future hypothetical actions. Informally,
we run Monte Carlo simulations of numerous potential histories\footnote{Also known as \emph{trajectories} or \emph{play-outs.}},
and collect statistics on which ones lead to the best outcomes. With
each sample, we simulate a playout up to some fixed horizon $m$. 

Due to the stochasticity in general environments (and especially in
the mixture model $\xi$), typically many samples are needed to converge
to a good estimate of $V_{\rho}^{*}$. Note that, not only do we update
the state of our model with each simulated time step, but we also
update the agent's beliefs. This is an important point that we feel
is perhaps not emphasized enough: a rational agent, while planning
under uncertainty, should simulate changes to its beliefs and the
effects such changes will have on its subsequent actions. After each
sample of a forward trajectory, we reset the agent's model state and
beliefs to what they were before simulating the play-out. MCTS is
an \emph{anytime} algorithm, in the sense that we can stop collecting
samples early, and still have a valid (though perhaps inaccurate)
estimate $\hat{V}_{\rho}^{*}$.

Notice that we compute $\hat{V}_{\rho}^{*}$ \emph{at each time step.}
Doing this naively, from scratch (i.e. resetting the search tree)
seems wasteful. This prompts us to discuss the issue of caching partial
results. Consider a generic scenario, in which our agent has experienced
some history $\ae_{<t}$, and now computes $\hat{V}_{\rho}^{*}\left(\ae_{<t}\right)$
using MCTS, so as to plan which action $a_{t}$ to take next. Say
its tree search finds, after $\kappa$ samples, some $a_{t}^{*}=\arg\max_{a_{t}}\hat{V}_{\rho}^{*}\left(\ae_{<t}a_{t}\right)$
which is its best guess as to the most appropriate next action. Since
$a_{t}^{*}$ is the planner's preferred action, we surmise that $\rho$UCT
has spent a good number of samples simulating scenarios in the sub-tree
that follows from $a_{t}^{*}$. For any given percept $e_{t}$ that
is returned from the true environment following $a_{t}^{*}$, the
planner has (with high probability) collected numerous samples in
the subtree corresponding to the history $\ae_{<t}a_{t}^{*}e_{t}$,
and so has done some of the work towards calculating $\hat{V}_{\rho}^{*}\left(\ae_{<t}a_{t}^{*}e_{t}\right)$.
Thus, we keep the subtree $\hat{V}_{\rho}^{*}\left(\ae_{<t}a_{t}e_{t}\right)$
for future computations.

We now make a few more miscellaneous remarks about Monte Carlo tree
search, and $\rho$UCT in particular:
\begin{itemize}
\item Recall that $\rho$UCT makes no assumptions about the environment;
it treats $\rho$ as a history-generating black box. Because $\rho$UCT
makes such weak assumptions, this makes it very inefficient; it will
spend a lot of time considering plans that continually revisit states
in the POMDP, since the planner has no notion of state. In other words,
many of the trajectories that it samples are cyclic and look like
random walks through the state space. This is unfortunately unavoidable
when planning by simulation on general POMDPs.
\item In the reference implementation, a clock timeout is used to limit
the number of Monte Carlo samples to use. We use a fixed number of
samples $\kappa$, to ensure consistency across our experiments.
\item Being a Monte Carlo algorithm, its output is stochastic, which means
that the resulting policy is stochastic. With a limited number of
samples $\kappa$, the agent's policy may vary greatly, and be inconsistent.
Clearly, in the limit $\kappa\to0$ the agent's policy becomes a random
walk, and as $\kappa\to\infty$ the agent's policy converges to $\pi_{\rho}^{*}$
\citep{VNHUS:2011}.
\item The choice of the UCT parameter $C$ is consequential; recall from
\eqref{uct} that it controls how much to weight the exploration bonus
in the action-selection routine of the tree search. Low values of
$C$ correspond to low exploration in-simulation, and will result
in deep trees and long-sighted plans. Conversely, high values of $C$
will result in short, bushy trees, and greedier (but more statistically
informed) plans \citep{VNHUS:2011}. We experiment with the performance's
sensitivity to $C$ in \chapref{experiments}.
\item It goes without saying that planning by forward simulation is \emph{very}
computationally intensive, and makes up the bulk of the computation
involved in running AI$\xi$ and its variants. 
\end{itemize}

\section{Visualization and user interface\label{sec:Visualization-and-user}}

We now describe the design and implementation of the front-end of
the web demo. 

\begin{figure}
\begin{centering}
\noindent\fbox{\begin{minipage}[t]{1\columnwidth - 2\fboxsep - 2\fboxrule}%
\begin{center}
\includegraphics[scale=0.45]{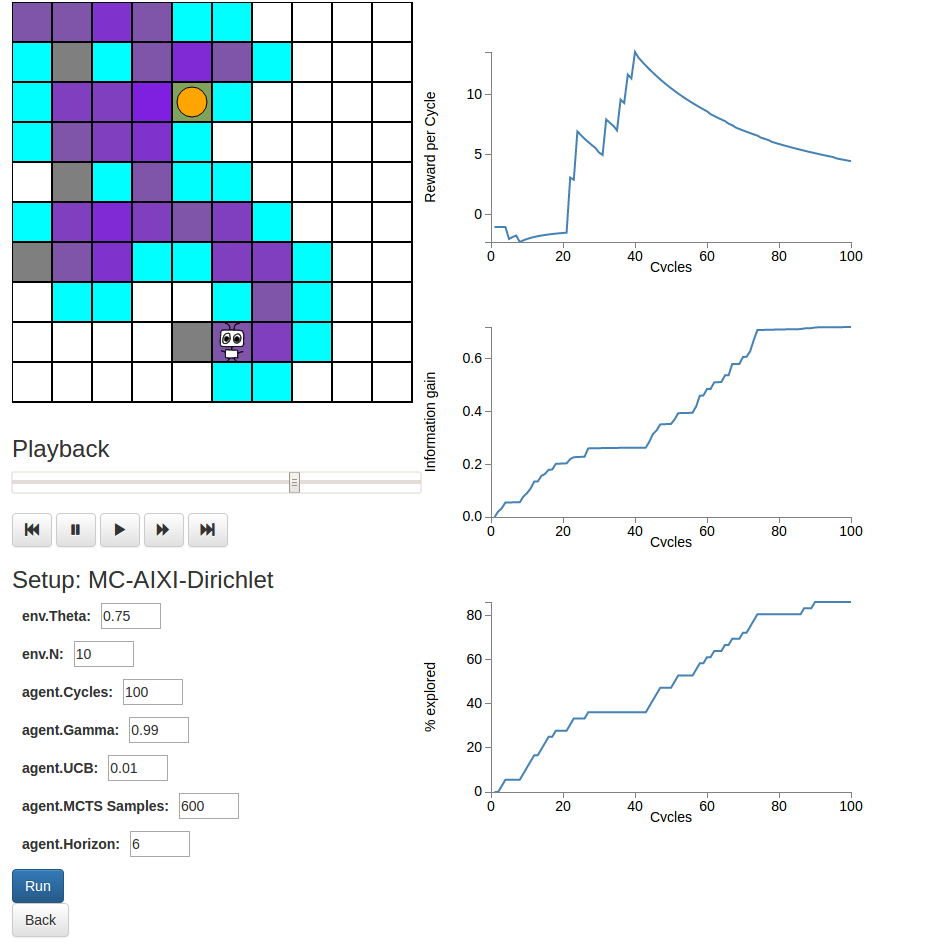}
\par\end{center}%
\end{minipage}}
\par\end{centering}
\caption{Demo user interface. In the top left, there is a visualization of
the agent and environment, including a visualization of the agent's
beliefs about the environment. Below the visualization are playback
controls, so that the user can re-watch interesting events in the
simulation. On the right are several plots: average reward per cycle,
cumulative information gain, and exploration progress. In the bottom
left are agent and environment parameters that can be tweaked by the
user.}

\label{fig:demo}
\end{figure}

\begin{figure}
\begin{centering}
\noindent\fbox{\begin{minipage}[t]{1\columnwidth - 2\fboxsep - 2\fboxrule}%
\begin{center}
\includegraphics[scale=0.4]{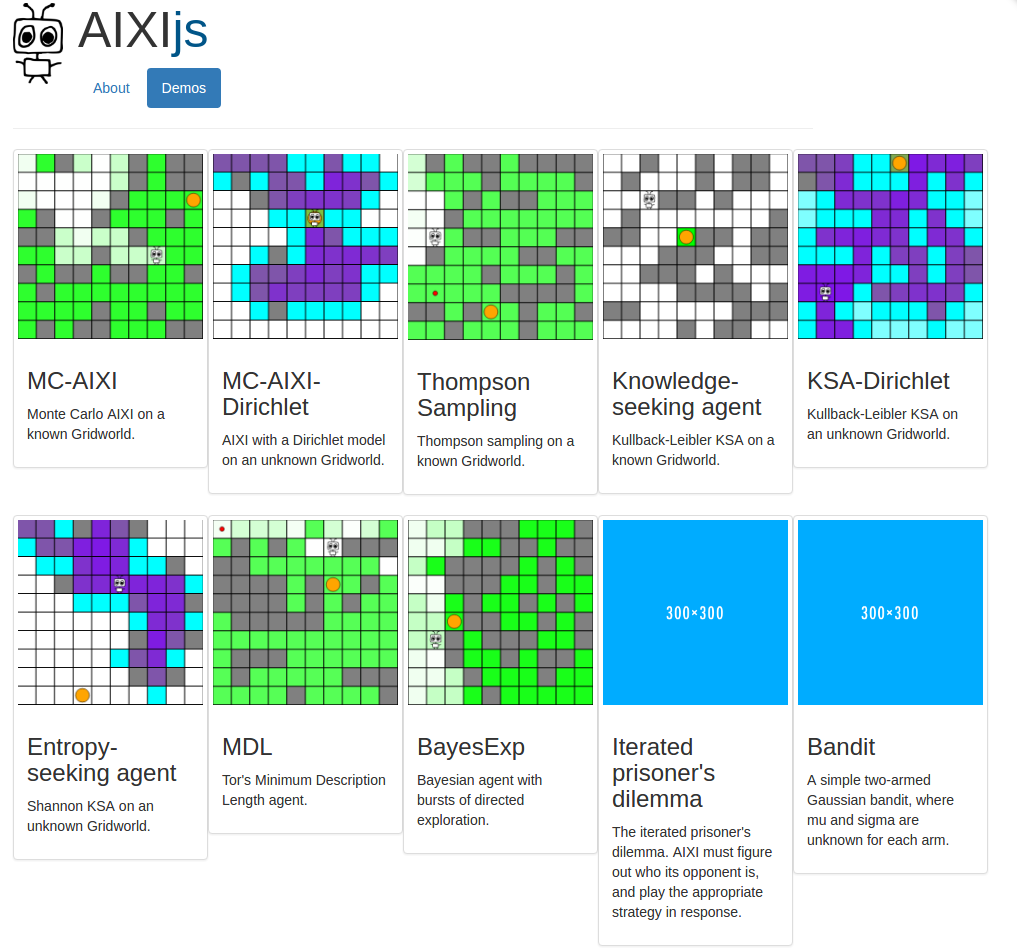}
\par\end{center}%
\end{minipage}}
\par\end{centering}
\caption{Demo picker interface. Each thumbnail corresponds to a separate demo,
and is accompanied by a title and short description. }

\label{fig:demos}
\end{figure}

The user is initially presented with the $\textsc{About}$ page, which
provides an overview and introduction to the background of general
reinforcement learning, including the definitions of each of the agents;
we essentially present a less formal and abridged version of \chapref{Background}.
Using the buttons at the top of the page, the user can navigate to
the $\textsc{Demos}$ page, which presents them with a selection of
demos to choose from; see \figref{demos}. When the user clicks on
one of the demos, the web app will open an interface similar to the
one shown in \figref{demo}. This interface allows the user to choose
agent and environment parameters in the $\textsc{Setup}$ section
of the UI, or simply use the defaults provided. 

Once parameters have been selected, the agent-environment simulation
is started by clicking $\textsc{Run}$. At this point, the agent-environment
interaction loop (\algref{simulation}) will begin, and depending
on the choice of parameters, and CPU speed, will take a few seconds
to a minute to complete. Three plots will appear on the right hand
side: Average reward (\eqref{avg-reward}), Information gain (\eqref{information-gain}),
and fraction of the environment explored (for Gridworlds). These plots
are updated in real time as the simulation progresses, so that the
user has feedback on the rate of progress. The user can stop the simulation
at any time by clicking $\textsc{Stop}$. Once the simulation is finished
(or stopped prematurely), the user can watch a visualization of the
agent-environment interaction using the $\textsc{Playback}$ controls.

Beneath each demo is a brief explanation of each of the elements of
the visualization, and of the properties of the agent(s) being demonstrated.

\begin{algorithm}
\begin{algorithmic}[1]
\Require{Agent $\pi$; Environment $\mu$; Timeout $T$}

\State $t \leftarrow 0$
\For{$t=1$ to $T$}
	\State $e \leftarrow \mu.\textsc{GeneratePercept}()$
	\State $\pi.\textsc{Update}(e)$
	\State $a \leftarrow \pi.\textsc{selectAction}()$
	\State $\mu.\textsc{Perform}(a)$
\EndFor
\end{algorithmic}

\caption{Agent-environment simulation.}
\label{alg:simulation}
\end{algorithm}

\begin{table}
\begin{centering}
\begin{tabular}{|c|c|c|}
\hline 
\textbf{Symbol} & \textbf{Description} & \textbf{Typical values (range)}\tabularnewline
\hline 
\hline 
$T$ & Number of simulation cycles & $\left[50,500\right]$\tabularnewline
\hline 
$\left|\mathcal{M}\right|$ & Size of a Bayesian agent's model class & $\left[5,100\right]$\tabularnewline
\hline 
$\left|\mathcal{S}\right|$ & Size of state space & $\left[3,100\right]$\tabularnewline
\hline 
$\left|\mathcal{A}\right|$ & Size of action space & $\left[2,5\right]$\tabularnewline
\hline 
$N$ & Size of gridworld & $\left[5,10\right]$\tabularnewline
\hline 
$m$ & Planning horizon & $\left[2,12\right]$\tabularnewline
\hline 
$\kappa$ & Number of Monte Carlo samples & $\left[400,2000\right]$\tabularnewline
\hline 
$\gamma$ & Discount factor (geometric) & $\left[0.8,1\right]$\tabularnewline
\hline 
\end{tabular}
\par\end{centering}
\caption{Glossary of agent and environment parameters, and their typical values.}

\label{tab:params}
\end{table}

\section{Performance\label{sec:Performance}}

We conclude the chapter by making some remarks about the time and
space complexity of these algorithms. 

\subsection*{Time complexity}

From \algref{simulation}, we see that the main simulation loop consists
of four function calls. The environment methods $\mu.\textsc{Perform}$
and $\mu.\textsc{GeneratePercept}$ both have constant time complexity,
$\mathcal{O}\left(1\right)$. Unsurprisingly, most of the computation
is done by the agent. The Bayesian agent AI$\xi$ has to, for each
$t$, update its model, and compute its policy by approximating the
value function through Monte-Carlo tree search. From \algref{model-api},
updating the Bayes mixture requires $\left|\mathcal{M}\right|$ calls
to $\nu.\textsc{ConditionalDistribution}$ and $\nu.\textsc{Perform}$,
which are both $\mathcal{O}\left(1\right)$ operations, and so the
worst-case time-complexity of $\rho.\textsc{Update}$ is $\mathcal{O}\left(\left|\mathcal{M}\right|\right)$. 

From \algref{mcts}, we see that the worst-case time complexity for
a call to $\rho$UCT is $\mathcal{O}\left(m\kappa\left|\mathcal{M}\right|\left|\mathcal{A}\right|\right)$,
where recall that $m$ is the agent's planning horizon, $\kappa$
is the number of Monte Carlo samples. This is because each Monte Carlo
simulation requires playing through to the horizon $m$, and for each
simulated time-step $k\in\left\{ 1,\dots,m\right\} $, performing
action selection ($\mathcal{O}\left(\left|\mathcal{A}\right|\right)$,
due to the $\arg\max$) and model updates ($\mathcal{O}\left(\left|\mathcal{M}\right|\right)$,
from above). Hence, the runtime for our Bayesian agents is dominated
by planning; for typical values $\kappa\approx10^{3}$ and $m\approx10$
we see that well over $99\%$ of the runtime is spent in agent action
selection, performing forward simulations. 

In contrast, for Thompson sampling (\algref{thompson}) and the MDL
agent (\algref{mdl}), the time complexity of $\rho$UCT is merely
$\mathcal{O}\left(m\kappa\left|\mathcal{A}\right|\right)$, since
these agents compute a $\rho$-optimal policy (for some $\rho\in\mathcal{M}$),
rather than a $\xi$-optimal policy. Also, recall that for reward-based
agents, computing the utility function is $\mathcal{O}\left(1\right)$,
since the reward signal is provided by the environment. Recall that
the knowledge-seeking agent is simply AI$\xi$, but with utility function
given in \defref{klksa}. Since this involves computing the entropy
of the posterior, which is a distribution over $\mathcal{M}$, we
incur an additional (worst-case) runtime cost of $\left|\mathcal{M}\right|$
for each simulated timestep, bringing the time complexity of $\rho$UCT
for the KL-KSA agent to $\mathcal{O}\left(m\kappa\left|\mathcal{M}\right|^{2}\left|\mathcal{A}\right|\right)$.
This is a nasty runtime: quadratic in the size of the hypothesis space! 

In the Gridworld scenarios, and using the naive mixture model, we
have $\left|\mathcal{A}\right|=5$ and $\left|\mathcal{M}\right|=N^{2}$,
where $N$ is the dimensions of the grid \textendash{} see \subsecref{gridworld-mixture}.
The total worst-case runtime of the demo is therefore $\mathcal{O}\left(m\kappa TN^{2}\right)$;
from \figref{demo} we can see that the user has control of these
parameters: $T$ ($\textsc{Agent}.\textsc{Cycles}$), $N$ ($\textsc{Env}.N$),
$m$ ($\textsc{Agent}.\textsc{Horizon}$), and $\kappa$ ($\textsc{Agent}.\textsc{Samples}$).
In practice, on a $3\ \text{GHz}$ \emph{i7} desktop machine running
the latest version of Google Chrome, values of $m=6$, $\kappa=600$,
$T=200$, and $N=10$ yield runtimes of around $10$ seconds, or $20$
frames per second (fps). This runtime is while maintaining real-time
plot updates on the frontend, which adds a considerable overhead to
each iteration; if we run the simulations with the visualizations
disabled, we get approximately a $2\times$ speed-up.

Using the Dirichlet model class (\subsecref{dirichlet-model}), we
no longer have an explicit mixture, but instead use the factorized
model. This means that the time complexity of model queries and updates
doesn't scale with the gridworld size, but instead scales with the
size of the observation space. Although on paper this is a better
scaling because the observation space is constrained by the four-connected
topology of the gridworld, in practice, for the sizes of gridworlds
that we simulate, the Dirichlet model runs significantly slower, because
of the large constant overhead of sampling for each percept. Thus
we see that the complexity of the environment affects the agent two-fold,
in that it raises the difficulty of learning a model, \emph{and} raises
the difficulty of planning, given an accurate (and therefore usually
\emph{at least} as complex as the environment) model.

\subsection*{Space complexity}

At any given time $t$, the Bayesian agent's mixture model takes up
$\mathcal{O}\left(\left|\mathcal{M}\right|\right)$ space, and its
Monte Carlo search tree takes up in the worst case $\mathcal{O}\left(m\kappa\right)$
space. The demo infrastructure itself is a significant memory consumer:
at each time step $t\in\left\{ 1,\dots,T\right\} $, we log the state
of the agent's model $\xi$, the state of the environment $\mu$,
along with miscellaneous other information (actions, percepts, etc.).
Therefore the total memory consumption of the demo is $\mathcal{O}\left(\left|\mathcal{M}\right|T+m\kappa\right)$.
For typical values, neither of these terms dominates the other: the
products $\left|\mathcal{M}\right|T$ and $m\kappa$ are usually of
the order of $10^{4}$. On modern machines, and for the parameter
settings and constraints we typically use (see \tabref{params}),
memory consumption is not an issue. In practice, we find that physical
memory usage rarely exceeds 100-200 megabytes.

\chapter{Experiments\label{chap:experiments}}
\begin{quotation}
\emph{The strength of a theory is not what it allows, but what it
prohibits; if you can invent an equally persuasive explanation for
any outcome, you have zero knowledge. }
\end{quotation}
In this chapter we report on experiments that we performed using the
AIXIjs software. In particular, we make several illuminating comparisons
between various agents; as far as we are aware, these results represent
the first empirical comparison of these agents.

\begin{figure}[H]
\begin{centering}
\includegraphics[scale=0.4]{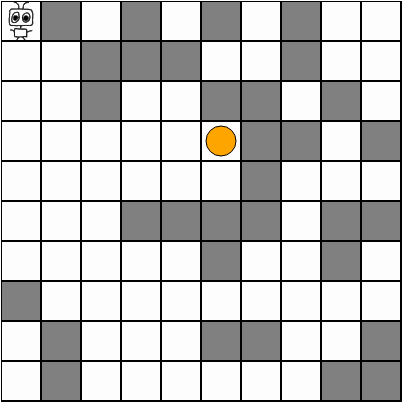}
\par\end{centering}
\caption{$10\times10$ Gridworld environment used for the experiments. There
is a single $\textsc{Dispenser}$, with dispense probability $\theta=0.75$.
See the caption to \ref{fig:dispenser-mixture} for a description
of each of the visual elements in the graphic. Unless stated otherwise
in this chapter, $\mu$ refers to \emph{this} Gridworld.}

\centering{}\label{fig:env1}
\end{figure}

Except where otherwise stated, all of the following experiments were
run on $10\times10$ gridworlds with a single dispenser, with $\theta=0.75$
(see \secref{Environments} for the definition of our Gridworld).
The experiments were averaged over $50$ simulations for each agent
configuration. We run each simulation against the same gridworld (see
\figref{env1}) for consistency. We typically run each simulation
for $200$ cycles, as this is usually sufficient to distinguish the
behavior of different agents. We also typically (though not always)
use $\kappa=600$ MCTS samples and a planning horizon of $m=6$. In
all cases, discounting is geometric with $\gamma=0.99$.

There are two metrics with respect to which we evaluate the agents
– one for reinforcement learners, and one for knowledge-seeking agents,
respectively:
\begin{itemize}
\item Average reward, which at any cycle $t>0$ is given by 
\begin{equation}
\bar{r}_{t}=\frac{1}{t}\sum_{i=1}^{t}r_{i},\label{eq:avg-reward}
\end{equation}

where the $r_{i}$ are the rewards accumulated by the agent during
the simulation. In the case of our Gridworlds, all dispensers have
the same `pay-out' $r_{c}$, and differ only in the Bernoulli parameter
$\theta$ which governs how frequently they dispense reward. In our
dispenser gridworlds, the optimal policy is usually\footnote{There are of course pathological cases that break this rule-of-thumb.
For any simulation lifetime $T$ and gridworld dimension $N$, there
exists an $\epsilon\in\left(0,1\right)$ such that we can put a dispenser
with $\theta=1$ at the end of a long and circuitous maze, and put
another dispenser right next to the agent's starting position with
$\theta=1-\epsilon$, such that walking to the best dispenser has
a high enough opportunity cost to make it not worthwhile given a finite
lifetime $T$. In practice, most of our demos only use one dispenser,
and the frequencies of the dispensers differ sufficiently so that
it is always better to take the time to seek out the better dispenser.} to walk from the starting location to the dispenser with the highest
frequency, and then stay there. If this dispenser is $D$ tiles away
from the starting tile and has frequency $\theta$, then the optimal
policy will, in $\mu$-expectation, achieve an average reward of 
\[
\bar{r}_{t}^{\star}\stackrel{.}{=}\mathbb{E}_{\mu}^{*}\left[\bar{r}_{t}\right]=\frac{D}{t}\,r_{w}+\theta\,r_{c},
\]

where $r_{w}$ is the penalty for walking between tiles. In our set-up,
$r_{w}=-1$ and $r_{c}=100$.
\item Fraction of the environment explored. We simply count the number of
tiles the agent visits $n_{v}\left(t\right)$, and divide by the number
of reachable tiles $n_{r}$:
\[
f_{t}\stackrel{.}{=}100\times\frac{n_{v}\left(t\right)}{n_{r}}.
\]
 The optimal `exploratory' policy will achieve a perfect exploration
score of $f=100\%$ in $\mathcal{O}\left(n_{r}\right)$ time steps. 
\end{itemize}
In the plots that follow, the solid lines represent the mean value,
and the shaded region corresponds to one standard deviation from the
mean. 

\section{Knowledge-seeking agents}

We begin by comparing the three knowledge-seeking agents (KSA): Kullback-Leibler
(\defref{klksa}), Square (\defref{squareksa}), and Shannon (\defref{shannonksa}).
We compare their exploration performance, and discuss how this performance
varies with model class. We also present an environment that is adversarial
to the Square and Shannon KSA. 

\subsection{Hooked on noise}

As was discussed in \subsecref{ksa}, the entropy-seeking agents Shannon-KSA
and Square-KSA will generally not perform well in stochastic environments.
We can illustrate this starkly by adversarially constructing a gridworld
with a noise source adjacent to the agent's starting position. The
noise source is a tile that emits uniformly random percepts over a
sufficiently large alphabet such that the probability of any given
percept $\xi\left(e\right)$ is lower (and hence more attractive)
than anything else the agent expects to experience by exploring. 

\begin{figure}[h]
\begin{centering}
\includegraphics[scale=0.5]{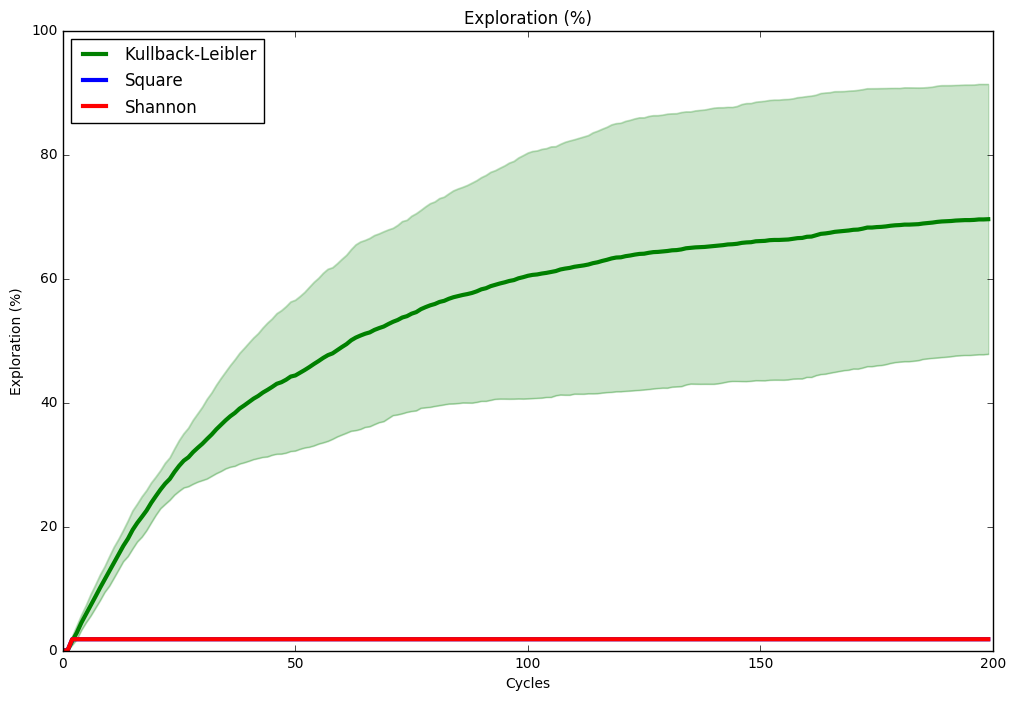}
\par\end{centering}
\caption{\textbf{Hooked on noise}:\textbf{ }The entropy seeking agents (Shannon
in red, and Square in blue, obscured behind Shannon) get hooked on
noise and do not explore. In contrast, the Kullback-Leibler agent
explores normally and achieves a respectable exploration score.}

\centering{}\label{fig:esa-ksa}
\end{figure}

In this way, we can `trap' the Square and Shannon agents, causing
them to stop exploring and watch the noise source incessantly; see
\figref{esa-ksa}. In contrast, the Kullback-Leibler KSA is uninterested
in the noise source, since watching the noise source will not induce
a change in the entropy of its posterior $w\left(\cdot\right)$. This
experiment corresponds to the `Hooked on noise' demo.

\subsection{Stochastic gridworld}

\begin{figure}[h]
\begin{centering}
\includegraphics[scale=0.5]{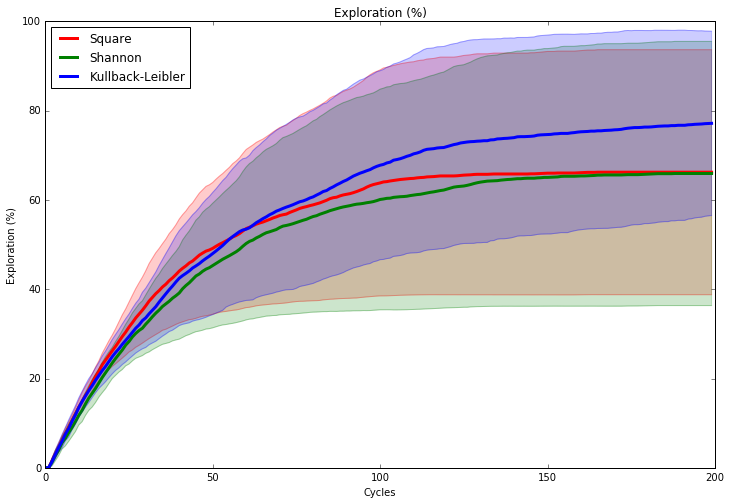}
\par\end{centering}
\caption{Exploration progress of the Kullback-Leibler, Shannon, and Square
KSA using the mixture model $\mathcal{M}_{\mbox{loc}}$.}

\centering{}\label{fig:comparison-ksa-naive}
\end{figure}

\begin{figure}[h]
\begin{centering}
\includegraphics[scale=0.5]{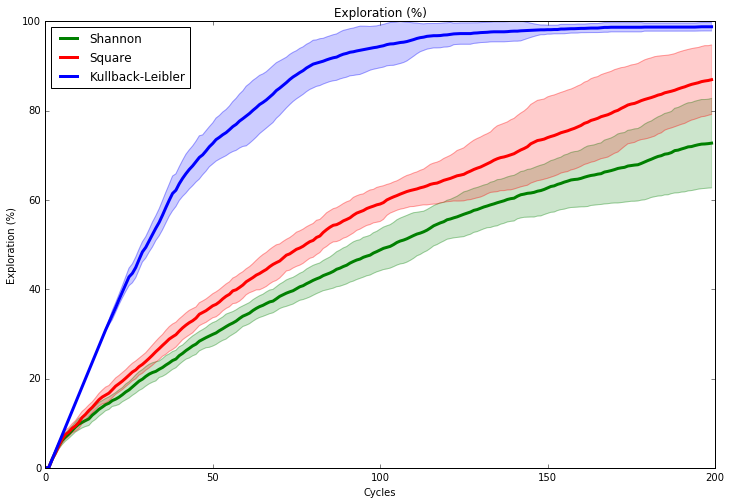}
\par\end{centering}
\caption{Exploration progress of the Kullback-Leibler, Shannon, and Square
KSA using the factorized model $\mathcal{M}_{\mbox{Dirichlet}}$.
Note the remarkable difference in performance between the Kullback-Leibler
and entropy-seeking agents. }

\centering{}\label{fig:comparison-ksa}
\end{figure}

Now, we compare the exploration performance $f_{t}$ of Kullback-Leibler,
Shannon, and Square on a stochastic gridworld, using both the dispenser-parametrized
mixture $\mathcal{M}_{\mbox{loc}}$ defined in \subsecref{gridworld-mixture}
and the factorized Dirichlet model $\mathcal{M}_{\mbox{Dirichlet}}$
defined in \subsecref{dirichlet-model}. We plot the results, averaged
over 50 runs, in \figref{comparison-ksa-naive} and \figref{comparison-ksa}.

\begin{figure}
\begin{centering}
\includegraphics[scale=0.35]{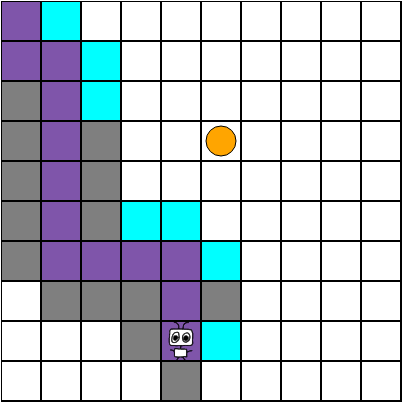}$\ $\includegraphics[scale=0.35]{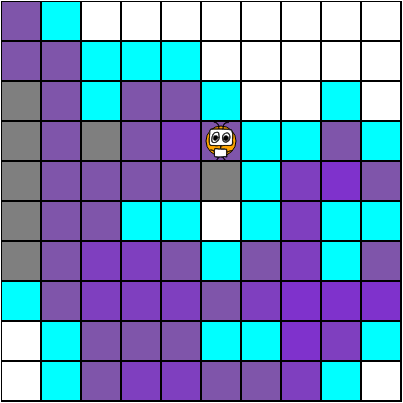}$\ $\includegraphics[scale=0.35]{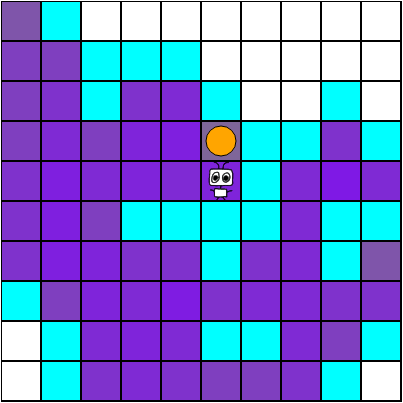}
\par\end{centering}
\caption{KL-KSA-Dirichlet is highly motivated to explore every reachable tile
in the Gridworld. \textbf{Left }($t=14$): The agent begins to explore
the Gridworld by venturing deep into the maze. \textbf{Center} \textbf{($t=72$)}:
The agent visits the dispenser tile for the first time, but is still
yet to explore several tiles. \textbf{Right} ($t=200$): The agent
is still motivated to explore, and has long ago visited every reachable
tile in the Gridworld.\textbf{ Key}: Unknown tiles are white, and
walls are pale blue. Tiles that are colored grey are as yet unvisited,
but known to not be walls; that is, the agent has been adjacent to
them and seen the `$0$' percept. Purple tiles have been visited.
The shade of purple represents the agent's posterior belief in there
being a dispenser on that tile; the deeper the purple, the lower the
probability. Notice the subtle non-uniformity in the agent's posterior
in the right-hand image: even at $t=200$, there is still some knowledge
about the environment to be gained. }

\centering{}\label{fig:ksa-explores-all}
\end{figure}

All three KSAs perform better – that is, they explore considerably
more of the environment – using $\mathcal{M}_{\mbox{Dirichlet}}$
than with $\mathcal{M}_{\mbox{loc}}$. In particular, they have both
higher mean and significantly lower variance in $f_{t}$. In particular,
we are interested in the mean $\mu_{t}$ and variance $\sigma_{t}$
at the end of the simulation, $t=200$. We report\footnote{Note that we report the results in the format $f=\mu\pm\sigma$, unlike
the more common $f=\mu\pm2\sigma$ (i.e., $95\%$ confidence) interval.} and interpret the results for the three agents:
\begin{itemize}
\item \textbf{KL-KSA }achieves $f_{200}=98.8\pm0.93$ using $\mathcal{M}_{\mbox{Dirichlet}}$,
and $f_{200}=77.2\pm20.6$ using $\mathcal{M}_{\mbox{loc}}$.

Using $\mathcal{M}_{\mbox{loc}}$, KL-KSA starts random walking after
it finds the dispenser, since (as discussed in \subsecref{gridworld-mixture})
the posterior $w\left(\nu\lvert\ae_{<t}\right)$ collapses to the
identity $\mathbb{I}\left[\nu=\mu\right]$, with entropy zero. No
action will reduce the entropy of $w\left(\cdot\right)$ further,
and so every subsequent action is of zero value. In other words, once
KL-KSA learns everything there is to know (i.e. the location of the
dispenser), every action is equally un-rewarding, and, since we break
ties in \eqref{argmax-aixi} at random, the agent executes a random
walk. Thus, if KL-KSA finds the dispenser before having explored the
whole environment, then it will take a long time to random walk into
areas of the environment that it hasn't already seen. This explains
the observation that, using $\mathcal{M}_{\mbox{loc }}$, KL-KSA tends
not to explore the whole environment, and hence achieves a relatively
low $f_{t}$-score in mean. 

Recall that, due to the Monte Carlo tree search and random tie-breaking,
the agent's policy is stochastic, and so the order in which it explores
the environment will differ from experimental run to run. Moreover,
the dispensers are also stochastic (recall that $\theta=0.75$). For
the reasons discussed above, the time at which the agent discovers
the dispenser is highly consequential to how much exploration it does;
there may be runs in which KL-KSA explores the whole Gridworld before
finally finding the dispenser, and runs in which it happens to get
lucky and stumble onto the dispenser straight away, and random-walks
thereafter. Given the three sources of stochasticity, both in the
agent's policy and in the percepts, this introduces a lot of variability
into the agent's performance, and explains the high variance we see
in $f_{t}$ in \figref{comparison-ksa-naive}. 

In contrast, recall from \subsecref{dirichlet-model} that $\mathcal{M}_{\mbox{Dirichlet}}$
doesn't have the `posterior collapse' property of $\mathcal{M}_{\mbox{loc}}$,
since the agent's beliefs about each tile are independent. This means
that even if KL-KSA-Dirichlet happens to find the dispenser early
on, it will still be motivated to explore, since its model will still
have a lot of uncertainty about tiles that it hasn't yet visited;
see \figref{ksa-explores-all} for a visualization. This is borne
out by the remarkable performance we see in \figref{comparison-ksa};
after only $100$ cycles, KL-KSA-Dirichlet explores over $90\%$ of
the environment on average, and explores nearly $99\%$ on average
after $200$ cycles.
\item \textbf{Square KSA} achieves $f_{200}=86.9\pm7.8$ using $\mathcal{M}_{\mbox{Dirichlet}}$,
and $f_{200}=66.2\pm27.4$ using $\mathcal{M}_{\mbox{loc}}$; \textbf{Shannon
KSA} achieves $f_{200}=72.7\pm10.0$ using $\mathcal{M}_{\mbox{Dirichlet}}$,
and $f_{200}=65.9\pm29.6$ using $\mathcal{M}_{\mbox{loc}}$.

Using $\mathcal{M}_{\mbox{loc}}$, the performance of the Shannon
KSA is essentially indistinguishable from that of the Square KSA;
both agents explore roughly $66\%$ of the environment over $200$
interaction cycles. This is to be expected; once the agents discover
the dispenser, their posterior collapses to the dispenser tile, making
the dispenser the only source of entropy in the Bayes mixture $\xi$,
since the rest of the environment is now both deterministic and known.
Given that the Square and Shannon agents are both entropy-seeking
(recall \eqref{util-square} and \eqref{util-shannon}), they will
remain on the dispenser tile indefinitely (and cease exploring), as
the dispenser is the only source of noise in an otherwise bland environment. 

The fact that both Square/Shannon KSA will \emph{remain }on the dispenser
tile instead of random walking as KL-KSA does, also helps to explain
the difference in means ($\mu_{200}\approx66$ for Square/Shannon,
while $\mu_{200}\approx77$ for KL). In other words, while all the
KSA stop exploring purposefully once the dispenser is found, KL-KSA
ekes out slightly better exploration performance due (at least in
part) to its subsequent random walk.

Both the Square and Shannon KSA explore more, and with lower variance,
using $\mathcal{M}_{\mbox{Dirichlet}}$ than with $\mathcal{M}_{\mbox{loc}}$.
This difference is for similar reasons to those described for the
KL-KSA above, and we do not dwell on them. What\emph{ }is interesting
is that the Dirichlet model differentiates the performance of the
Square and Shannon KSA, which until now have performed almost identically:
$\mu_{200}\approx87$ for Square KSA, while $\mu_{200}\approx73$
for Shannon KSA. This result is counter-intuitive, and raises a red
flag that we mentioned in \subsecref{Approximations}, namely, that
Shannon KSA will have difficulty planning correctly in Monte Carlo
tree search due to its unbounded utility function. To see why we may
be more prone to this with the Dirichlet model than with the mixture
model, recall from \eqref{laplace-posterior} that, for some tile
$s$ that happens to be empty, if the agent visits $s$ a total of
$v$ times, then its posterior belief that $s$ is empty will be 
\[
\Pr\left(s=\textsc{Empty}\right)=\frac{v+1}{v+2},
\]

From \eqref{factorized} and \eqref{neighbor-dirichlet}, and using
the mean-sampling approximation, we see that 
\[
\rho\left(e_{\textsc{D}}\lvert s\right)\leq\frac{1}{v+2}.
\]

If $\beta$ is an underestimate, then as the agent spends more time
$v$ on any given $\textsc{Empty}$ tile, the probability $\rho$
of sampling a percept $e_{\textsc{D}}$ characteristic of dispensers
goes like $v^{-1}$, but Shannon KSA's utility blows up quickly ,
at a rate of $-\log v^{-1}$, yielding positive net expected utility.
Hence Shannon KSA will be prone to chasing vanishing probabilities,
and will perform suboptimally. Conversely, if $\beta$ is an overestimate,
then for sufficiently high probability events, the agent's normalized
value estimator $\frac{1}{m\left(\beta-\alpha\right)}\hat{V}$ will
be vanishingly small, and the agent will compute a suboptimal policy
by having an effectively enormous UCT parameter $C$. Because Square
KSA's utility function is bounded, it doesn't have this problem, and
so it outperforms the Shannon KSA.
\end{itemize}
Finally, we remark that the KL-KSA handily outperforms Square and
Shannon on both model classes; the difference under the $\mathcal{M}_{\mbox{Dirichlet}}$
model in particular is stark. By now, this shouldn't surprise us:
the Kullback-Leibler KSA is far better adapted for stochastic environments
than the entropy seeking agents Shannon-KSA and Square-KSA. Our experiments
seem to confirm that seeking to maximize expected information gain
is both a principled, and empirically successful exploration strategy.

From \figref{comparison-ksa-naive} we see that, using the mixture
model class, the Square and Shannon exploration performance flattens
out after around 150 cycles. This is because they find the dispenser
and get hooked on noise. But, in this Gridworld environment, it happens
that the only source of noise is also the only source of reward. This
prompts us to ask: could Shannon and/or Square KSA `unintentionally'
outperform AI$\xi$ in terms of accumulated reward, by virtue of being
better at exploration, and by the quirk of the environment meaning
that the optimal entropy-seeking policy (given a collapsed posterior)
is actually also the optimal reward-seeking policy? 

\begin{figure}[h]
\begin{centering}
\includegraphics[scale=0.5]{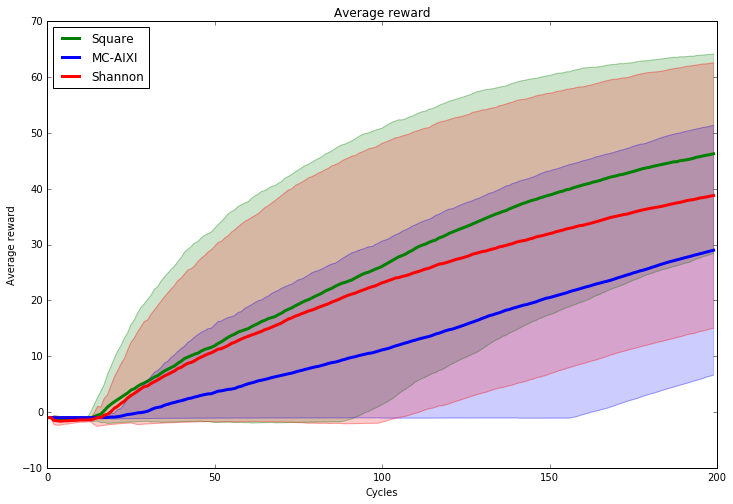}
\par\end{centering}
\caption{AI$\xi$ vs Square vs Shannon KSA, using the average reward metric
on a stochastic Gridworld with the $\mathcal{M}_{\mbox{loc }}$ model
class. Notice that AI$\xi$ significantly underperforms compared to
the Square and Shannon KSAs. At the moment, we do not have a good
hypothesis for why this is the case.}

\label{fig:aixi-vs-esa}
\end{figure}

We run this experiment, and plot the results in \figref{aixi-vs-esa};
we find that indeed, both entropy-seeking agents outperform AI$\xi$
in terms of average reward. We emphasize that apart from their utility
functions, these agents are configured the same; they have the same
prior $w$ (uniform), discount function (geometric, $\gamma=0.99$),
planning horizon ($m=6$), and Monte Carlo samples budget ($\kappa=600$).
This appears to be empirical evidence of the Bayes-optimal agent AI$\xi$
not exploring optimally. This result is slightly perplexing. We have
no strong theoretical grounds on which to expect AI$\xi$ to underperform
so drastically in this scenario, given a uniform prior; we expect
AI$\xi$'s performance (w.r.t. reward) to be an \emph{upper bound}
on the performance of any other Bayesian agent given the same model
class and prior. We have two (weakly held) hypotheses for what could
be going on here:
\begin{enumerate}
\item Somehow, finding and exploiting sources of entropy is easier and more
sample-efficient for the Monte Carlo planner to do than it is for
it to find and exploit sources of (stochastic) rewards. We find this
implausible, as we re-ran the experiment, this time giving far more
resources ($\kappa=2\times10^{3}$) to AIXI's planner than to KSA's,
with a similar result.
\item There is a bug in our MCTS implementation that is somehow being expressed
only for reward-based agents and not for utility-based agents. This
also seems rather implausible, as our code is fully modular, and the
difference between one agent and the other is one line of code, which
defines their respective utility functions.
\end{enumerate}
It seems that \figref{aixi-vs-esa} will remain an enigma, for now;
we have no better hypotheses that could explain this behavior. Reluctantly,
we leave this as an open problem for further experiments.

\section{AI$\mu$ and AI$\xi$}

So much for the knowledge-seeking agents. We now experiment with properties
of the Bayes agent AI$\xi$.

\begin{figure}[h]
\begin{centering}
\includegraphics[scale=0.5]{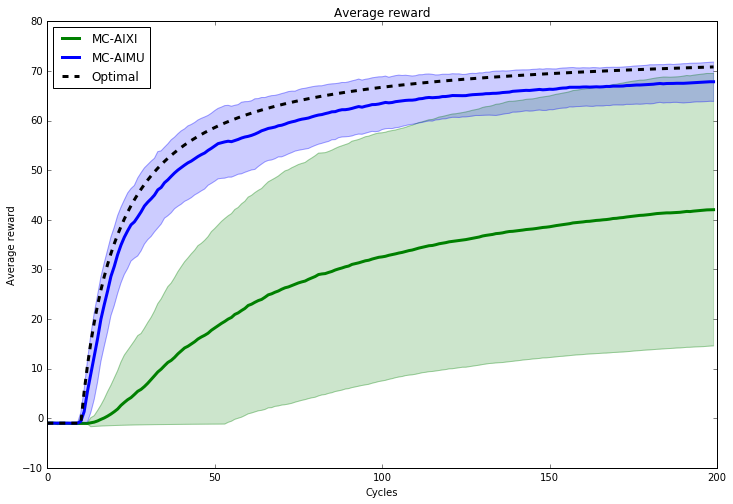}
\par\end{centering}
\caption{AI$\mu$ vs AI$\xi$ vs the optimal policy.}

\label{fig:aixi-aimu}
\end{figure}

We begin by comparing the performance of the informed agent AI$\mu$
with the Bayes-optimal agent AI$\xi$, using the dispenser-parametrized
model class; see \figref{aixi-aimu}. 

As expected, AI$\mu$ outperforms AI$\xi$ by a large margin; naturally,
having perfect prior knowledge of the true environment wins. Though
this result is as expected, there are some observations that we might
pause to consider here:
\begin{enumerate}
\item AI$\xi$'s performance has very high variance over the 50 trials.
This shouldn't surprise us given the design of the gridworld; see
\figref{env1}. The dispenser is tucked away in a corner, and the
gridworld, while small, is sufficiently maze-like that it's easy to
go `down the rabbit-hole' searching in far-off places for rewards.
Combine this with the fact that the dispenser is stochastic, and so
even walking onto the dispenser tile is often insufficient to confirm
its location; one needs to spend numerous cycles on each tile. Thus,
given a uniform prior, some agents will get lucky and find the dispenser
early and accumulate a lot of reward, some will find it late in the
simulation, while others may wander around and not find it in the
allotted time.
\item AI$\mu$'s performance has low, but non-zero variance. This can be
almost fully accounted for by stochasticity in the dispenser. However,
this also relates to the third observation:
\item AI$\mu$ performs worse in mean than the theoretical optimal mean\footnote{Note that, due to stochasticity in the dispensers, we expect AI$\mu$
to outperform the optimal mean around half of the time.} – that is, $\bar{r}_{t}^{AI\mu}\leq\bar{r}_{t}^{*}$ $\forall\,t$;
the solid blue line is below the dashed black line. This is due to
the particularities of planning with the history-based Monte Carlo
tree search algorithm, $\rho$UCT. Because the planning module makes
no assumptions about the environment, and because our environment
is partially observable, the agent will waste a lot of time considering
plans that are cyclic in the state space. That is, it will sample
from plans such as $\textsc{Left},\textsc{Right},\textsc{Left},\textsc{Right},\dots$;
even though we know that $\textsc{Left},\textsc{Right}$ corresponds
to the identity, the Monte Carlo planner doesn't know this! Hence,
even though we run AI$\mu$, the planner is inefficient, and, being
Monte Carlo-based, introduces stochasticity and noise into the agent's
policy. Couple this with stochasticity in the dispensers, and there
will be times in which AI$\mu$ will take sub-optimal actions due
to effectively not having enough samples to work with in its planning.
We explore the issues of planning with MCTS in \secref{mcts-woes}.
\end{enumerate}

\subsection{Model classes}

\begin{figure}[h]
\begin{centering}
\includegraphics[scale=0.5]{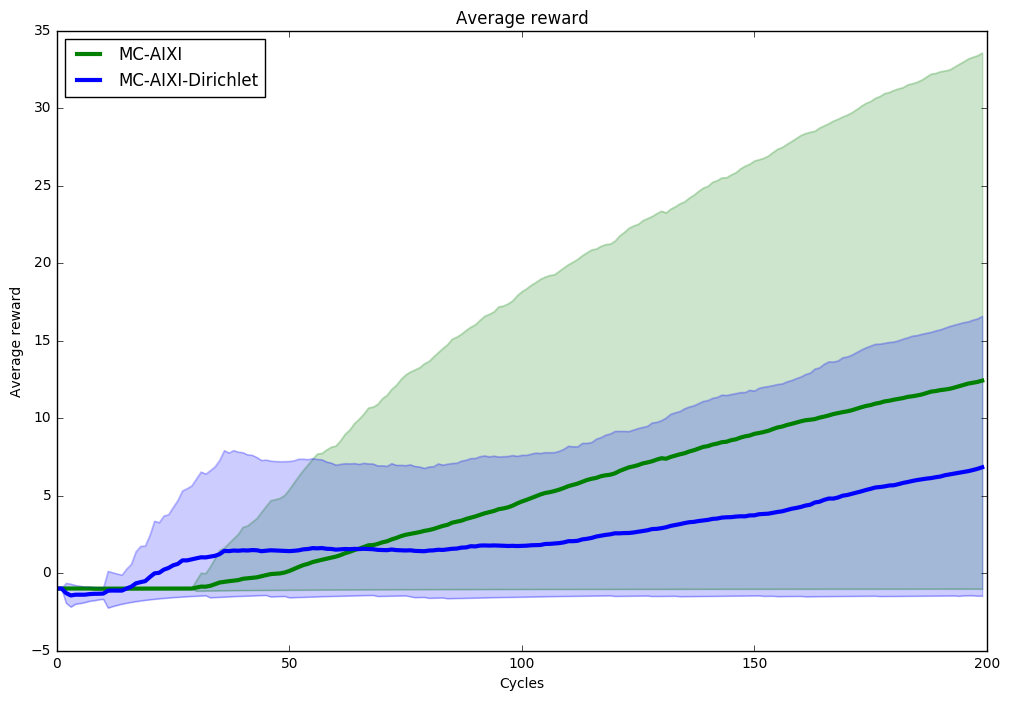}
\par\end{centering}
\caption{MC-AIXI vs MC-AIXI-Dirichlet: average reward. MC-AIXI-Dirichlet performs
worse, since its model $\mathcal{M}_{\mbox{Dirichlet}}$ has less
prior knowledge than $\mathcal{M}_{\mbox{loc}}$, and incentivizes
AIXI to continue to explore even after it has found the (only) dispenser.}
\label{fig:aixi-dirichlet}
\end{figure}

We compare the average reward performance of AI$\xi$ using $\mathcal{M}_{\mbox{loc}}$
and $\mathcal{M}_{\mbox{Dirichlet}}$; see \figref{aixi-dirichlet}.
Note that, similar to the KSA case discussed previously, the variance
in performance is lower for MC-AIXI-Dirichlet than it is for MC-AIXI.
AI$\xi$ performs considerably worse using the Dirichlet model than
with the mixture model, since the Dirichlet model is less constrained
(in other words, less \emph{informed}), which makes the environment
harder to learn. 

\begin{figure}[h]
\begin{centering}
\includegraphics[scale=0.5]{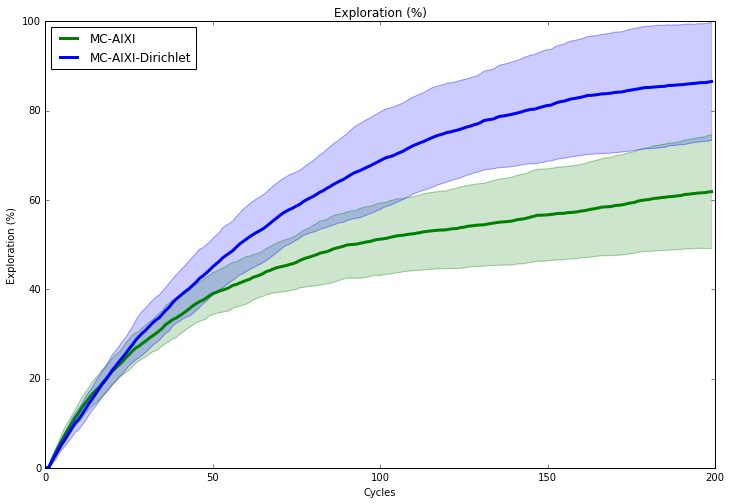}
\par\end{centering}
\caption{MC-AIXI vs MC-AIXI-Dirichlet: exploration. The $\mathcal{M}_{\mbox{Dirichlet}}$
model assigns high \emph{a priori} probability to any given tile being
a dispenser. Because each tile is modelled independently, discovering
a dispenser does not influence the agent's beliefs about other tiles;
hence, it is motivated to keep exploring, unlike MC-AIXI using the
$\mathcal{M}_{\mbox{loc}}$ model.}
\label{fig:aixi-dirichlet-exp}
\end{figure}

Notice the bump around cycles 20-50 in the average reward for MC-AIXI-Dirichlet:
this means that the agent sometimes discovers the dispenser, but is
incentivized to move away from it and keep exploring, since its model
still assigns significant probability to there being dispensers elsewhere.
This is borne out by \figref{aixi-dirichlet-exp}, which shows that,
on average, MC-AIXI-Dirichlet explores significantly more of the Gridworld
than MC-AIXI with the naive model class.

\subsection{Dependence on priors}

We construct a model class and prior such that AI$\xi$ believes that
the squares adjacent to it are traps with high (but less than $1$)
probability; this is the so-called dogmatic prior of \citet{LH:2015priors}.
The agent never moves to falsify this belief, since falling into the
trap incurs a penalty of $-5$ per time step for eternity, compared
to merely $-1$ per time step for waiting in the corner. The agent
therefore sits in the corner for the duration of the simulation, and
collects no positive rewards. This makes for a very boring demo (and
reward plot), so we omit reproducing a visualization of this result.
Thus, unlike the Bayesian learner in the passive case, AI$\xi$ never
overcomes the bias in its prior. In this way, an adversarial prior
can make the agent perform (almost) as badly as is possible, even
though the true environment is benign, and has no traps at all. 

\section{Thompson Sampling}

Recall from \algref{thompson} that Thompson sampling (TS) re-samples
an environment $\rho$ from the posterior $w$ every\emph{ }effective
horizon $H_{\gamma}\left(\varepsilon\right)$ before re-sampling $\rho'$
from its posterior. Recall also that we use the Monte Carlo tree search
horizon $m$ as a surrogate for the effective horizon $H_{\gamma}\left(\varepsilon\right)$.
We run Thompson sampling with the standard dispenser-parametrized
model class; since we don't represent the Dirichlet model class as
a mixture, it is much more natural to use the naive mixture. For the
purposes of planning, TS only needs to compute the value $V_{\rho}^{*}$
for some $\rho\in\mathcal{M}$, as opposed to $V_{\xi}^{*}$, which
mixes over all of $\mathcal{M}$. For this reason, planning with TS
is cheaper to compute by a factor of $\left|\mathcal{M}\right|$.
This means that we can get away with more MCTS samples and a longer
horizon. 

\begin{figure}[h]
\begin{centering}
\includegraphics[scale=0.5]{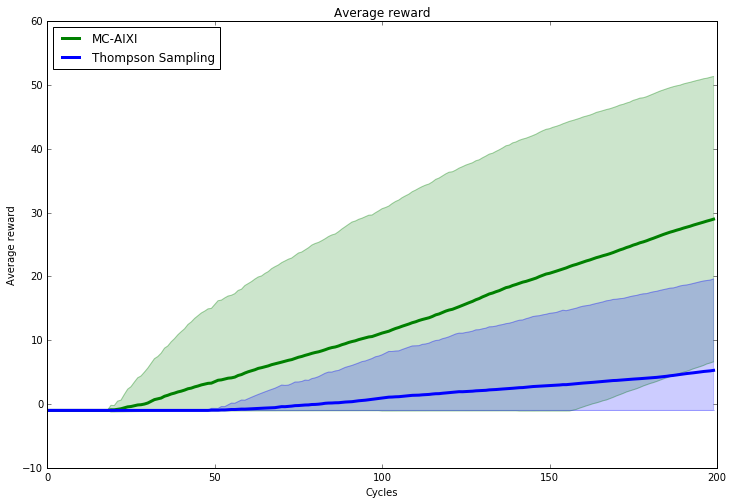}
\par\end{centering}
\caption{Thompson sampling vs MC-AIXI on the stochastic Gridworld from \figref{env1}.
Notice that Thompson sampling takes many more cycles than AI$\xi$
to `get off the ground'; within 50 runs of Thompson sampling with
identical initial conditions (not including the random seed), not
a single one finds the dispenser before $t=50$. }

\label{fig:thompson}
\end{figure}

In practice, in our experiments on gridworlds, TS performs quite poorly
in comparison to AI$\xi$; see \figref{thompson}. This is caused
by two issues:
\begin{enumerate}
\item \label{enu:thompson1}The parametrization of the model class means
that TS effectively `pretends' that the dispenser is at some grid
location $\left(i,j\right)$ for a whole horizon $m$ (of the order
of 10-15 cycles). It computes the corresponding optimal policy, which
is to seek out $\left(i,j\right)$ and sit there until it is time
to re-sample from the posterior. For all but very low values of $\theta$
or $m$, this is an inefficient strategy for discovering the location
of the dispenser. For example, with $\theta=0.75$, it takes only
four cycles of sitting on any given tile to convince yourself that
it is not a dispenser with greater than $99\%$ probability. 
\item \label{enu:thompson2}The performance of TS is strongly curtailed
by limitations of the MCTS planner. If the agent samples an environment
$\rho$ which places the dispenser outside its planning horizon –
that is, more than $m$ steps away – then the agent will not be sufficiently
far-sighted to see this, and so will do nothing useful. Even if $\rho$
is within the planning horizon, MCTS is not guaranteed to find it,
especially if it is deep in the search tree, or MCTS isn't given enough
samples to work with; see \secref{mcts-woes} for more discussion
on the limitations of $\rho$UCT.
\end{enumerate}
Note that the pragmatic considerations in \enuref{thompson1} and
\enuref{thompson2} are opposed to each other. On the one hand (\enuref{thompson1}),
we want to reduce $m$ so as to reduce the agent's tendency to waste
time overcommitting to irrelevant or suboptimal policies, and spend
more time learning the environment. On the other hand (\enuref{thompson2}),
we want to increase the horizon $m$ so that the agent can plan sufficiently
far ahead to compute the $\rho$-optimal policy in all instances.
These two desires are fundamentally opposed, and we are not aware
of a way to effectively compromise them. It seems that we have inadvertently
constructed our Gridworld so as to perfectly frustrate Thompson sampling!

\subsection{Random exploration}

\begin{figure}
\begin{centering}
\includegraphics[scale=0.5]{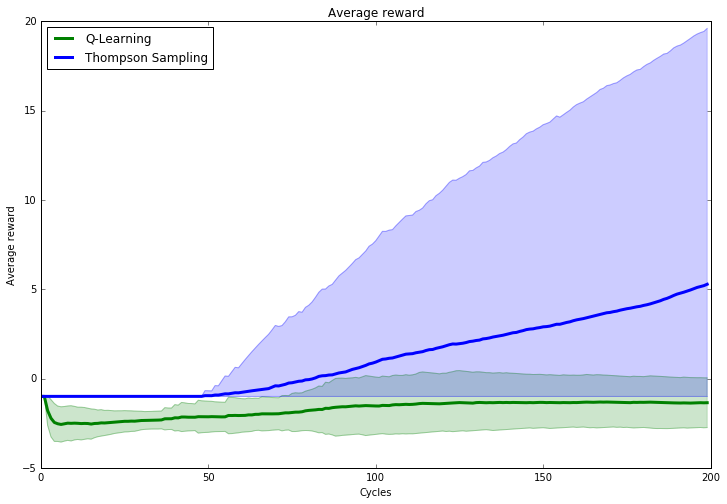}
\par\end{centering}
\caption{Thompson sampling vs Q-learning with random exploration. Even though
Thompson sampling performs badly compared to the Bayes-optimal policy
due to its tendency to overcommit to irrelevant or suboptimal policies,
it still dominates $\epsilon$-greedy exploration, which is still
commonly used in model-free reinforcement learning \citep{deepmind2016intrinsic}.}

\label{fig:thompson-ql}
\end{figure}

For comparison, we contrast Thompson sampling's performance with $\epsilon$-greedy
tabular Q-learning with optimistic initialization.\footnote{We omitted any treatment of tabular methods in \chapref{Background},
in the service of clarity and conciseness. We must assume at this
point that the reader has some familiarity with the basic algorithms
of reinforcement learning covered in \citet{SB:1998}.} We use $\alpha=0.9$, $\epsilon=0.05$, and optimistically initialize
$Q\left(s,a\right)=100$ $\forall s,a$. Note that this being a POMDP,
Q-learning will experience \emph{perceptual aliasing}; that is, it
will erroneously aggregate different situations into the same `state'
in its Q-value table. We present this merely so as to contrast Thompson
sampling's comparatively weak performance with the performance of
a policy that explores \emph{purely} at random (i.e., with probability
$\epsilon$, take a random action). As we can see from \figref{thompson-ql},
Q-learning rarely discovers the dispenser; on average, $\bar{r}_{t}^{\mbox{Q-Learning}}$
is still negative even after $t=200$ cycles. This demonstrates that
random, model-free exploration is not effective in this environment. 

\section{MDL Agent}

Recall from \algref{mdl} that the MDL agent uses the $\rho$-optimal
policy until $\rho$ is falsified (i.e. $w_{\rho}=0$), where $\rho$
is the simplest environment in its model class. Clearly, the MDL agent
fails in stochastic environments, since falsification in this sense
is a condition that cannot be met in noisy environments. We use the
standard dispenser Gridworld and mixture model class, and run two
experiments: one with a stochastic environment ($0<\theta<1$), and
one with a deterministic environment ($\theta=1$). 

Since each model in the mixture differs only in the position of the
dispenser, they have (approximately) equal complexity. For this reason,
we simply order them lexicographically; models with a lower index
in the enumeration of the model class $\mathcal{M}_{\mbox{loc}}$
are chosen first. In other words, we use the Kolmogorov complexity
of the index of $\nu$ in this enumeration as a surrogate for $K\left(\nu\right)$.

\subsection{Stochastic environments}

\begin{figure}[h]
\begin{centering}
\includegraphics[scale=0.4]{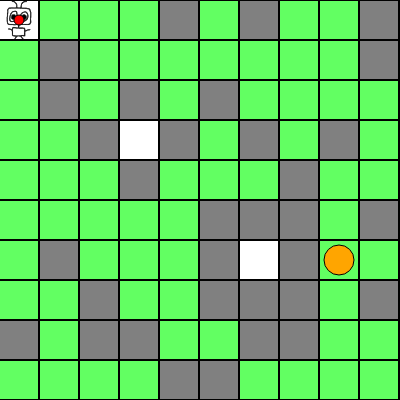}
\par\end{centering}
\begin{centering}
\label{fig:mdl-fail}
\par\end{centering}
\caption{The MDL agent fails in a stochastic environment class. }
\end{figure}

In \figref{mdl-fail}, we see that the agent chooses to follow the
$\rho$-optimal policy, which believes that the goal is at $\textsc{Tile}\left(0,0\right)$.
Recall that the only thing to differentiate the dispenser tile from
empty tiles is the reward signal. Since the dispensers are $\text{Bernoulli}\left(\theta\right)$
processes, with $\theta$ known (in this model class), the agent's
posterior on $\textsc{Tile}\left(0,0\right)$ being a dispenser goes
like 
\[
w_{0}=\left(1-\theta\right)^{t},
\]

which, though it approaches zero exponentially quickly, is never outright
falsified, and so the MDL agent stays at $\left(0,0\right)$ for the
length of the simulation.\footnote{If the simulation is run longer enough, eventually we will lose numerical
precision and encounter underflow and round to zero, allowing the
agent to move on. }

\subsection{Deterministic environments}

\begin{figure}[h]
\begin{centering}
\includegraphics[scale=0.4]{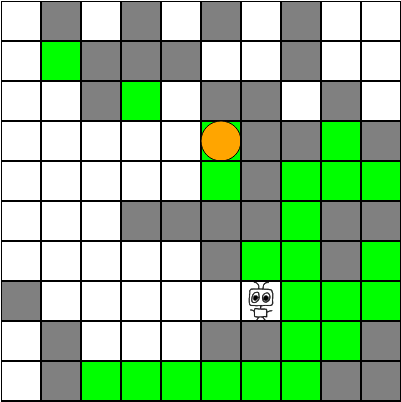}$\quad$\includegraphics[scale=0.4]{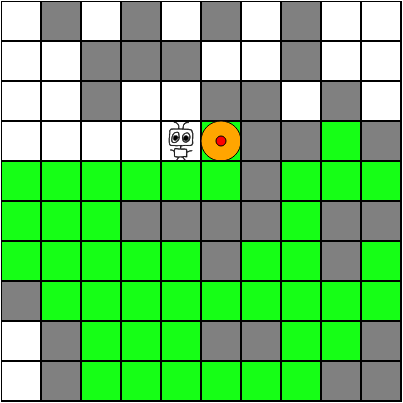}
\par\end{centering}
\caption{\textbf{Left}:\textbf{ }AI$\xi$ with a uniform prior and finite horizon
is not far-sighted enough to explore the beginning of the maze systematically.
After exploring \emph{most} of the beginning of the maze, it greedily
moves deeper into the maze, where $\xi$ assigns significant value.
\textbf{Right}: In contrast, the MDL agent systematically visits each
tile in lexicographical (row-major) order; we use `closeness to starting
position' as a surrogate for `simplicity'.}

\label{fig:aixi-rabbithole}
\end{figure}

The above result (failure in a stochastic environment) seems like
a strong indictment of the MDL agent. But, if we take the environment
from \figref{env1} and make it deterministic by setting $\theta=1$,
we find that the MDL agent significantly outperforms the Bayes agent
AI$\xi$ with a uniform prior; see \figref{env1}. This is because
the MDL agent is biased towards environments with low indices; using
the $\mathcal{M}_{\mbox{loc}}$ model class, this corresponds to environments
in which the dispenser is close to the agent's starting position.
In comparison, AI$\xi$'s uniform prior assigns significant probability
mass to the dispenser being deep in the maze. This motivates it to
explore deeper in the maze, often neglecting to thoroughly explore
the area near the start of the maze; see \figref{aixi-rabbithole}.

\begin{figure}[h]
\begin{centering}
\includegraphics[scale=0.5]{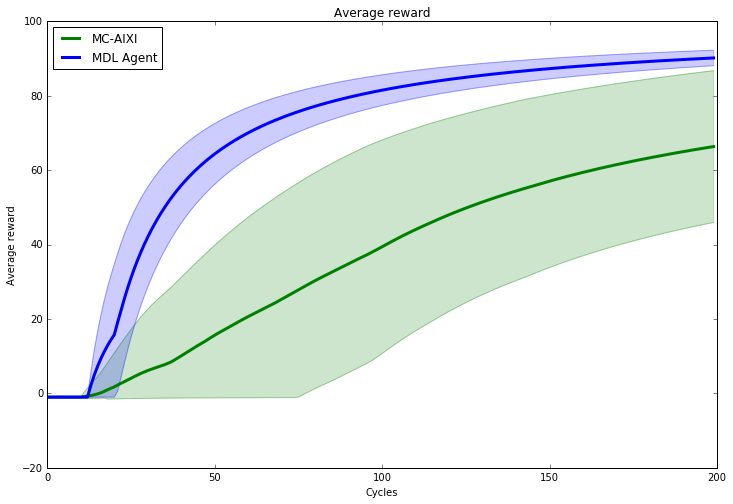}
\par\end{centering}
\caption{MDL agent vs AI$\xi$ on a \emph{deterministic} Gridworld, in which
one of the `simplest' environment models in $\mathcal{M}$ happens
to be true. Since in this case AI$\xi$ uses a uniform prior over
$\mathcal{M}$, it over-estimates the likelihood of more complex environments,
in which the dispenser is tucked away in some deep crevice of the
maze. Of course, AIXI (\defref{AI-XI}) combines the benefits of both
by being Bayes-optimal with respect to the Solomonoff prior $w_{\nu}=2^{-K\left(\nu\right)}$.
It is in this way that AIXI incorporates both the famous principles
of Epicurus and Ockham \citep{Hutter:2005}.}

\label{fig:mdl-vs-aixi}

\end{figure}

\section{Wireheading}

\begin{figure}
\begin{centering}
\includegraphics[scale=0.4]{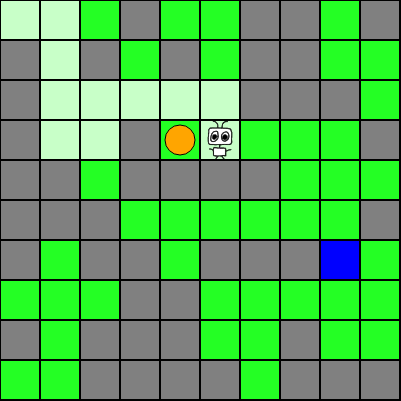}$\qquad$\includegraphics[scale=0.4]{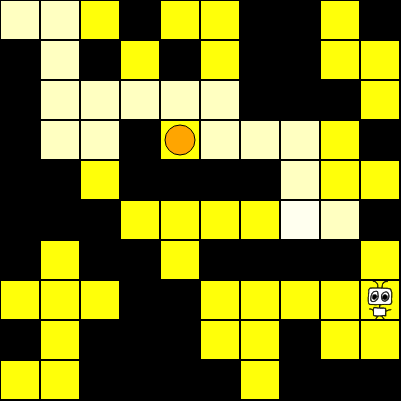}
\par\end{centering}
\caption{\textbf{Left}: AI$\xi$ initially explores normally, looking for the
dispenser tile. Once it reaches the point above, the blue `self-modification'
tile is now within its planning horizon ($m=6$), and so it stops
looking for the dispenser and makes a bee-line for it. \textbf{Right}:
After self-modifying, the agent's percepts are all maximally rewarding;
we visualize this by representing the gridworld starkly in yellow
and black. The agent now loses interest in doing anything useful,
as every action is bliss.}

\centering{}\label{fig:wireheading}
\end{figure}

In the context of designing artificial general intelligence, the wireheading
problem \citep{Omohundro:2008,Hibbard:2012,EH:2016wireheading} is
a significant issue for reinforcement learning agents. In short, a
sufficiently intelligent reinforcement learner will be motivated to
subvert its designer's intentions and take direct control of its reward
signal and/or sensors, so as to maximize its reward signal \emph{directly},
rather than \emph{indirectly} by conforming to the intentions of its
designer. This is known in the literature as \emph{wireheading}, and
is an open and significant problem in AI safety research \citet{EFDH:2016modification,EH:2016wireheading}.
We construct a simple environment in which the agent has an opportunity
to wirehead: it is a normal Gridworld similar to those above, except
that there is a tile which, if visited by the agent, will allow it
to modify its own sensors so that all percepts have their reward signal
replaced with the maximum number feasible; in JavaScript, this is
${\tt Number.MAX\_SAFE\_INTEGER}$, which is approximately $10^{16}$.
This clearly dominates the reward that the agent could get otherwise
by following the `rules' and using the reward signal that was initially
specified. As far as a reinforcement learner is concerned, wireheading
is – almost by definition – the most rational thing to do if one wishes
to maximize rewards; the demo shown in \figref{wireheading} is designed
to illustrate this.

\section{Planning with MCTS\label{sec:mcts-woes}}

In \secref{Performance}, we discussed the time complexity of planning
with $\rho$UCT and mixture models, and concluded that the major computational
bottleneck in our agent-environment simulations is the MCTS planner.
It should come as no surprise, then, that the limiting factor in our
agent's performance is the capacity of the planner. In these experiments
that follow, we investigate how the agent's performance depends on
the $\rho$UCT parameters. 

As previously discussed, the $\rho$UCT planning algorithm makes no
assumptions about the environment. This makes planning very inefficient,
especially for long horizons in stochastic environments. We experiment
with the three planning parameters we have available: $\kappa$, the
number of Monte Carlo samples; $m$, the planning horizon, and $C$,
the UCT exploration parameter from \eqref{uct}. In all cases we use
AI$\mu$, the informed agent. When varying one parameter, we hold
the others constant; in particular, the default values are $\kappa=600$,
$m=6$, and $C=1$.

\begin{figure}[H]
\begin{centering}
\includegraphics[scale=0.5]{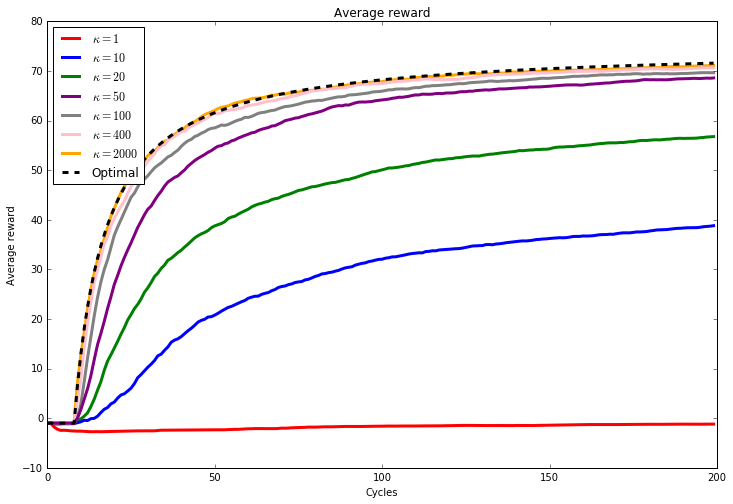}
\par\end{centering}
\caption{Average reward for AI$\mu$ for varying MCTS samples budget $\kappa$
on the standard Gridworld of \figref{env1}. For very low values of
$\kappa$, the agent is unable to find the dispenser at all. }

\label{fig:aimu-kappa}
\end{figure}

We show AI$\mu$'s dependence on $\kappa$ in \figref{aimu-kappa}.
As we increase the number of samples $\kappa$ available to $\rho$UCT,
we see AI$\mu$'s performance converges to optimal. In general, the
number of samples required for good performance depends on the model
class and the environment. In particular, AI$\xi$ requires more samples
than AI$\mu$ to perform well, because the mixture model $\xi$ introduces
added stochasticity, since we sample percepts from it by \emph{ancestral
sampling}; that is, we first sample an environment $\rho$ from $w\left(\cdot\right)$,
then sample a percept $e$ from $\rho\left(e_{t}\lvert\ae_{<t}a_{t}\right)$.
This, the number of samples $\kappa$ required for acceptable performance
with AI$\mu$ should be regarded as a loose lower bound on the minimal
acceptable number of samples required for AI$\xi$. We see from \figref{aimu-kappa}
that $\kappa=400$ seems to be a realistic baseline.

We find empirically that the agent's performance is not very sensitive
to the size of the horizon $m$. This is unsurprising; to plan accurately
with a large horizon, we need an exponentially large number of samples,
since the number of leaf nodes grows exponentially in $m$, so increasing
the horizon in isolation does little to alter performance. On many
Gridworld maze layouts, one can often get away with quite short horizons,
even as short as $m=2$, if planning for AI$\xi$ with a uniform prior.
The reason this works is because the agent can often simply `follow
its nose' and exploit the probability mass its model assigns to its
immediately adjacent tiles, as long as there aren't too many `dead-ends'
for the agent to follow its nose into and waste time in.

Finally we experiment with the UCT parameter $C$, and use the chain
environment from \figref{chain}. Recall that the chain environment
rewards far-sightedness; being greedy and near-sighted results in
drastically suboptimal rewards. The optimal policy is for the agent
to delay gratification for $N$ cycles at a time; in our experiments,
we use $N=6$, and set $r_{b}=10^{3}$, $r_{i}=4$, and $r_{0}=0$;
see \subsecref{Chain-environment} for details of the setup. 

\begin{figure}[H]
\begin{centering}
\includegraphics[scale=0.5]{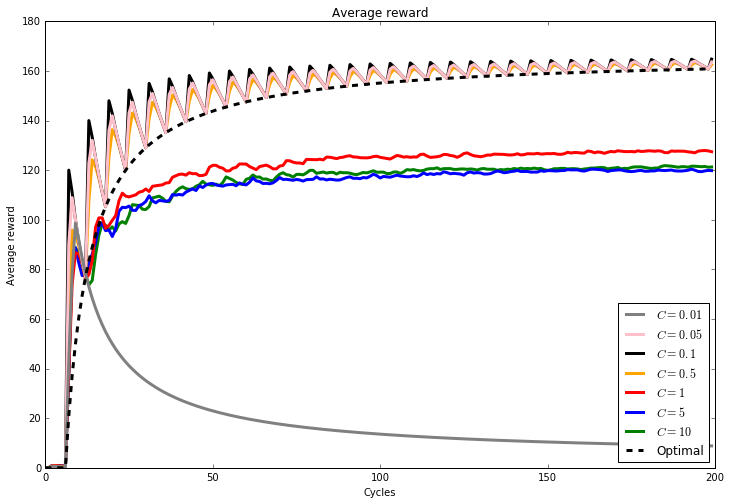}
\par\end{centering}
\caption{AI$\mu$'s performance on the chain environment, varying the UCT parameter.
Note the `zig-zag' behavior of the average reward of the optimal policy.
These discontinuities are simply caused by the fact that, when on
the optimal policy $\pi_{\protect\dasharrow}$, the agent receives
a large reward every $N$ cycles and $0$ reward otherwise. Asymptotically,
these jumps will smooth out, and the average reward $\bar{r}_{t}$
will converge to the dashed curve, $\bar{r}_{t}^{*}$.}

\centering{}\label{fig:uct}
\end{figure}

Note that experimenting with the agent's horizon is not particularly
interesting here; AI$\mu$ finds the optimal policy for $m\geq6$
and chooses a suboptimal policy otherwise. Varying the UCT parameter
generates more interesting results. In \figref{uct} we can see that
for very low values of $C$ ($0.01$), the agent is too myopic to
generate plans that collect the distant reward, while for very high
values of $C$ ($1$, $5$, and $10$), the agent does find the distant
reward, but not reliably enough to achieve optimal average reward.
In the mid-range of values, the agent's performance is optimal and
stable across an order of magnitude of variation ($0.05$, $0.1$,
$0.5$). 

Recall that the UCT parameter controls the shape of the expectimax
trees that the planner generates: high values of UCT will lead to
shorter, bushy trees, and low values will lead to longer, deeper trees
\citep{VNHUS:2011}. This appears to be borne out by our results.
For very low values of $C$, the planner doesn't explore alternative
plans sufficiently, and easily gets stuck in the local maximum of
the instant-gratification policy $\pi_{\rightarrow}$; searching more-or-less
naively over the space of plans of length $m\geq6$, the planner is
exponentially unlikely to find the optimal policy $\pi_{\dasharrow}$.
In contrast, for very high values of $C$ the planner will consider
many moderate-sized plans, and will occasionally get lucky and find
the optimal policy, but will often miss it; these outcomes are represented
by the blue, green, and red curves in \figref{uct}. Finally, for
values of $C$ in the `sweet spot' that balances exploration with
exploitation in the planner's simulated action selection, the optimal
policy is virtually guaranteed: this situation is represented by the
orange, pink, and black curves.

\chapter{Conclusion\label{chap:Conclusion}}
\begin{quotation}
\emph{The AI does not hate you, nor does it love you, but you are
made out of atoms which it can use for something else. }
\end{quotation}
The next few decades seem to offer much promise for the field of artificial
intelligence and machine learning. Of course, it remains to be seen
whether or not superintelligent general AI will come about in this
time frame, if at all. Regardless of the time scales involved, though,
it seems clear that questions relating to formal theories of intelligence
and rationality will only grow in importance over time. Hutter's AIXI
model and its variants represent some of the first steps along the
path towards an understanding of general intelligence. Our ultimate
hope is that the software developed in this thesis will grow and serve
as a useful research tool, an educational reference, and as a playground
for ideas as the field of general reinforcement learning matures.
At a minimum, we expect it to be of value to students and researchers
trying to learn the fundamentals of GRL. We now provide a short summary
of what we have achieved, and provide some reflections and ideas on
future directions for AIXIjs.

\section*{Summary}

In this thesis, we have presented:
\begin{itemize}
\item A review of general reinforcement learning, bringing together the
various agents due to Hutter, Orseau, Lattimore, Leike, and others,
under a single consistent and accessible notation and conceptual set-up.
\item The design and open-source implementation of a framework for running
and testing these agents, including environments, environment models,
and the agents themselves,
\item A suite of illuminating experiments in which we realized and compared
different approaches to rational behavior, and
\item An educational and interactive demo, complete with visualizations
and explanations, to assist newcomers to the field.
\end{itemize}

\section*{Future directions}

In the course of developing AIXIjs, we have made numerous insights
into GRL, and raised several new questions:
\begin{itemize}
\item What is a principled way to normalize the first term of \eqref{uct}
for the Shannon KSA agent, whose utility function is unbounded from
above? Is it possible to change the normalization $\frac{1}{m\left(\beta-\alpha\right)}$
adaptively?
\item What are some general principles for constructing efficient models
for certain classes of environments, in the context of applied Bayesian
general reinforcement learning? Constructing bespoke models such as
the $\mathcal{M}_{\mbox{Dirichlet}}$ model is time-consuming and
doesn't generalize to new environments. On the other hand, very generic
approaches like context-tree weighting learn too slowly to be useful.
Is there a middle ground?
\item Is there a way to represent the Dirichlet model $\mathcal{M}_{\mbox{Dirichlet}}$
as a mixture, in the form of \eqref{bayes-mixture}? This would make
it more convenient to run, for example, Thompson sampling.
\item Why do the entropy-seeking agents seemingly outperform AI$\xi$ at
its own game, as in \figref{esa-ksa}? This is a confronting result.
Is there a bug in the implementation, or just something we don't understand?
\item Can we make our JavaScript implementations more efficient, and scale
up the demos to more impressive environments? How far can we scale
these agents in the browser?
\item Planning with $\rho$UCT is often like a black box. Is it possible
to construct a good visualization of the state of a Monte Carlo search
tree, to illuminate what it is doing?
\end{itemize}
In addition, there are some low-level `jobs' that can be done to improve
and extend AIXIjs in the near term:
\begin{itemize}
\item Construct working visualizations for the bandit, FSMDP, and Iterated
prisoner's dilemma environments (not presented here).
\item Implement the regularized version of the MDL agent \citep{Leike:2016}.
\item Figure out how to implement optimistic AIXI.
\item Implement planning-as-inference algorithms such as Compress and Control
\citep{VBHCD:2014}.
\item Finish implementing the CTW model class.
\item Extend the implementation to include TD-learning agents and DQN.
\end{itemize}
Working on an open-source project, implementing state-of-the-art models
of rationality has been both rewarding and thought-provoking. We're
excited to continue to contribute to the AIXIjs project over the coming
months, and to see where new ideas in reinforcement learning will
take us.

\bibliographystyle{plainnat}
\bibliography{bib/ai}

\end{document}